\documentclass{ipol}

\usepackage[font={sf,footnotesize}]{caption}
\usepackage{graphicx}
\usepackage{amsmath}
\usepackage{amssymb}
\usepackage{caption}
\usepackage{subcaption}
\usepackage{float}
\usepackage{listings}
\usepackage{parskip}
\usepackage{ulem}
\usepackage{transparent}
\usepackage[ruled,vlined,lined,linesnumbered]{algorithm2e}

\SetCommentSty{mycommfont}

\usepackage[affil-sl]{authblk}

\usepackage{cleveref}

\ipolSetTitle{CS-TRD: a Cross Sections Tree Ring Detection Method
}
\ipolSetAuthors{Henry Marichal\ipolAuthorMark{1},
Diego Passarella,\ipolAuthorMark{2}
Gregory Randall\ipolAuthorMark{1}}
\ipolSetAffiliations{%

\ipolAuthorMark{1} Instituto de Ingeniería Eléctrica, Facultad de Ingeniería, Universidad de la República, Uruguay(\texttt{henry.marichal@fing.edu.uy / randall@fing.edu.uy})

\ipolAuthorMark{2} Sede Tacuarembó, CENUR Noreste, Universidad de la República, Uruguay(\texttt{diego.passarella@cut.edu.uy})
}
\begin{document}

\begin{ipolAbstract}
This work describes a Tree Ring Detection method for complete Cross-Sections of Trees (CS-TRD) that detects, processes, and connects edges corresponding to the tree's growth rings. The method depends on the parameters for the Canny Devernay edge detector ($\sigma$), a resize factor, the number of rays, and the pith location. The first five are fixed by default. The pith location can be marked manually or using an automatic pith detection algorithm. Besides the pith localization, CS-TRD is fully automated and achieves an F-Score of 89\% in the UruDendro dataset (of Pinus taeda) and of 97\% in the Kennel dataset (of Abies alba) without specialized hardware requirements. 
\end{ipolAbstract}

\begin{ipolCode}
A Python 3.11 implementation of CS-TRD is available on the web page of this article\footnote{\url{https://ipolcore.ipol.im/demo/clientApp/demo.html?id=77777000390}}. Usage instructions are included in the \verb|README.md| file of the archive. The associated online demo is accessible through the website.
\end{ipolCode}

\ipolKeywords{image edge detection, dendrochronology, tree ring detection}

\section{Introduction}
\label{sec:intro}
\begin{figure*}
\begin{center}
   \includegraphics[width=0.3\textwidth]{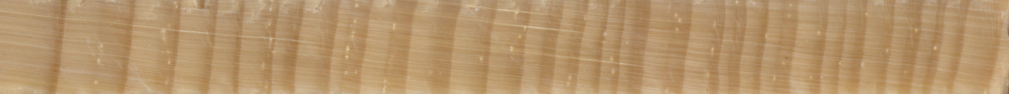}
   \includegraphics[width=0.3\textwidth]{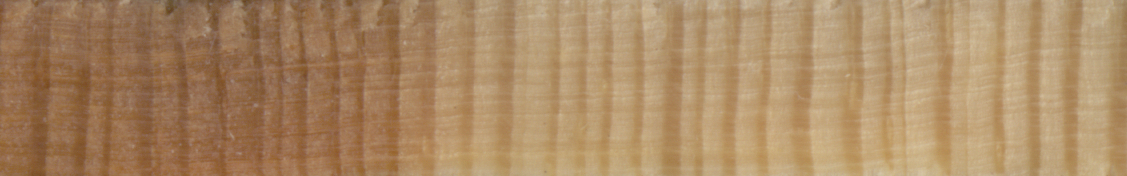}
   \includegraphics[width=0.3\textwidth]{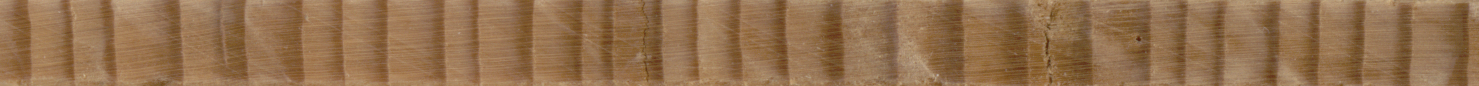}
   \caption{Examples of core tree-ring images taken from a dataset with 239 images \cite{FABIJANSKA2017279}.}
   \label{fig:coreimg}
\end{center}
\end{figure*}

\begin{figure*}
\begin{centering}
    \begin{subfigure}{0.3\textwidth}
    \includegraphics[width=\textwidth]{F02a}
    \label{fig:ddbb-F02a}
    \end{subfigure}
    \begin{subfigure}{0.3\textwidth}
    \includegraphics[width=\textwidth]{F02b}
    \label{fig:ddbb-F02b}
    \end{subfigure}
    \begin{subfigure}{0.3\textwidth}
    \includegraphics[width=\textwidth]{F02c}
    \label{fig:ddbb-F02c}
    \end{subfigure}
    \begin{subfigure}{0.3\textwidth}
   \includegraphics[width=\textwidth]{F02d}
    \label{fig:ddbb-F02d}
    \end{subfigure}
    \begin{subfigure}{0.3\textwidth}
   \includegraphics[width=\textwidth]{F02e}
    \label{fig:ddbb-F02e}
    \end{subfigure}
    \begin{subfigure}{0.3\textwidth}
   \includegraphics[width=\textwidth]{F03c}
    \label{fig:ddbb-F03c}
    \end{subfigure}
     \begin{subfigure}{0.3\textwidth}
   \includegraphics[width=\textwidth]{F07b}
    \label{fig:ddbb-F07b}
    \end{subfigure}
    \begin{subfigure}{0.3\textwidth}
   \includegraphics[width=\textwidth]{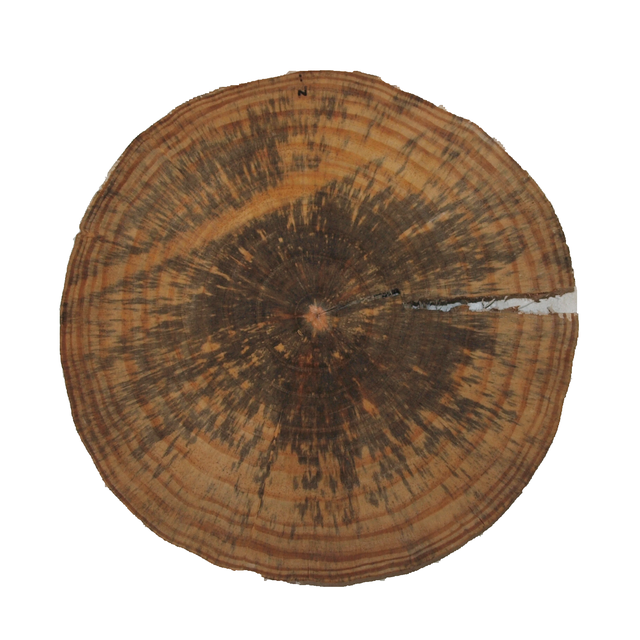}
    \label{fig:ddbb-L02b}
    \end{subfigure}
    \begin{subfigure}{0.3\textwidth}
   \includegraphics[width=\textwidth]{L03c}
    \label{fig:ddbb-L03c}
    \end{subfigure}
   \caption{Some examples of images from the UruDendro dataset. Note the variability of the images and the presence of fungus (image L02b), knots (images F07b and F03c), and cracks (images F02e and L03c). The first five images are from the same tree at different heights, as the text explains in Section \ref{sec:database}. }
   \label{fig:ddbb}
\end{centering}
\end{figure*}

Most of the available methods for dendrochronology (measurement and dating of tree growth rings) use images taken from cores (small cylinders crossing all the tree growth rings) instead of complete transverse cross sections. Figure \ref{fig:coreimg} illustrates that core images are rectangular sections. Using cores for the analysis presents some advantages. The rings are measured on a small portion of the trunk, which can be assumed as a sequence of bands with repetitive contrast, simplifying the image analysis.
However, this method provides limited information on annual tree growth because it results in a circular approximation based on a rectangular section. In many cases, ring growth is not uniform, leading to significant errors.
An application that needs the study of the whole cross-section is when studying the angular homogeneity of the ring-tree pattern trying to detect the so-called compression wood \cite{Duncker2014} for which the lack of homogeneity in the growing pattern produces differential mechanical properties. On the downside,
cross-section analysis implies the felling of the tree and includes the challenge of generating a pattern of (almost) concentric closed curves representing the tree rings. As Figure \ref{fig:ddbb} shows, several factors increase the difficulty of the task: wood knots, fungi appearing as black spots with shapes following radial directions, and cracks that can be very wide.

This article presents a founded method for automatically delineating tree rings on cross-section RGB images. The approach takes advantage of the knowledge of the tree cross-section's general structure and redundant information on a radial profile for different angles around the tree's pith. 

The method is evaluated over two public datasets with two different species and outperforms the other publicly available comparable method \cite{inbd} in accuracy.

\Cref{sec:antecedentes} briefly review previous work in the field. \Cref{sec:method} describes the proposed automatic cross-section tree-ring detection algorithm (CS-TRD). \Cref{sec:algorithm} presents in more detail the algorithms and \Cref{sec:implementation} discuss implementation details.
\Cref{sec:ExperimentsAndResults} briefly presents a dataset for developing and testing the proposed approach and discusses some experimental results. \Cref{sec:conclusions} concludes and propose future work.

\section{Previous work}
\label{sec:antecedentes}

Tree ring detection is an old and essential problem in forestry, and it has multiple uses in dendroecology, dendrochronology, forest management, and other applications. Due to the species idiosyncrasies, many practitioners still use a manual approach, using a ruler or other (manual) tree-ring measuring system, which is a tedious and time-consuming task that requires an expert operator.

Among the strategies proposed in recent years for the automation of tree ring delineation over wood cross-section images, we can mention the classic image-processing approaches that try to detect borders and then reconstruct the ring pattern \cite{CerdaHM07, Norell2007GreyWP, Zhou2012, Henke2014SemiautomaticTR, KennelBS15, JacobiSets} and the algorithms based on deep learning methods which try to learn the solution from the data \cite{inbd}. The core approaches are more extended, resulting in a significantly greater number of data sets. Hence, the deep learning methods are generally designed for core dendrochronology \cite{DeepDendro, Polek2022AutomationOT}. 

Cerda et al. \cite{CerdaHM07} proposed a classic image processing approach for detecting entire growth rings based on the Generalized Hough Transform. This work already suggests using the general geometrical structure of the tree rings, which we use in our approach, as illustrated in Figure \ref{fig:algo}. Norell et al. use the Grey Weighted Polar Distance Transform \cite{Norell2007GreyWP} to process end faces acquired in sawmill environments. Still, the method implies using rectangular sections, including the pith, and avoiding knots or other disturbances, diminishing the generality of the approach. Zhou proposes a much simpler method \cite{Zhou2012}, which resembles the traditional manual procedure in which two perpendicular lines across the slice are traced. The watershed method is applied to the profiles to obtain the peaks corresponding to each ring. Henkel et al. use an active contours approach \cite{Henke2014SemiautomaticTR} alone or coupled with a Dual-Tree Complex Wavelet Transform \cite{KennelBS15} based on the evolution of a partial differential equation that includes terms related to the image content and to the curve itself. Many PDE-based algorithms are time-consuming, making them difficult to use in real-time applications. Makela et al. proposed a method based on Jacobi Sets to detect the ring pattern and pith location \cite{JacobiSets}. All these methods rely on detecting the edges corresponding to the tree rings and different strategies to reconstruct the pattern. In all cases, the pith is the center of a general structure. Most of these works were tested against a few images (ranging from 7 to 20). Unfortunately, the code and used images are unavailable in these cases for testing and comparing the results with other methodologies.

Deep learning approaches have become more popular in recent years and have naturally been applied to this problem. Still, the scarcity of labeled data for a given species is a significant problem in the area, as the methods must be tailored to the particularities of each species. Gillert et al., \cite{inbd} proposed a method for cross-section tree-ring detection named Iterative Next Boundary Detection Network (INBD) but applied it to high-resolution images of shrub cross-sections specially conditioned for microscopy observation. In this case, the images are not only of a particular species, but also the microscopic resolution introduces specific characteristics. Starting from the pith, the method infers the annual ring at each iteration step, detecting ring by ring from the medulla to the tree's bark. A problem can arise if an intermediate ring is badly processed. In those cases, the error propagates and affects all the rings outward. The method was trained and tested on images obtained under the specific conditions of microscopic images of shrub species. This is the unique method for which we have access to the code, allowing us to compare it with our approach. In \Cref{sec:inbd}, we illustrate the results of applying their model to our datasets, training their method for our particular images. 

Besides the INBD method, most deep-learning-based approaches are applied to core images. For example, recently, Polek et al.\cite{Polek2022AutomationOT} applied a deep learning approach to process cores of coniferous species. Fabijańska et al. proposed both a classic image-processing approach \cite{FABIJANSKA2017279} (based on the linking of image gradient peak detected pixels) and a convolutional neural network one \cite{DeepDendro} for detecting tree rings over core images. Comparing both methods, they reported a precision of 43\% and a recall of 51\% for the classical approach, a precision of 97\%, and a recall of 96\% for the deep learning one. Without the code or the data, it is impossible to verify these claims with other species or datasets (their experiments concern three ring-porous wood species).

In short, deep learning-based methods are almost all for cores, and the absence of labeled databases makes it difficult to use them on complete slices. Classical methods are generally based on edge detection and build rings out of them. Most reported methods lack the code or the data to verify their claims. Still, their analysis leads to the assumption that the presence of perturbations such as fungus, cracks, and knots strongly affects the performance because the construction of each ring depends on the previous ones. Therefore, an error in the pith or rings close to the center propagates to the rest of the structure.

\section{Proposed Approach}

In this section, we present the main ideas of the method for tree-ring delineation over RGB cross-section images, called  CS-TRD for Cross-Section Tree-Ring Detector.
\label{sec:method}
\subsection{Assumptions}
Our tree-ring detection algorithm is heavily based on the structural characteristics of the problem:
\begin{itemize}
    \item The use of the whole horizontal cross-section of a tree (slice) instead of a wood dowel (or core), as most dendrochronology approaches do. 
    \item The following properties generally define the rings on a slice:
    \begin{enumerate}
        \item The rings are roughly concentric, even if their shape is irregular. This means that two rings can't cross.
        \item Several rays can be traced outwards from the slice pith. Those rays will cross each ring only once.
        \item We are interested only in the rings corresponding to the latewood to earlywood transitions, namely the \textit{annual rings}.
        \end{enumerate}
\end{itemize}

The principal idea of the method proposed in this work is the definition and use of a general structure formed by the rings, as explained in \Cref{sec:definitions}. 

\subsection{Definitions}
\label{sec:definitions}
\begin{figure*}
\begin{center}
   \begin{subfigure}{0.40\textwidth}
   \def\svgwidth{\linewidth}
   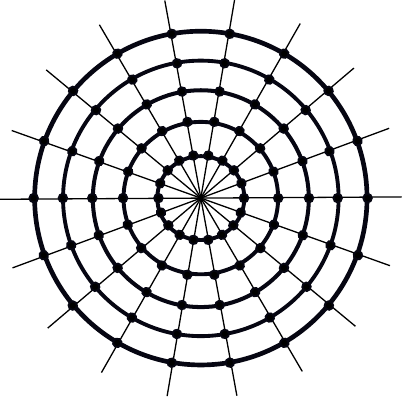
   \caption{}
   \label{fig:definitionsA}
   \end{subfigure}
   \hfill
   \begin{subfigure}{0.40\textwidth}
   \def\svgwidth{\linewidth}
\begingroup%
  \makeatletter%
  \providecommand\color[2][]{%
    \errmessage{(Inkscape) Color is used for the text in Inkscape, but the package 'color.sty' is not loaded}%
    \renewcommand\color[2][]{}%
  }%
  \providecommand\transparent[1]{%
    \errmessage{(Inkscape) Transparency is used (non-zero) for the text in Inkscape, but the package 'transparent.sty' is not loaded}%
    \renewcommand\transparent[1]{}%
  }%
  \providecommand\rotatebox[2]{#2}%
  \newcommand*\fsize{\dimexpr\f@size pt\relax}%
  \newcommand*\lineheight[1]{\fontsize{\fsize}{#1\fsize}\selectfont}%
  \ifx\svgwidth\undefined%
    \setlength{\unitlength}{424.76705933bp}%
    \ifx\svgscale\undefined%
      \relax%
    \else%
      \setlength{\unitlength}{\unitlength * \real{\svgscale}}%
    \fi%
  \else%
    \setlength{\unitlength}{\svgwidth}%
  \fi%
  \global\let\svgwidth\undefined%
  \global\let\svgscale\undefined%
  \makeatother%
  \begin{picture}(1,0.33375362)%
    \lineheight{1}%
    \setlength\tabcolsep{0pt}%
    \put(0,0){\includegraphics[width=\unitlength,page=1]{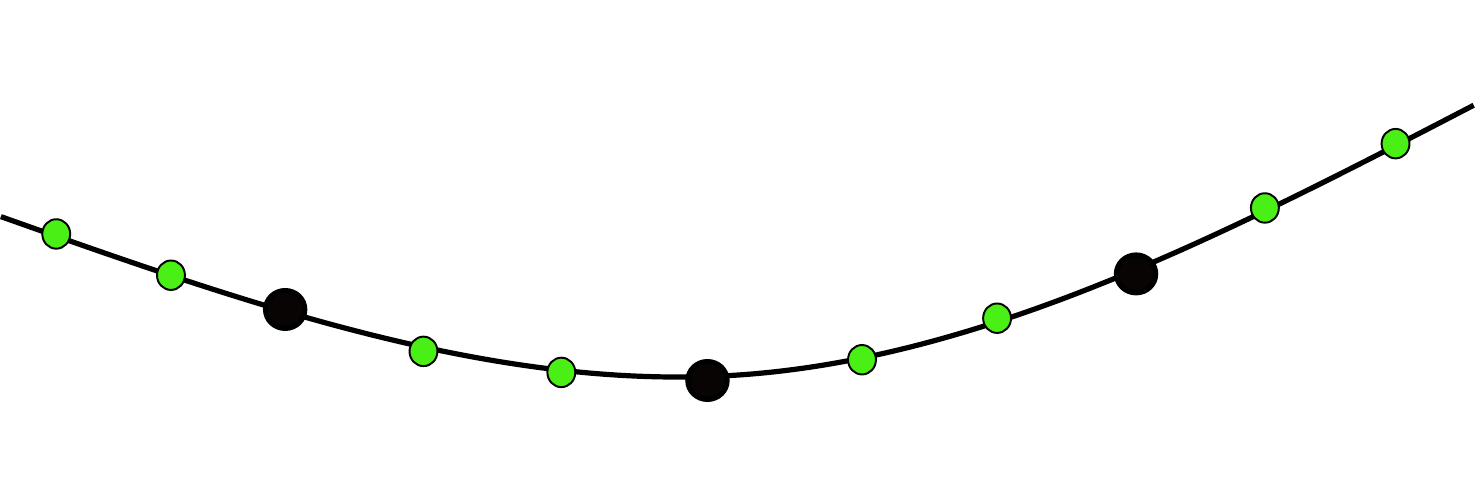}}%
    \put(0.21061995,0.13954937){\color[rgb]{0,0,0}\makebox(0,0)[lt]{\lineheight{1.25}\smash{\begin{tabular}[t]{l}$N_{i-1}$\end{tabular}}}}%
    \put(0.44415971,0.00643505){\color[rgb]{0,0,0}\makebox(0,0)[lt]{\lineheight{1.25}\smash{\begin{tabular}[t]{l}$p_{n}$\end{tabular}}}}%
    \put(0.4123015,0.11240774){\color[rgb]{0,0,0}\makebox(0,0)[lt]{\lineheight{1.25}\smash{\begin{tabular}[t]{l}$N_i$\end{tabular}}}}%
    \put(0.66370372,0.163992){\color[rgb]{0,0,0}\makebox(0,0)[lt]{\lineheight{1.25}\smash{\begin{tabular}[t]{l}$N_{i+1}$\end{tabular}}}}%
    \put(0.53704455,0.0470616){\color[rgb]{0,0,0}\makebox(0,0)[lt]{\lineheight{1.25}\smash{\begin{tabular}[t]{l}$p_{n+1}$\end{tabular}}}}%
    \put(0.32189371,0.04411435){\color[rgb]{0,0,0}\makebox(0,0)[lt]{\lineheight{1.25}\smash{\begin{tabular}[t]{l}$p_{n-1}$\end{tabular}}}}%
    \put(0,0){\includegraphics[width=\unitlength,page=2]{ChainPuntosYNodosGrande_Layer_1.pdf}}%
  \end{picture}%
\endgroup%

   \caption{}
   \label{fig:definitionsB}
   \vspace{5pt}
   \def\svgwidth{\linewidth}
   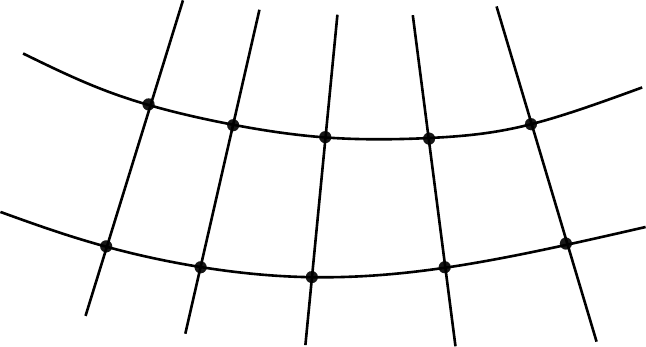
   \caption{}
   \hfill
   \label{fig:definitionsC}
   \end{subfigure}
   \caption{(a) The whole structure, called \textit{spider web}, is formed by a \textit{center} (which corresponds to the slice pith), $N_r$ \textit{rays} (in the drawing $N_r=18$) and the \textit{rings} (concentric curves). In the scheme, the \textit{rings} are circles, but in practice, they can be (strongly) deformed as long as they don't intersect another \textit{ring}. Each ray intersects a ring only once in a point called \textit{node}. (b) A curve is a set of connected \textit{points} (small green dots). Some of those \textit{points} are the intersection with \textit{rays}, named \textit{nodes} (black dots). A chain is a set of connected \textit{nodes}. In this case, the \textit{node} $N_i$ is the \textit{point} $p_n$. (c) Each \textit{Chain} $Ch_k$ and $Ch_{k+1}$, intersects the  \textit{rays} $r_{m-1}$, $r_{m}$ and $r_{m+1}$ in \textit{nodes} $N_{i-1}$, $N_{i}$ and $N_{i+1}$. 
   } 
   \label{fig:definitions}
\end{center}
\end{figure*}
 To explain the approach, we need some naming definitions; see Figure \ref{fig:definitions}. We call \textit{spider web} the global structure of the tree-rings we are searching for, which is depicted in a general way in  Figure \ref{fig:definitions}.a. It comprises a \textit{center}, associated with the slice's pith, which is the origin of a certain number of \textit{rays}. The \textit{rings} are concentric and closed curves that do not cross each other. Each \textit{ring} is formed by a \textit{curve} of connected points. Each \textit{ray} crosses a \textit{curve} only once. The \textit{rings} can be viewed as a \textit{curve} of points with \textit{nodes} in the intersection with the \textit{rays}. A \textit{chain} is a set of connected \textit{nodes}. As shown in Figure \ref{fig:definitions}.b, a \textit{curve}  is a set of chained nodes (small green dots in the figure, noted $P_i$). Depending on the position of the \textit{curve} concerning the \textit{center}, some of those \textit{points} are  \textit{nodes} (larger black dots in the figure, denoted $N_i$ hereafter). 
 The larger the number $Nr$ of \textit{rays}, the better the precision of the reconstruction of the \textit{rings}. We fix $Nr = 360$. Note that this is the ideal setting. In actual images,  \textit{rings} can disappear without forming a closed curve,  the \textit{rings} can be strongly deformed, etc. 
 
 Figure \ref{fig:definitions}.c illustrates the nomenclature used in this paper: \textit{Chains} $Ch_k$ and $Ch_{k+1}$, intersect the  \textit{rays} $r_{m-1}$, $r_{m}$ and $r_{m+1}$ in \textit{nodes} $N_{i-1}$, $N_{i}$ and $N_{i+1}$. To explain the method, we use this terminology. \textit{Chains} merge to form a \textit{ring}. The \textit{rays} determine a sampling of the \textit{curves}, producing \textit{chains}.

\subsection{Approach}

The method takes as input an image of the disk without background and the biological center of the disk (pith) (see~\Cref{fig:algo}.b) and returns the annual rings (see~\Cref{fig:algo}.i).~\Cref{fig:algo} illustrates the output of each stage of the method, which are fully described in~\Cref{sec:algorithm}.

Briefly, the method works as follows: from the center (disk pith), rays are traced (see~\Cref{fig:definitions}.a). The Canny edge detection method is then applied to the image (see~\Cref{fig:algo}.d), and by calculating the angle between the edge normals and the rays, most of the edges that do not belong to the latewood to earlywood transitions are eliminated. At the end of this filtering stage, we obtain both noisy edges and the edges of the rings we are looking for; additionally, some edges may be missing because they were not detected (see~\Cref{fig:algo}.e). Until now, the method is similar to the one proposed by Cerda et al. \cite{CerdaHM07}; the more challenging step is how the pixels belonging to the edges are grouped to form rings. 

In the next stage, the edges obtained in the previous phase enforce the \textit{spider web} structure, which describes the general properties of the annual rings. To achieve this, all edges are subsampled using the rays of \textit{spider web}: for each edge, we only retain the intersections between it and the respective rays (the nodes), forming what we call chains (see ~\Cref{fig:algo}.f). 

The final stage involves grouping all chains that belong to the same ring by imposing the SpiderWeb structure through a smoothness condition (see ~\Cref{fig:algo}.g, ~\Cref{fig:algo}.h and~\Cref{equ:similiraty_criterio}). The chain grouping is performed iteratively to connect chains near areas with stronger edge information. Once no more chains are connected in the current neighborhood, the method moves on to the next region with strong edge information. This iterative process, which prioritizes edge information along with the smoothness condition, is the main contribution of our method to the community. Now, we describe the method in detail.

\section{Algorithm}
\label{sec:algorithm}

\begin{figure*}
\begin{center}
   \begin{subfigure}{0.3\textwidth}
   \includegraphics[width=1\linewidth]{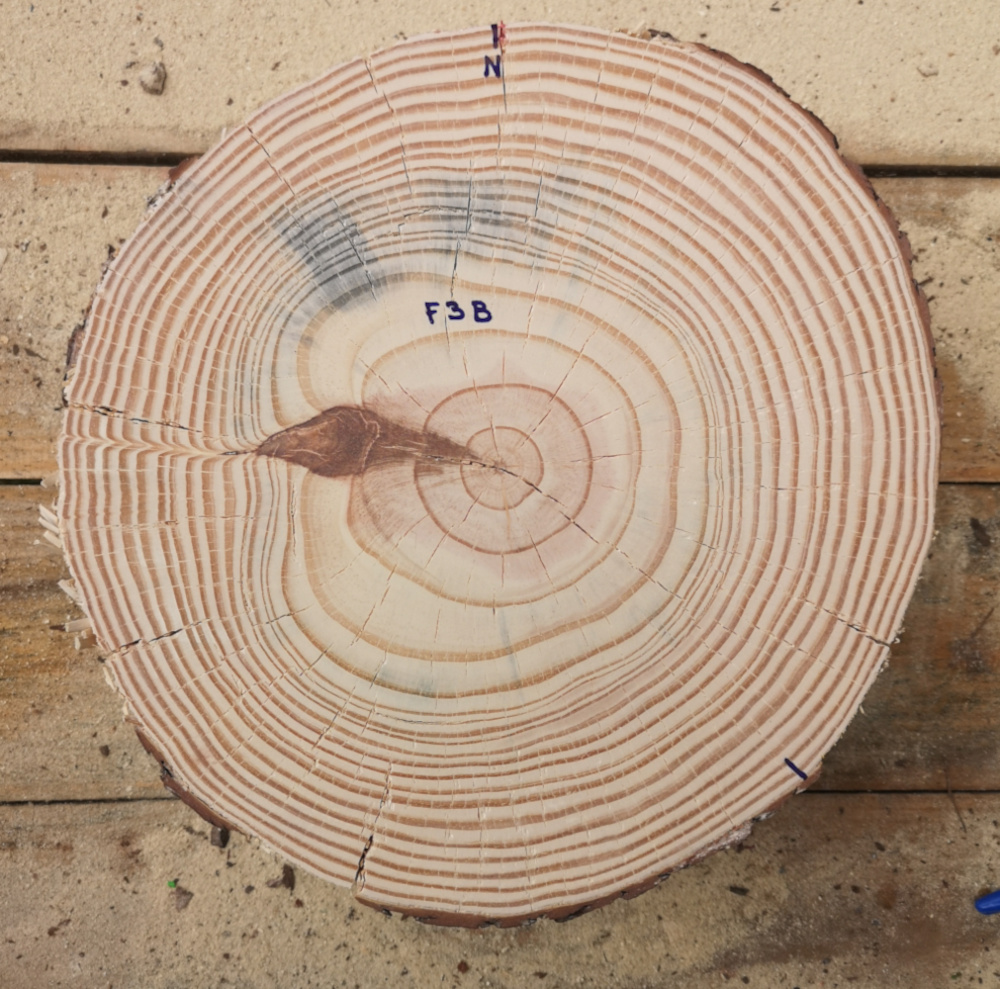}
   \caption{}
   \end{subfigure}   
   \begin{subfigure}{0.3\textwidth}
   \includegraphics[width=1\linewidth]{F03d_segmentation.jpg}
   \caption{}
   \end{subfigure}   
   \begin{subfigure}{0.3\textwidth}
   \includegraphics[width=1\linewidth]{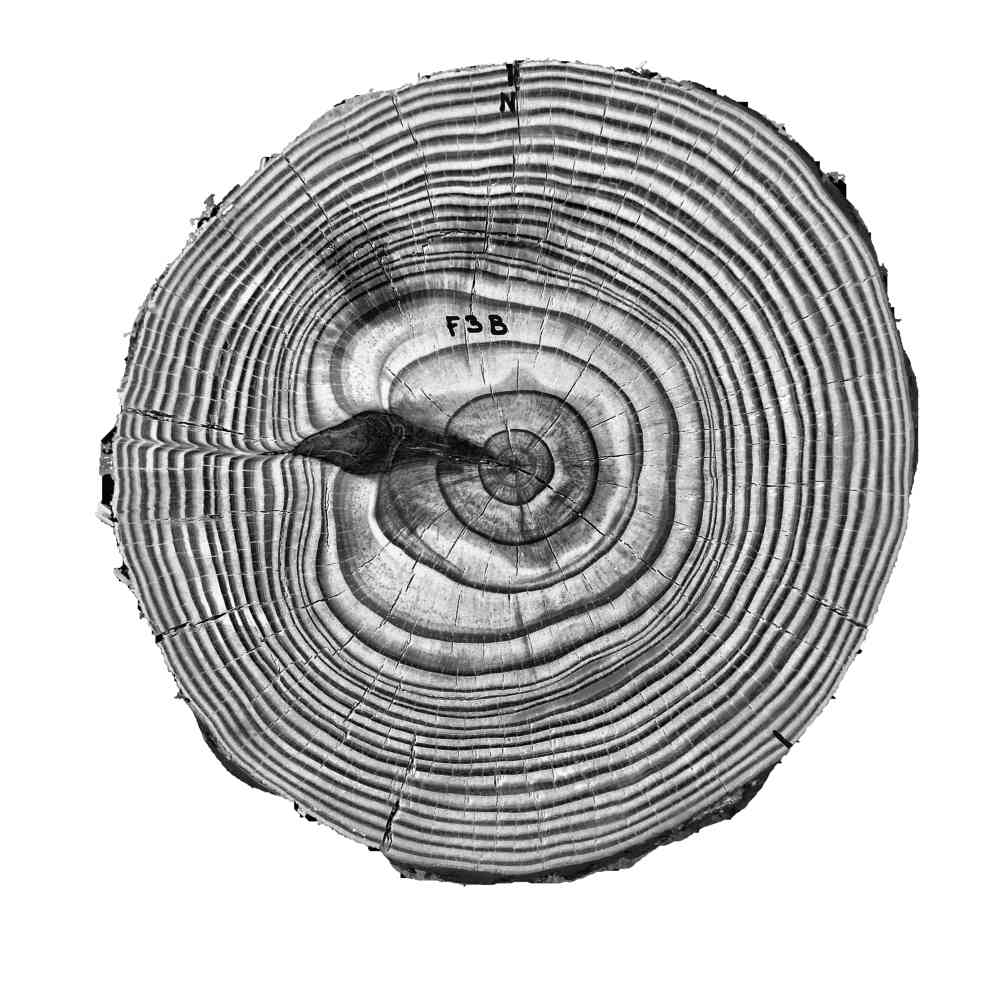}
   \caption{}
   \end{subfigure}   
   \begin{subfigure}{0.3\textwidth}
   \includegraphics[width=1\linewidth]{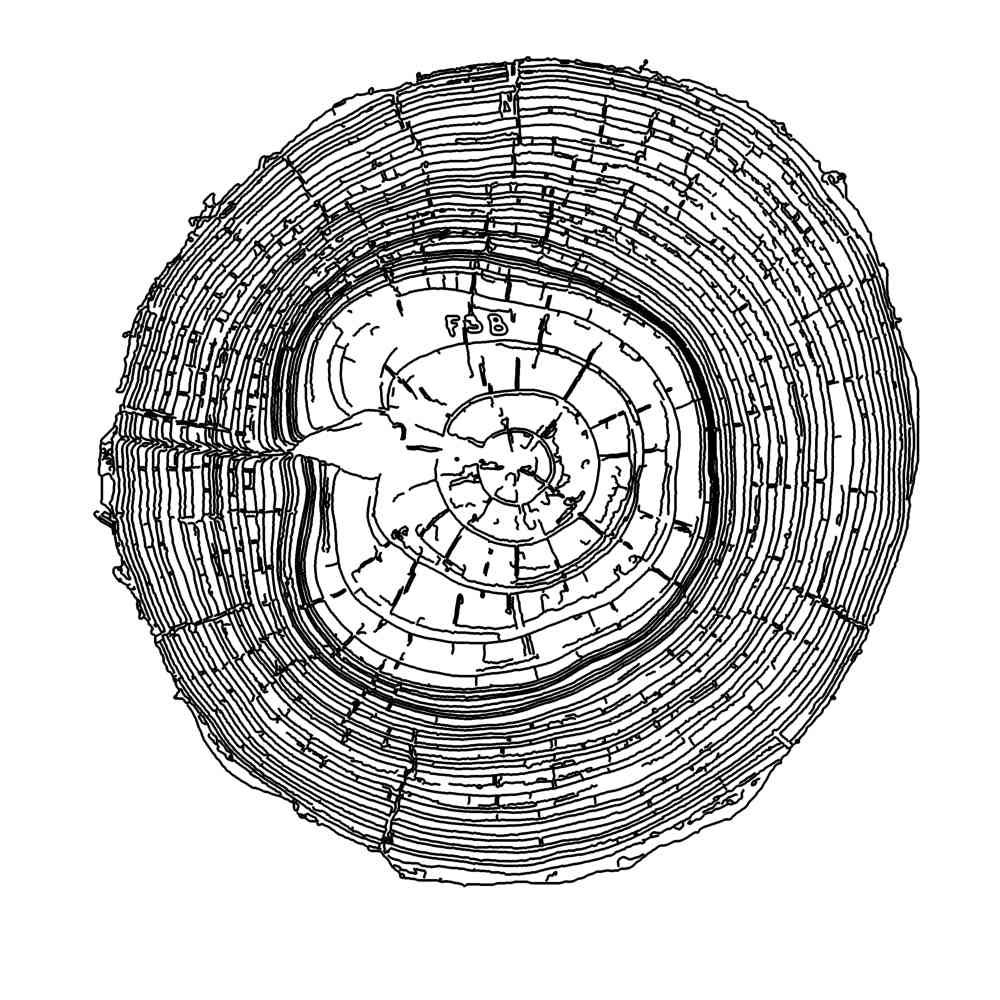}
   \caption{}
   \end{subfigure}   
   \begin{subfigure}{0.3\textwidth}
   \includegraphics[width=1\linewidth]{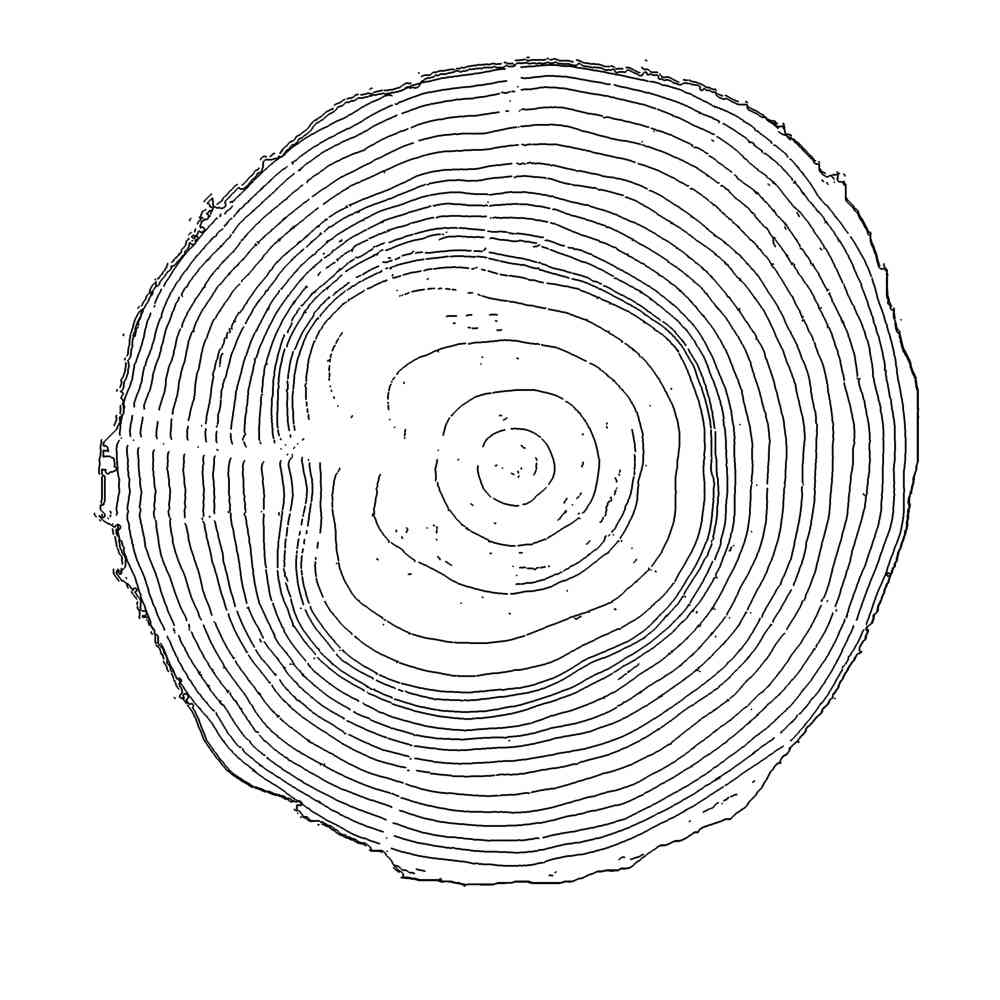}
   \caption{}
   \end{subfigure}   
   \begin{subfigure}{0.3\textwidth}
   \includegraphics[width=1\linewidth]{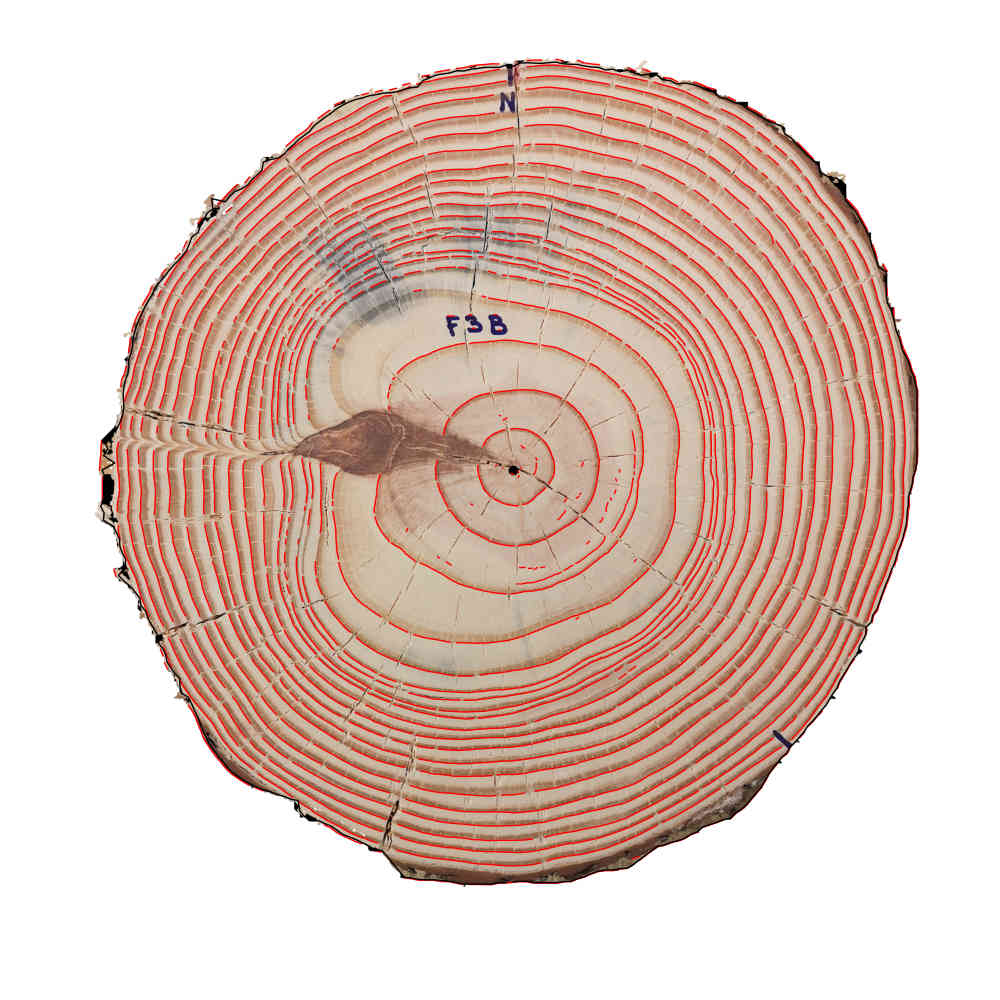}
   \caption{}
   \end{subfigure}   
   \begin{subfigure}{0.3\textwidth}
   \includegraphics[width=1\linewidth]{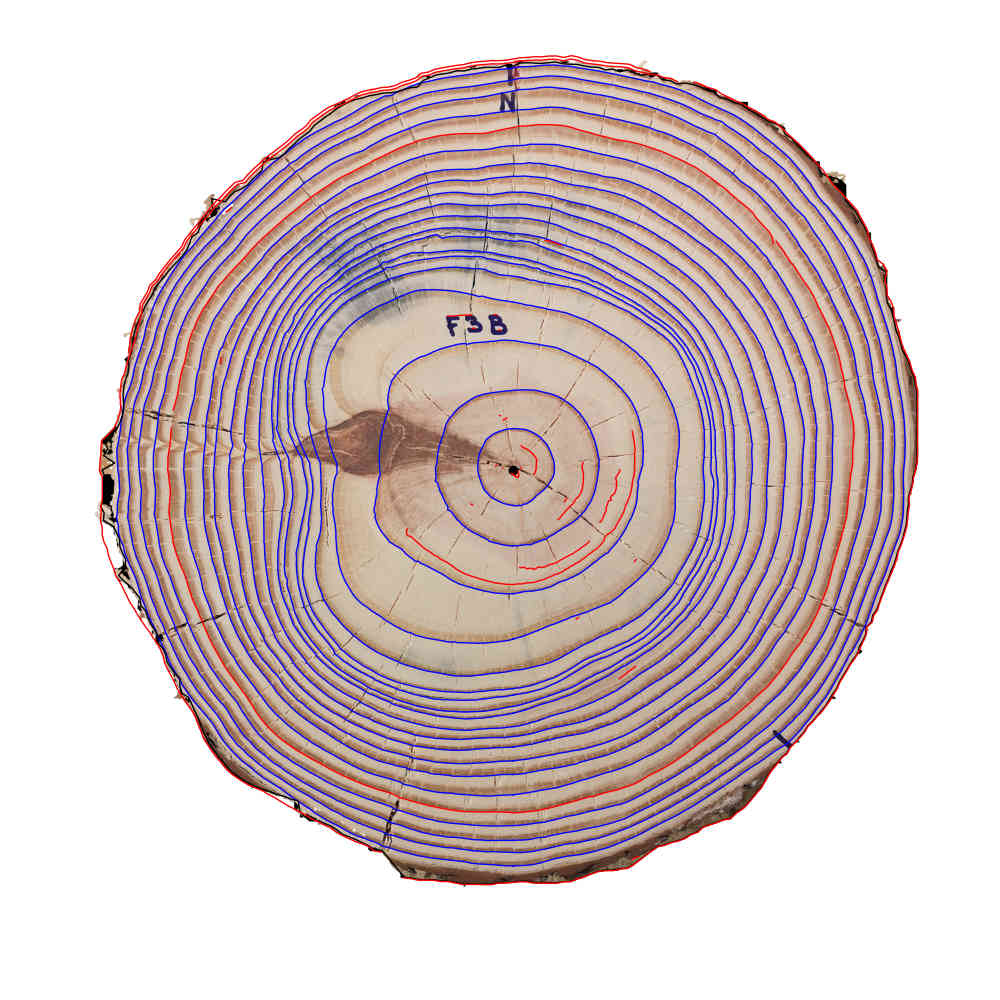}
   \caption{}
   \end{subfigure}   
   \begin{subfigure}{0.3\textwidth}
   \includegraphics[width=1\linewidth]{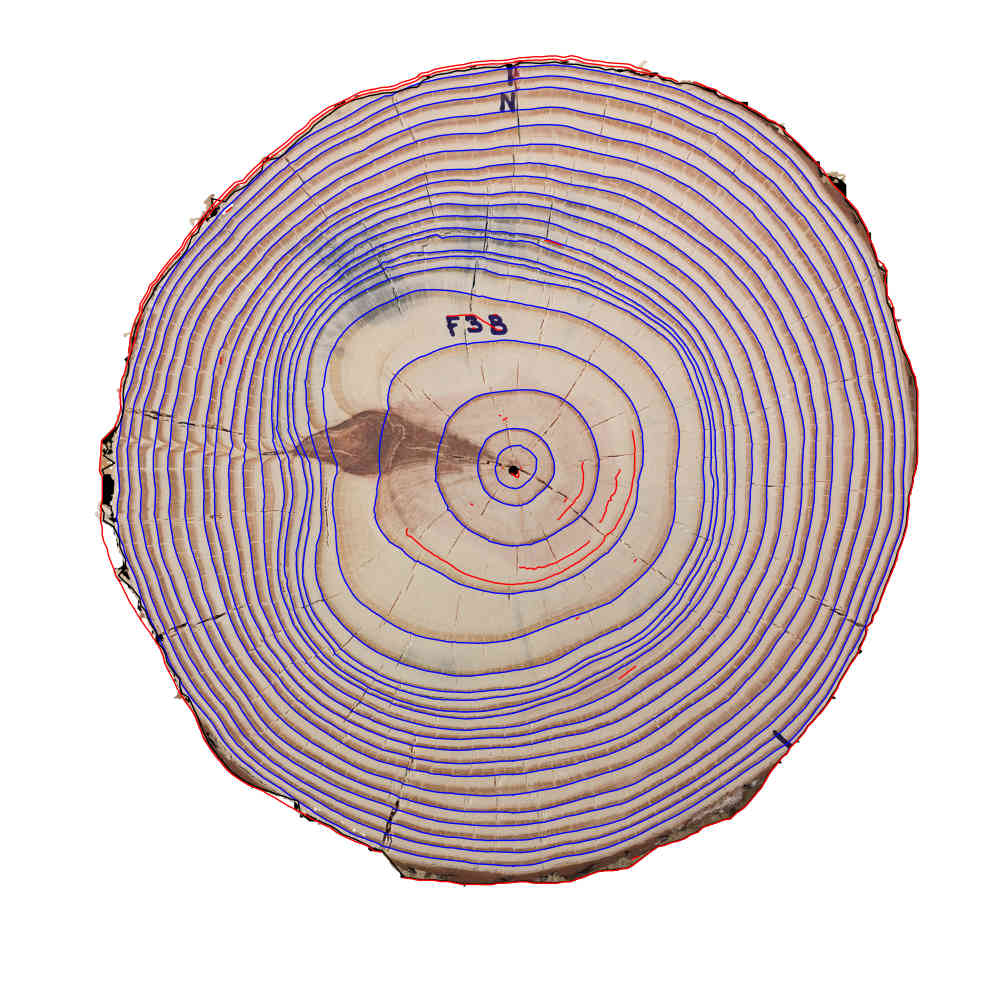}
   \caption{}
   \end{subfigure}   
   \begin{subfigure}{0.3\textwidth}
   \includegraphics[width=1\linewidth]{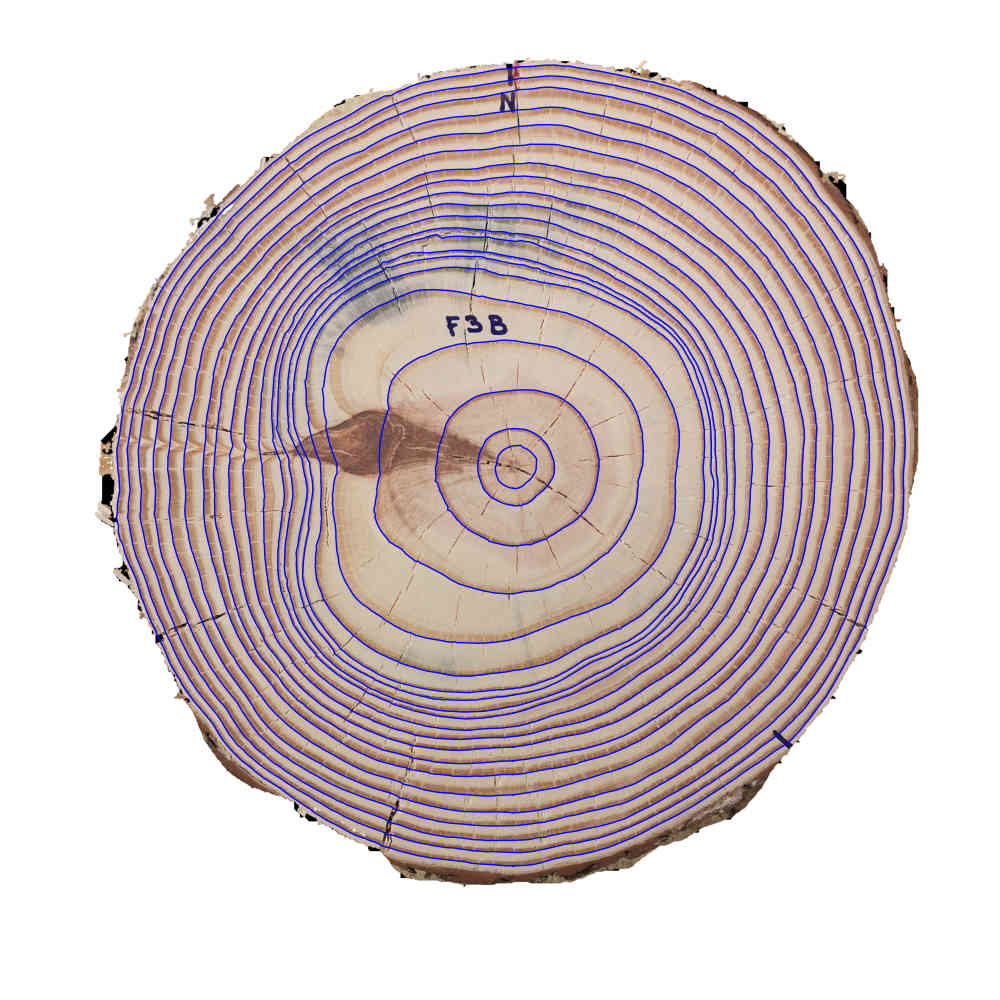}
   \caption{}
   \end{subfigure}   
      \caption{Principal steps of the CS-TRD algorithm: (a) Original image, (b) Background subtraction, (c) Pre-processed image (resized,  equalized, grayscale conversion), (d) Canny Devernay edge detector, (e) Edges filtered by the direction of the gradient, (f) Detected chains, (g) Connected chains, (h) Post-processed chains and (i) Detected tree-rings.}
   \label{fig:algo}
\end{center}
\end{figure*}

\Cref{algo:Globalalgo} depicts the CS-TRD method and \Cref{fig:algo} illustrates its intermediate results. The input is an image of a tree slice. First, we subtract the background, applying a deep learning-based approach \cite{salientObject} based on a two-level nested U-structure ($U^2Net$). 

Given the image without a background as shown in \Cref{fig:algo}.b, we must find the set of pixel chains representing the annual rings (dark to clear transitions). We need the center $c=(cy,cx)$ of the \textit{spider web} (the tree's pith) as input. This fundamental point can be manually marked or can be automatically detected \cite{ipolPith}. Here, we consider that this point is given (in the demo, both options are available).  Note that the values of parameters $\alpha$,  $N_{r}$, and $m_{c}$ are fixed once and for all.

\begin{algorithm}[!htbp]
  \KwIn{$Im_{in}$, // segmented input image. Background pixels are set to 255 (white)  \\ 
        $cy$, $cx$, // pith's coordinates, in pixels: center of the \textit{spider web}  \\
        $\sigma$, // Canny edge detector gaussian kernel parameter \\
        $height$, $width$, // dimensions of the image after the resize step \\
        $\alpha$, // threshold on the collinearity of the edge filtering \\
        $N_{r}$, // number of rays \\
        $m_{c}$ // minimum chain length \\
   }
  \KwOut{A json object $l\_rings$, each element is a closed \textit{chain} representing a tree-ring.}
  \SetAlgoLined
  \LinesNumbered
  \BlankLine 
    $Im_{pre}, cy, cx$ $\leftarrow$ preprocessing($Im_{in}, height, width, cy, cx$) \\
    $m\_ch_{e}, G_x, G_y$ $\leftarrow$ canny\_deverney\_edge\_detector($Im_{pre}$, $\sigma$) //  described in \cite{DevernayIPOL} \\
    $l\_ch_{f}$ $\leftarrow$ filter\_edges($m\_ch_{e}$,  $cy$, $cx$, $G_x$, $G_y$, $\alpha$, $Im_{pre}$)\\
    $l\_ch_{s},l\_nodes_s$ $\leftarrow$ sampling\_edges($l\_ch_{f}$, $cy$, $cx$, $N_r$, $m_c$, $Im_{pre}$)\\
    $l\_ch_{c},l\_nodes_c$ $\leftarrow$ connect\_chains($l\_ch_{s}$, $l\_nodes_s$, $cy$, $cx$) \\
    $l\_ch_p$ $\leftarrow$ postprocessing($l\_ch_{c}$, $l\_nodes_c$)\\
    $l\_rings$ $\leftarrow$ chain\_to\_labelme\_json($l\_ch_{p}$, $height$, $width$, $cy$, $cx$, $Im_{in}$) \\
    \KwRet{$l\_rings$}
\caption{Tree-ring detection algorithm}
\label{algo:Globalalgo}
\end{algorithm}

\subsection{Preprocessing} 
The first step in \Cref{algo:Globalalgo} is to preprocess the input image to increase the method's performance. We resize the image to a fixed $1500 \times 1500$ pixels size via Lanczos interpolation, then convert it to grayscale, and a histogram equalization is applied (\Cref{fig:algo}.c).

\subsection{Canny-Devernay edge detector} 
Line 2 of \Cref{algo:Globalalgo} corresponds to the edge detection stage. We apply the sub-pixel precision Canny Devernay edge detector \cite{Devernay95anon-maxima, DevernayIPOL}. The output of this step is a list of pixel chains corresponding to the image edges. Besides some noise-derived ones, we can group those edges into three  classes: 

\begin{itemize}
    \item $Edges_T$: produced by the tree growing process. It includes the edges that form the rings. Considering a pith outward direction, they can be of two types: those produced by early wood to late wood transitions (clear to dark) and latewood to early wood transitions (dark to clear). We are interested in detecting the former ones, hereon called annual rings.
    \item $Edges_R$: radial edges produced by cracks, fungi, or other phenomena.
    \item Other edges produced by wood knots and noise.
\end{itemize}

The gradient vector is normal to the edge and encodes the local direction and sense of the transition. The Canny Devernay filter produces the $G_x$ and $G_y$ matrices with  $x$ and $y$ gradient components, respectively, and $m\_ch_{e}$, the matrix of edge chains, in which successive rows list the coordinates $(x,y)$ of the chained pixels of a given edge, and the row [-1,-1] signals the end of an edge chain.

The Canny Devernay edge detector has three parameters: The standard deviation of the Gaussian kernel $\sigma$, and the gradient modulus thresholds $th_{low}$ and $th_{high}$, associated with the threshold with hysteresis filtering on the edge points. We only adjust the parameter $\sigma$; the other two are fixed (see \Cref{exp:size}). \Cref{fig:algo}.d show the output of this stage. We slightly modify the Canny Devernay filter IPOL implementation \cite{DevernayIPOL} to get the matrices $G_x$ and $G_y$ as output.

\subsection{Filtering the edge chains} 
Line 3 of  \Cref{algo:Globalalgo} corresponds to the edge filtering stage. Given the center $c=(cy,cx)$ and a point $p_i$ over an edge \textit{curve}, the angle $\delta(\vec{c p_i},\vec{G_{p_i}})$ between the vector $\vec{c p_i}$ 
and the gradient vector $\vec{G_{p_i}}$ at the the point  $p_i$  is given by:

 \begin{equation}
 \delta(\vec{c p_i},\vec{G_{p_i}}) =\arccos\left( \frac{\vec{c p_i} \times \vec{G_{p_i}}}{\|\vec{c p_i}\|\|\vec{G_{p_i}}\|}\right) \label{equ:filter_angle}   
 \end{equation}

We filter out all \textit{points} $p_i$ for which $
 \delta(\vec{c p_i},\vec{G_{p_i}})   \geq \alpha$. We fix  $\alpha = 30$ degrees. 
 
Two edge groups are filtered out: early wood transitions of the $Edges_T$ set gradients, which point inward, and $Edges_R$ set gradients, roughly normal to the \textit{rays}.  The algorithm produces the list $l\_ch_f$, including mainly the ring's edges and a disk border chain in the last position. \Cref{fig:algo}.e show the output of this stage.

\subsection{Sampling edges} Line 4 of  \Cref{algo:Globalalgo}, is the edge sampling stage which produces the \textit{nodes} depicted in \Cref{fig:definitions}.b. We sample each \textit{curve} of the $l\_ch_{f}$ list using the number of rays  $N_r$. The algorithm has two parameters: $N_r$ (360 by default) and $m_c$, the minimum number of nodes in a \textit{chain}. Every \textit{chain} has two endpoints, so we fix  $m_c=2$. The algorithm produces two lists: $l\_ch_{s}$ (of \textbf{Chain} objects)  and  $l\_nodes_{s}$ (of \textbf{Nodes} objects), which includes the nodes of all chains. These lists include two artificial \textit{chains}, one of type \textit{center} with $N_r$ nodes with the exact pith coordinates but different angular orientations, and one corresponding to the border. 
The inclusion of these artificial chains is beneficial for the connecting chain stage. The attribute $type$ of the object \textbf{Chain} identifies if it is an artificial or a standard chain.  

\Cref{fig:algo}.f shows this stage output. Standard chains are in red, and center and border chains are in black. Due to the sampling, there are fewer chains than edges. \Cref{fig:algo}.e has more (noisy) curves around the pith than \Cref{fig:algo}.f. Some small curves whose sampled length is shorter than $m_{c}$ are discarded. In that sense, this parameter filters out "short" chains.

\begin{figure}
    \centering
    \includegraphics[width=0.25\textwidth]{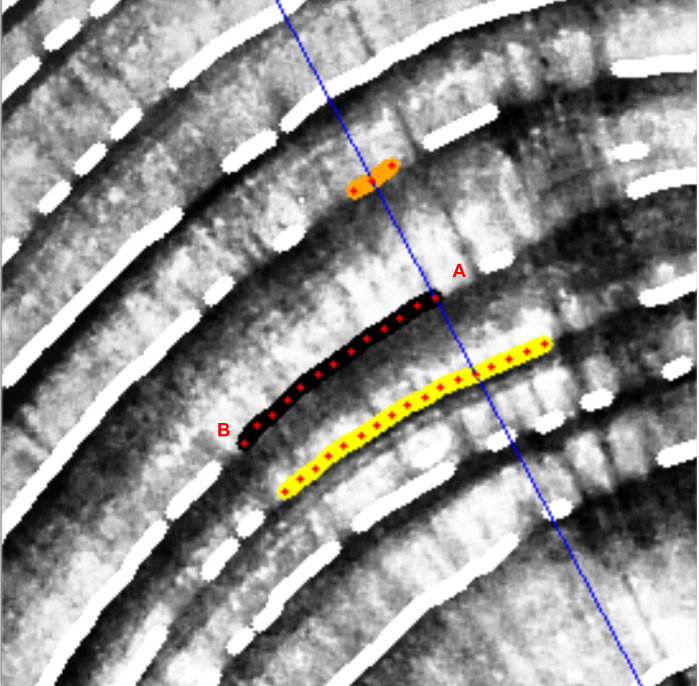}
    \caption{A given chain (in black) with two endpoints A and B. Its nodes (in red) appear at the intersection between the Canny Devernay curve and the rays. The ray at endpoint A is in blue. Other chains detected by Canny Devernay are in white. Endpoint A's inward and outward chains are in yellow and orange, respectively.}
    \label{fig:cproperties}
\end{figure}

Every chain has two endpoint nodes, A and B. Endpoint A is always the furthest node clockwise, while endpoint B is the most distant node counterclockwise. We use the concepts of \textit{outward} and \textit{inward} (concerning a given endpoint) in the chain attributes. Given the corresponding \textit{ray} of a \textit{chain} endpoint, we find the first \textit{chain} that intersects it going from the chain to the center along the \textit{ray} (named \textit{inward}) and the first \textit{chain} that intersects that \textit{ray} moving away from the center (named \textit{outward}), as shown in \Cref{fig:cproperties}. Chains are superposed over the gray-level image. The ray at endpoint A is blue, and the nodes are red and lie at the intersection between the rays and the edges. Orange and yellow chains are the \textit{visible} ones for the black chain at endpoint A (outward and inward, respectively).

We use three metric distances between chains. Given $EndPoint_{j}$ for the current chain $Ch_j$, and  $EndPoint_{k}$ for chain $Ch_k$, distances are defined as:

\begin{itemize}
\item \textbf{ Euclidean } defined as 
\begin{equation}
d_e = \sqrt{\left(x_j-x_k\right)^{2} + \left(y_j-y_k\right)^{2}}   \label{equ: Euclidean}
\end{equation}
Where $(x_j, y_j)$ and $(x_k, y_k)$ are the cartesian coordinates of $EndPoint_j$ and $EndPoint_k$.

\item \textbf{Radial Difference} defined as 
\begin{equation}
d_{rd} = \|r_j - r_k\|    \label{equ:radial}
\end{equation}
$r_j$ is the  Euclidean distance between $EndPoint_j$ and the pith center, and $r_k$ is the  Euclidean distance between $EndPoint_k$ and the pith center.

\item \textbf{Angular} Given the endpoints ray support angle, $\theta_j$ and $\theta_k$, this distance is defined as 
\begin{equation}
d_a =    \left(\theta_j-\theta_k+360\right) mod \hspace{0.2cm} 360 \label{equ:angular}
\end{equation}
Where $\theta_j$ and $\theta_k$ are the ray's direction supporting $EndPoint_j$ and $EndPoint_k$ (both in degrees), respectively, $mod$ refers to the module operation.  
\end{itemize}

\subsection{Connect chains}
\label{connect} 
We must now group these chains to form the rings (Line 5 of  \Cref{algo:Globalalgo}). As this section is one of the main parts of the algorithm, we will divide it into two subsections: one that explains the general idea of chain connection and another that details the algorithms themselves.

\subsubsection{General Logic of Chain Connection}
\label{sec:connect_general_logic}

The input of the algorithm is a set of chains. Some of these chains are spurious, produced by noise, small cracks, knots, etc., but most are part of the desired rings, as seen in Figure \ref{fig:algo}.f.

\begin{figure}
\begin{center}
   \begin{subfigure}{0.45\textwidth}
    \def\svgwidth{\linewidth}
    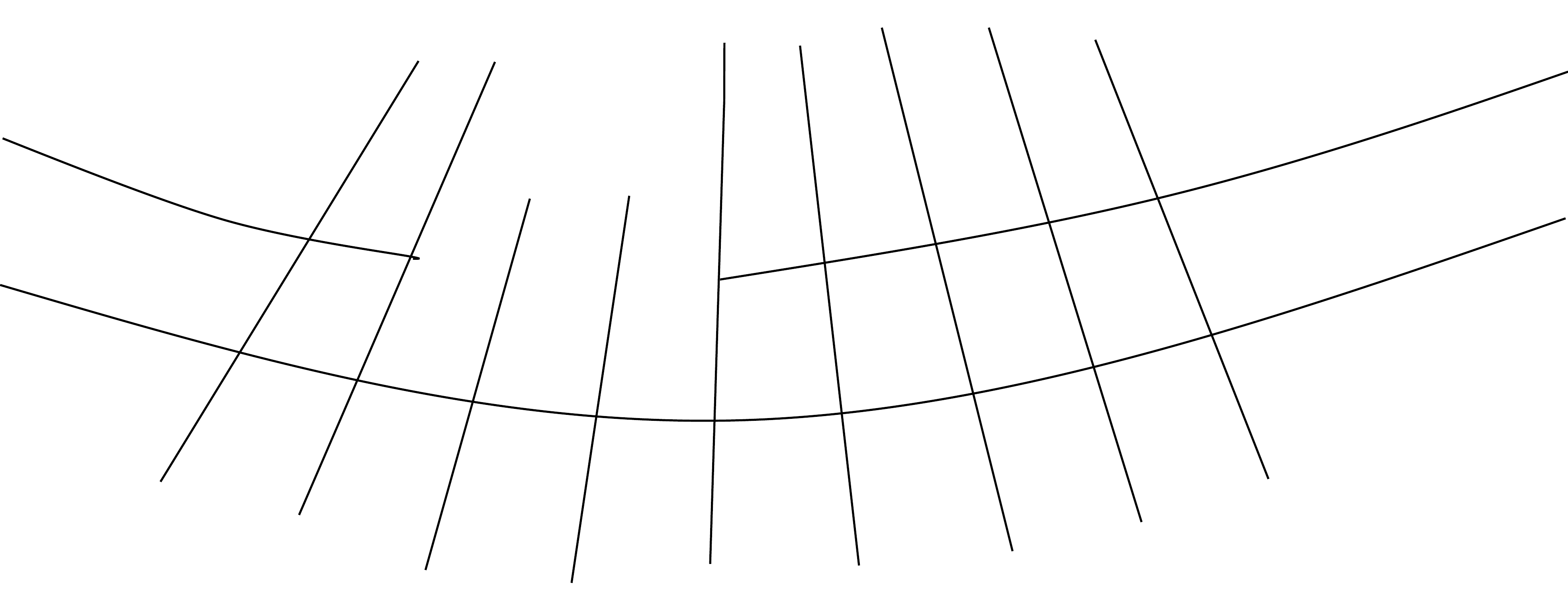
    \caption{} 
    \end{subfigure}
   \begin{subfigure}{0.45\textwidth}
    \def\svgwidth{\linewidth}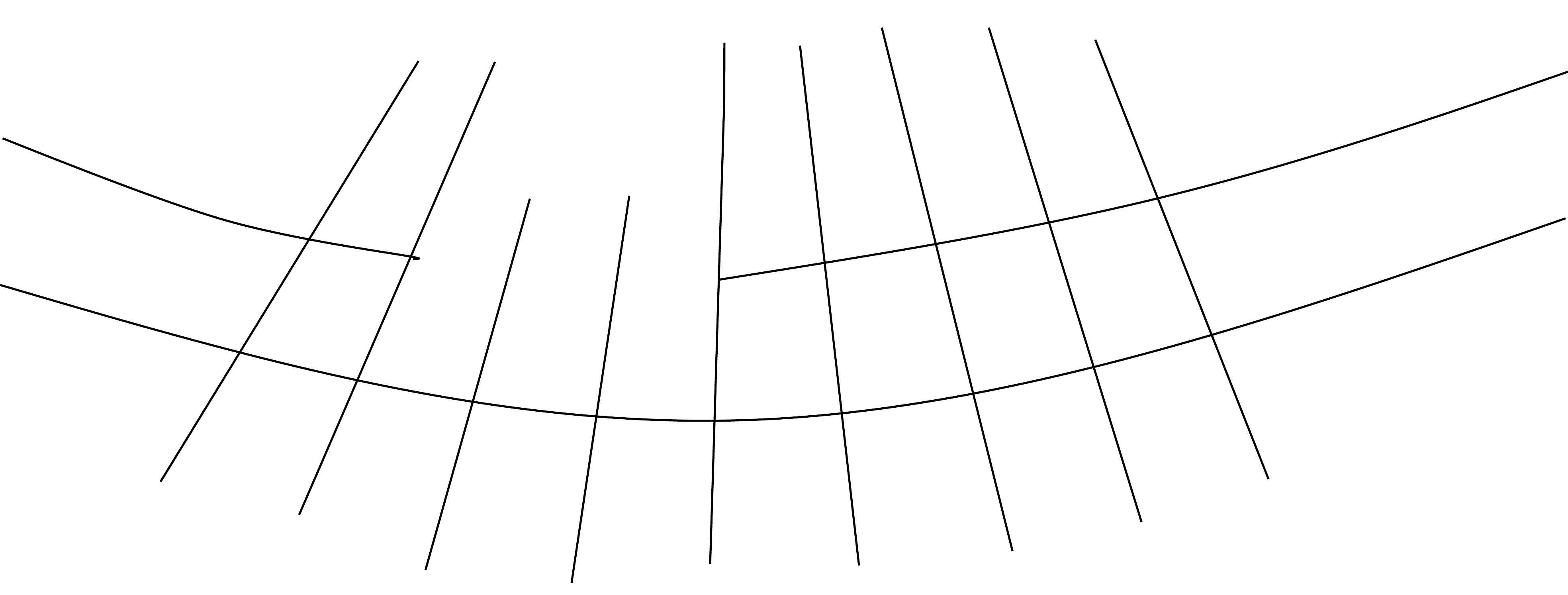
    \caption{} 
    \end{subfigure}
    \caption{An illustration of the \textit{connectivity} issue. (a) The question is if endpoint \textit{A} of $Ch_3$ must be connected to endpoint \textit{B} of $Ch_2$ (red dashed line) or to endpoint \textit{B} of $Ch_1$ (blue dashed line). In figure (b), the same question can be posed for the connection between endpoint B of $Ch_1$ and endpoint A of $Ch_2$. This connection is forbidden because $Ch_1$ and $Ch_2$ intersect (the endpoints are on the same \textit{ray}). Note that we represent the connections by line segments for clarity, but these are curves in the image space, as we interpolate between \textit{chain} endpoints in polar geometry.} 
   \label{fig:connectivityIssue}
\end{center}
\end{figure}

In general, several chains form a ring. To connect them, we must decide if the endpoints of two given chains can be connected, as illustrated by \Cref{fig:connectivityIssue}.a. We use a support chain ($Ch_0$ in the figure) to determine if those chains must be connected.

\begin{figure}[ht]
\begin{center}
   \begin{subfigure}{0.47\textwidth}
   \def\svgwidth{\linewidth}
\begingroup%
  \makeatletter%
  \providecommand\color[2][]{%
    \errmessage{(Inkscape) Color is used for the text in Inkscape, but the package 'color.sty' is not loaded}%
    \renewcommand\color[2][]{}%
  }%
  \providecommand\transparent[1]{%
    \errmessage{(Inkscape) Transparency is used (non-zero) for the text in Inkscape, but the package 'transparent.sty' is not loaded}%
    \renewcommand\transparent[1]{}%
  }%
  \providecommand\rotatebox[2]{#2}%
  \newcommand*\fsize{\dimexpr\f@size pt\relax}%
  \newcommand*\lineheight[1]{\fontsize{\fsize}{#1\fsize}\selectfont}%
  \ifx\svgwidth\undefined%
    \setlength{\unitlength}{1151.90439299bp}%
    \ifx\svgscale\undefined%
      \relax%
    \else%
      \setlength{\unitlength}{\unitlength * \real{\svgscale}}%
    \fi%
  \else%
    \setlength{\unitlength}{\svgwidth}%
  \fi%
  \global\let\svgwidth\undefined%
  \global\let\svgscale\undefined%
  \makeatother%
  \begin{picture}(1,0.31317875)%
    \lineheight{1}%
    \setlength\tabcolsep{0pt}%
    \put(0,0){\includegraphics[width=\unitlength,page=1]{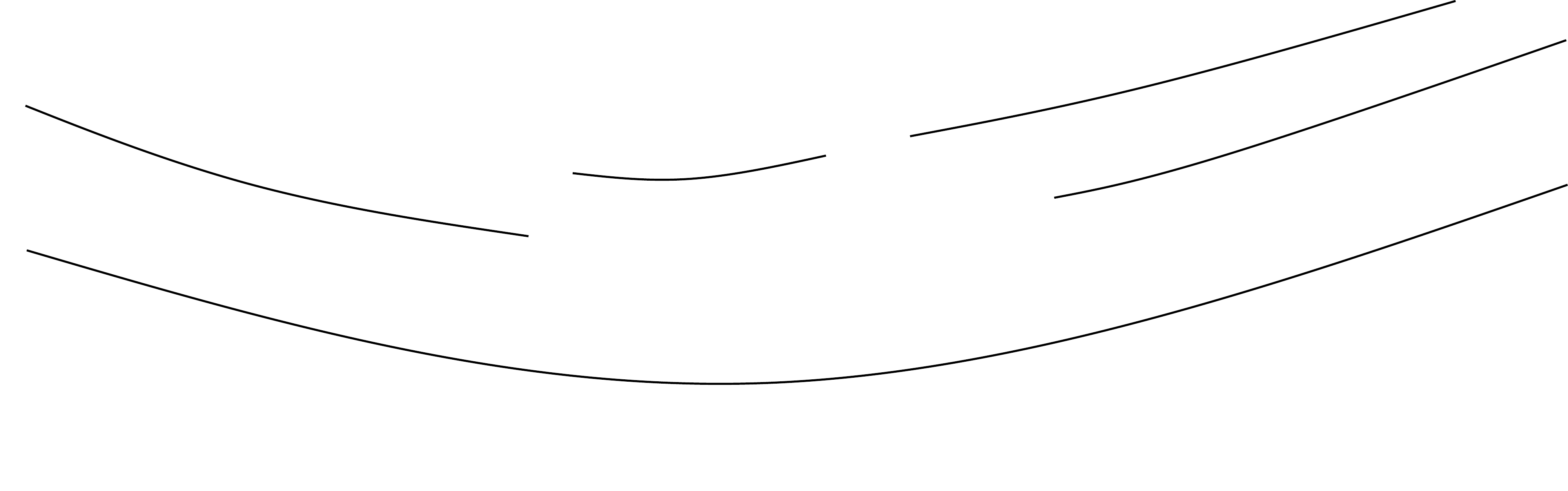}}%
    \put(0.83930899,0.1150164){\color[rgb]{0,0,0}\makebox(0,0)[lt]{\lineheight{1.25}\smash{\begin{tabular}[t]{l}$Ch_0$\end{tabular}}}}%
    \put(0.00544838,0.1873577){\color[rgb]{0,0,0}\makebox(0,0)[lt]{\lineheight{1.25}\smash{\begin{tabular}[t]{l}$Ch_1$\end{tabular}}}}%
    \put(0.39094554,0.22319857){\color[rgb]{0,0,0}\makebox(0,0)[lt]{\lineheight{1.25}\smash{\begin{tabular}[t]{l}$Ch_4$\end{tabular}}}}%
    \put(0.77542988,0.24741372){\color[rgb]{0,0,0}\makebox(0,0)[lt]{\lineheight{1.25}\smash{\begin{tabular}[t]{l}$Ch_6$\end{tabular}}}}%
    \put(0.62080298,0.25987261){\color[rgb]{0,0,0}\makebox(0,0)[lt]{\lineheight{1.25}\smash{\begin{tabular}[t]{l}$Ch_5$\end{tabular}}}}%
    \put(0,0){\includegraphics[width=\unitlength,page=2]{CincoChains.pdf}}%
    \put(0.190908,0.24840904){\color[rgb]{0,0,0}\makebox(0,0)[lt]{\lineheight{1.25}\smash{\begin{tabular}[t]{l}$Ch_3$\end{tabular}}}}%
    \put(0.10221908,0.148792){\color[rgb]{0,0,0}\makebox(0,0)[lt]{\lineheight{1.25}\smash{\begin{tabular}[t]{l}$Ch_2$\end{tabular}}}}%
  \end{picture}%
\endgroup%

    \caption{}
   \end{subfigure}
   \begin{subfigure}{0.47\textwidth}
   \def\svgwidth{\linewidth}
   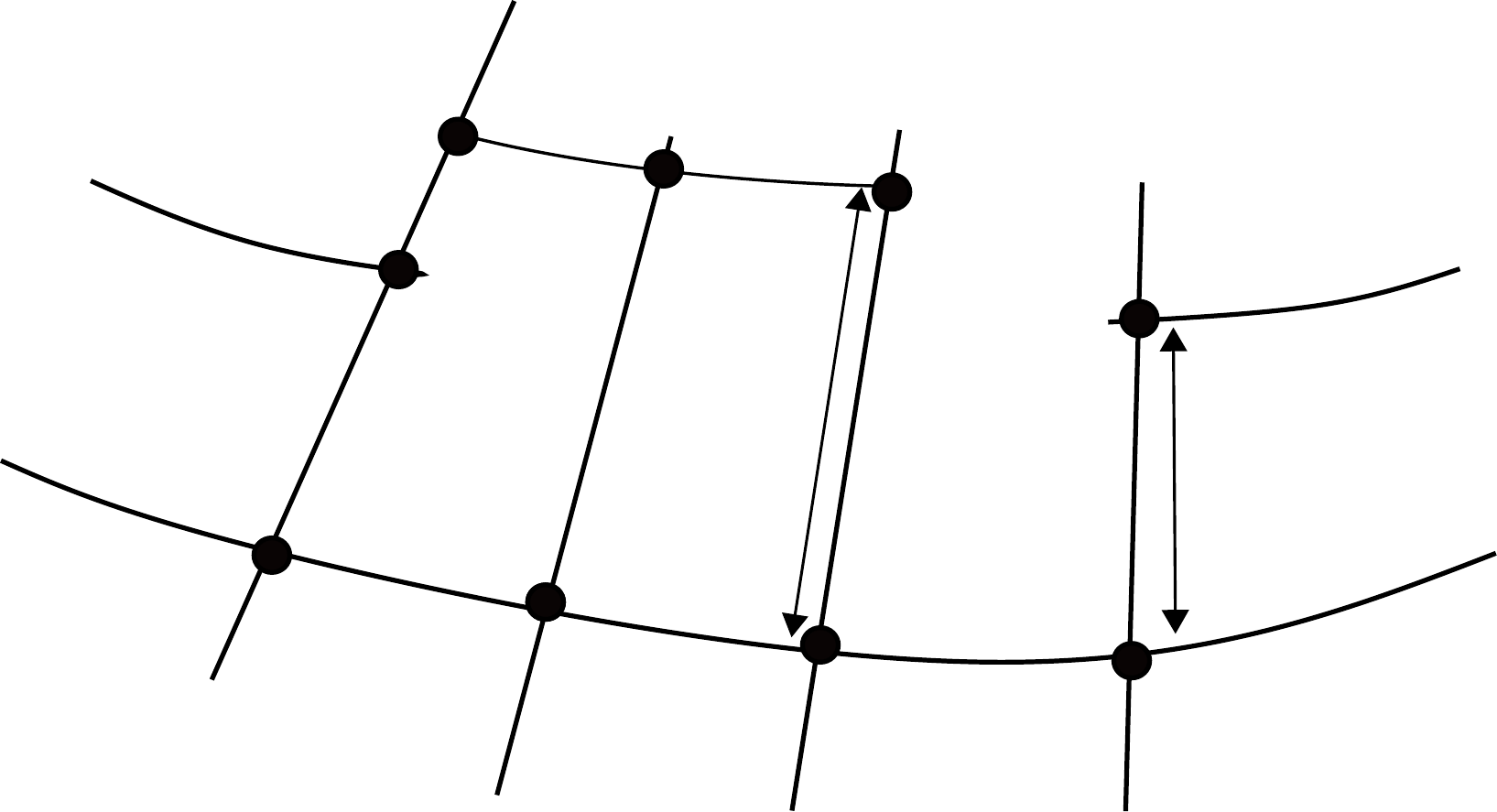
    \caption{}
   \end{subfigure}
\begin{subfigure}{0.47\textwidth}
   \def\svgwidth{\linewidth}
   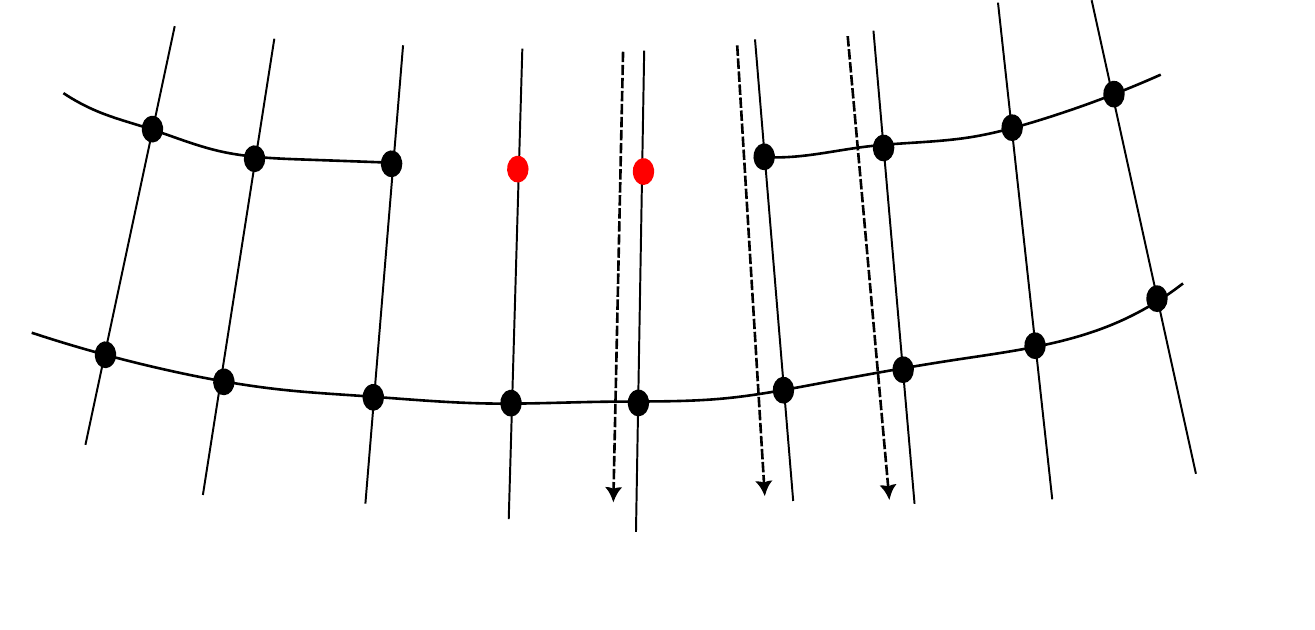
    \caption{}
\end{subfigure}
\caption{Connectivity nomenclature. (a) For the \textit{chain support} $Ch_0$, the set of  \textit{chain candidates} is formed by $Ch_1$, $Ch_2$, $Ch_4$, $Ch_5$ and $Ch_6$. \textit{Chain} $Ch_3$ is shadowed  by $Ch_1$ but $Ch_5$ is not shadowed by  $Ch_6$ because at least one endpoint of $Ch_5$ is visible from $Ch_0$. Note that a \textit{chain} becomes part of the \textit{candidate chains set} if at least one of its endpoints is visible from the \textit{chain support}. (b) Quantities used to measure the connectivity between \textit{chains}. $\delta R_i$ is the radial difference between two successive \textit{chains} along a \textit{ray} $R_i$ and $\delta N_i$ is the radial difference between two successive \textit{nodes} $N_i$ and $N_{i+1}$. Note that these nodes can be part of the same \textit{chain} or be part of two different \textit{chains} that may be merged. $Ch_i$ is the support chain. $Ch_i$ visible chains are $Ch_j$, $Ch_l$ and $Ch_k$. Chains $Ch_j$ and $Ch_k$ satisfy similarity conditions. (c) Chains $Ch_j$ and $Ch_k$ are candidates to be connected, and  $Ch_{i}$ is the support chain. $N_j^n$ ($N_k^n$) are the nodes of $Ch_j$ ($Ch_k$), with $n=0$ for the endpoint to be connected, and $r_s$ represents the radial distance to the center of $Node^s$. In red are the nodes created by an interpolation process between both endpoints.}.
\label{fig:ConnectivityNomenclature}
\end{center}
\end{figure}

To group chains that belong to the same ring, we proceed as follows:
\begin{enumerate}
    \item We order all the chains by length and begin by processing the longest, which we call \textit{Chain support}, $Ch_i$. Once we finish merging all the possible \textit{candidate chains} related to that one (the set $candidates_{Ch_i}$), we do the same with the next longest \textit{chain}, and so on. The reason for processing the longest chains first is that longer chains contain more edge information and more information about the tree ring structure, giving more robust results. Longer chains are more likely to be an edge belonging to an annual tree ring. Since consecutive rings have similar shapes, this iteration method propagates the shape of the longer chains.

        \item We find the chains that are visible from the \textit{Chain support} inwards (i.e., in the direction from \textit{Chain support} to the center).  
    Here, \textit{visible} means that a \textit{ray} that goes through one \textit{candidate chain} endpoint crosses the \textit{chain support} without crossing any other \textit{chains} in between. The set of \textit{candidate chains} of the \textit{Chain support} $Ch_i$ is named \textit{$candidates_{Ch_i}$}.  In \Cref{fig:ConnectivityNomenclature}.a this set is $candidates_{Ch_0} = \{Ch_1, Ch_2, Ch_4, Ch_5, Ch_6\} $.
    \textit{Chain} $Ch_3$ is shadowed  by $Ch_1$ and $Ch_5$ is not shadowed by $Ch_6$ because at least one of its endpoints are visible from $Ch_0$. The same process is made for the chains visible from the \textit{Chain support} outwards.

    \item We explore the set \textit{$candidates_{Ch_i}$} searching for  connections between them. By construction, the \textit{chain support} is not a candidate to be merged in this step. From the endpoint of a chain, we move forward angularly. The next endpoint of a nonintersecting \textit{chain} in \textit{$candidates_{Ch_i}$}  is a candidate to be connected to the first one. We say that two \textit{chains} intersect if there exists at least one \textit{ray} that crosses both \textit{chains}. For example, in Figure \ref{fig:ConnectivityNomenclature}.a, $Ch_6$  intersects with $Ch_5$ but not with $Ch_4$. 
    \label{connect:item_recorrer_cadenas}
    \item \label{connect:item_connectiviy_goodness} To decide if both chains must be connected, we must measure the \textit{connectivity goodness} between them, combining four criteria:
    \begin{enumerate}
        \item \textit{Radial tolerance for connecting chains}. The radial difference between the distance from each chain to be merged (measured at the endpoint to be connected) and the support chain must be small. For example, in Figure \ref{fig:ConnectivityNomenclature}.b, if we want to connect node $N_i$ of $Ch_l$ and node $N_{i+1}$ of $Ch_k$, we must  verify that $$ \delta R_i*(1-Th_{Radial\_tolerance})\leq \delta R_{i+1} \leq \delta R_i*(1+Th_{Radial\_tolerance})$$ Where $Th_{Radial\_tolerance}$ is a parameter of the algorithm. We call this condition \textit{RadialTol}.

        \item \textit{Similar radial distances of nodes in both chains}. For each chain, we define a set of nodes. For the chain $Ch_j$, this set is $N_j = \{ N_j^0, N_j^1,..., N_j^{n_{nodes}}\}$ where $n_{nodes}$ is the number of nodes to be considered,  fixed to $n_{nodes}=20$. See Figure \ref{fig:ConnectivityNomenclature}.c. We use the whole chain if it is shorter than $n_{nodes}$. We measure $\delta R_i$, the radial distance between each node in the given chain and the corresponding node (same ray) in the support chain, as illustrated in Figure \ref{fig:ConnectivityNomenclature}.b. This defines a set for each considered chain $j$ and $k$:
        $Set_j = \{ \delta R_j^0,..,\delta R_j^{n_{nodes}}\}$ and  $Set_k = \{ \delta R_k^0,..,\delta R_k^{n_{nodes}}\}$. We calculate its mean  and standard deviation $Set_j(\mu_j,\sigma_j)$ and $Set_k(\mu_k,\sigma_k)$. 
        This defines a range of radial distances associated with each chain: $Range_j = (\mu_j -    \textit{$Th_{Distribution\_size}$}*\sigma_j, \mu_j +  \textit{$Th_{Distribution\_size}$}*\sigma_j )$ and $Range_k = (\mu_k -   \textit{$Th_{Distribution\_size}$}*\sigma_k, \mu_k +   \textit{$Th_{Distribution\_size}$}*\sigma_k )$, where \textit{$Th_{Distribution\_size}$} is a parameter. There must be a non-null intersection between both distributions to connect both chains: $Range_j \cap Range_k \neq 0$. We call this condition \textit{SimilarRadialDist}.
        
        \item \textit{Regularity of the derivative}. Given two chains $Ch_j$ and $Ch_k$ which can be connected, let's call $Ch_{jk}$ the set of interpolated nodes between them (\Cref{fig:ConnectivityNomenclature}.c). The new virtual chain created by the connection between $Ch_j$ and $Ch_k$ will encompass the nodes of those two chains and $Ch_{jk}$. The parameter \textit{derivFromCenter} controls how are estimated the interpolated nodes between two chains, as the ones in red in   \Cref{fig:ConnectivityNomenclature}.c. If $derivFromCenter=1$, ray angle and radial distance from the center are used to estimate the position of the interpolated nodes. If its value is 0, the estimation is made by measuring the radial distance to the support chain. To test the regularity of the derivative, we define a set of nodes for each concerned chain. For the chain $Ch_j$, this set is $\{ N_j^0, N_j^1,..., N_j^{n_{nodes}}\}$ where $n_{nodes} = 20$ is the number of nodes to be considered. If the chain is shorter, we use all nodes. We compute the centered derivative in each node for all chains,  $\delta N^s = \frac{\|r_{s+1} - r_{s-1}\|}{2}$, where $r_s$ is the radial distance of the node $N^s$ to the center (i.e., the Euclidean distance between the node and the center). The set of the existing chains nodes is $Der(Ch_j, Ch_k) = \{\delta N_j^0, ..., \delta N_j^{n_{nodes}}, \delta N_k^0, ..., \delta N_k^{n_{nodes}} \} $. The condition \textit{RegularDeriv} is asserted if the greatest derivative in the interpolated chain is less or equal to the greater  derivative in the neighboring chains, with  tolerance $Th_{Regular\_derivative}$: 
        $$
        max(Der(Ch_{jk})) \leq max ( Der(Ch_j,Ch_k) ) \times Th_{Regular\_derivative} 
        $$

        \item \textit{Non-Overlapping Chain}. Finally, no other chain must exist between the chains to be connected. If another chain exists in between, it must be connected to the closer one. For example, in Figure \ref{fig:ConnectivityNomenclature}.a, it is impossible to connect chains $Ch_3$ and $Ch_5$ because between them appear $Ch_4$. We call this condition \textit{ExistChainOverlapping}. 
    \end{enumerate}
    Summarizing, in order to connect chains $Ch_j$ and $Ch_k$, the following condition must be met:
    
    \begin{equation}
            \mathbf{not}\textit{ExistChainOverlapping} \land \textit{RegularDeriv} \land ( \textit{SimilarRadialDist} \lor \textit{RadialTol} ) 
            \label{equ:similiraty_criterio}
    \end{equation}
    
    where $\lor$ and $\land$ stands for the logical \textit{or} and \textit{and} symbols, respectively, and the symbol $\mathbf{not}$ stands for the \textit{not} operator.
    
    The method iterates, searching for connectivity between chains over different neighborhood sizes. The parameter \textit{NeighbourdhoodSize} defines the maximum allowed distance, measured in degrees, for connecting two chains. 

    We iterate this process for the whole image nine times. In the first iteration, there are a lot of small chains, but in the second and third iterations, the concerned chains are already more extended and less noisy. As the merging process advances, we can relax the parameters to connect more robust chains. Table \ref{tab:connectivityParameters} summarizes the parameters used in each iteration.

    \begin{table}[!htbp]
    \centering
    \begin{tabular}{|l|l|l|l|l|l|l|l|l|l|}
    \hline
                       & 1   & 2   & 3   & 4   & 5   & 6   & 7   & 8   & 9   \\ \hline
    \textit{$Th_{Radial\_tolerance}$}        & 0.1 & 0.2 & 0.1 & 0.2 & 0.1 & 0.2 & 0.1 & 0.2 & 0.2 \\ \hline
    \textit{$Th_{Distribution\_size}$} & 2   & 2   & 3   & 3   & 3   & 3   & 2   & 3   & 3   \\ \hline
    \textit{$Th_{Regular\_derivative}$}      & 1.5 & 1.5 & 1.5 & 1.5 & 1.5 & 1.5 & 2   & 2   & 2   \\ \hline
    \textit{NeighbourdhoodSize} & 10  & 10  & 22  & 22  & 45  & 45  & 22  & 45  & 45  \\ \hline
    \textit{derivFromCenter }   & 0   & 0   & 0   & 0   & 0   & 0   & 1   & 1   & 1   \\ \hline
    \end{tabular}
    \caption{Connectivity Parameters. Each column is the parameter set used on that iteration.}
    \label{tab:connectivityParameters}
    \end{table}

    \item We proceed in the same manner in the outward direction.
\end{enumerate}

\subsubsection{Detailed Description of the Algorithms}

\Cref{algo:connectmainlogic} depicts the chain connection step's logic, described in the former section. 
It is iteratively called from an external function (Line 5 of \Cref{algo:Globalalgo}) with the parameter values defined in  \Cref{tab:connectivityParameters} columns. Chains produced at one iteration become the input chains for the next one.

As was mentioned, the logic for connecting chains is based on iterating over the \textit{support chain}. For each one, the set of visible chains is selected. The logic for selecting the support chain in each iteration (line 1~\Cref{algo:connectmainlogic}) is encapsulated in the \textbf{SystemStatus} object, which is the hub of our \textit{system}, containing all the necessary information to operate. The \textit{system} comprises all the chains and nodes and the intersection matrix $M$. This binary square matrix stores the intersection information of chains at each iteration. Rows and columns of $M$ span the chain list. Chain $Ch_j$ intersect chain $Ch_k$ if $M[j,k] = 1$. We say that two chains intersect if at least one ray crosses both chains. The intersection matrix, $M$, is an input argument for \Cref{algo:connectmainlogic}. The \textbf{SystemStatus} object updates the chains and nodes lists and the matrix $M$ whenever two chains are connected. This update is critical for our operation and signifies that the \textit{system} has been modified.

\Cref{algo:connectmainlogic} has a main loop from lines 2 to 19. Each iteration, the \textbf{SystemStatus} object produces the support chain ($Ch_i$). Support chains are iterated by following a neighborhood logic instead of sequentially going over the list to speed up the process.  Given a support chain, a secondary loop (lines 5 to 18) iterates over the inward (or outward) visible chains, as described in \Cref{fig:ConnectivityNomenclature}.a. In a third inner loop (lines 11 to 18), each candidate chain ($Ch_j$) is examined to find the closest chain ($Ch_k$) that meets the similarity goodness criteria (\Cref{equ:similiraty_criterio}). In summary, we search for the nearest chains to endpoints \textit{$Ch_{k}^a$} and \textit{$Ch_{k}^b$}. Between these candidate chains, we select the closest one ($Ch_k$) that does not intersect with the original chain $Ch_j$. Then, in line 17, chains $Ch_j$ and $Ch_k$ are connected. Connecting two chains implies generating new nodes between them (the new chain cannot have "holes"). Once no more chains can be connected, the main loop (lines 2 to 19) exits. In line 20, we iterate over all the chains in $l\_ch_s$, and if one has enough nodes, we complete it. The output is the lists of chains ($l\_ch_c$), nodes ($l\_nodes_c$), and the intersection matrix, $M$.

\begin{algorithm}[!htbp]
\KwIn{
$l\_ch_{s}$, // chains list \\
$l\_nodes_s$, // nodes list \\
$M$, //Binary Matrix  with intersection chains info \\
$cy$, $cx$, // pith's coordinates, in pixels: center of the \textit{spider web}  \\
$N_r$, // total number of rays\\
Parameters: \\
$th\_radial\_tolerance$, // Radial tolerance for connecting chains \\
$th\_distribution\_size$, // Chains Radial Difference Standard deviations for connecting chains \\
$th\_regular\_derivative$, // Chains Radial Derivative threshold for connecting chains \\
$neighbourhood\_size$, //Maximum angular distance allowed for connecting chains \\
$derivative\_from\_center$ // Define how nodes are interpolated \\
}
\KwOut{A list $l\_ch_{c}$, where each element is  a \textit{chain}, \\
$l\_nodes_{c}$,  where each element is a \textit{Node},\\
$M$, intersection matrix updated}
$state$ $\leftarrow$  $SystemStatus$($l\_ch_{s}$, $l\_nodes_s$, $M$ , $cy$, $cx$, $N_r$,$th\_radial\_tolerance$, $th\_distribution\_size$, $th\_regular\_derivative$, $neighbourhood\_size$, $derivative\_from\_center$)\\
\While{$state$.continue\_in\_loop()}{
    $Ch_i$ $\leftarrow$ $state$.get\_next\_chain()     \\
    $l\_s\_outward$, $l\_s\_inward$ $\leftarrow$ get\_chains\_in\_and\_out\_wards($l\_ch_{s}$,$Ch_i$)\\
    \For{$l\_candidates\_{Ch_i}$ in ($l\_s\_outward$, $l\_s\_inward$)}{
        $j\_pointer$ $\leftarrow$ 0\\
        \If{$l\_candidates\_{Ch_i}$ =  $l\_s\_inward$}{
             $location$ $\leftarrow$ "inward"
        }
        \Else{
             $location$ $\leftarrow$ "outward"
        }

        \While{length($l\_candidates\_{Ch_i}$) $>$ $j\_pointer$}{
             $Ch_j$ $\leftarrow$ $l\_candidates\_{Ch_i}$[$j\_pointer$]\\
             $l\_no\_intersection\_j$ $\leftarrow$ get\_non\_intersection\_chains($M$, $l\_candidates\_{Ch_i}$, $Ch_j$)\\
             $Ch_k^b$ $\leftarrow$ get\_closest\_chain\_logic($state$, $l\_candidates\_{Ch_i}$, $Ch_j, l\_no\_intersection\_j, Ch_i, location$, $B$)\\
             $Ch_k^a$ $\leftarrow$ get\_closest\_chain\_logic($state$, $l\_candidates\_{Ch_i}$, $Ch_j, l\_no\_intersection\_j, Ch_i, location$, $A$)\\
             $Ch_k$, $endpoint$ $\leftarrow$ select\_closest\_one($Ch_j$, $Ch_k^a$, $Ch_k^b$)\\
             connect\_two\_chains($state$, $Ch_j$, $Ch_k$, $l\_candidates\_{Ch_i}$, $endpoint$, $Ch_i$)\\
             $j\_pointer$ $\leftarrow$ update\_pointer($Ch_j$, $Ch_k$, $l\_candidates\_{Ch_i}$)\\
        }
    }
    $state$.update\_system\_status( $Ch_i$, $l\_s\_outward$, $l\_s\_inward$)\\
}
$l\_ch_{c}$,$l\_nodes_{c}$, $M$ $\leftarrow$ iterate\_over\_chains\_list\_and\_complete\_them\_if\_met\_conditions($state$)\\

\SetAlgoLined
\LinesNumbered
\BlankLine 
\KwRet{$l\_ch_{c}$,$l\_nodes_{c}$, $M$}
\caption{Connect Chains Main Logic}
\label{algo:connectmainlogic}
\end{algorithm}

Once we have a support chain ($Ch_i$), we iterate over the visible chains to connect them if they meet the similarity criteria as briefly shown in \Cref{algo:connectmainlogic}. 

The method $get\_closest\_chain\_logic$ (lines 14 and 15 of \Cref{algo:connectmainlogic}) provides the closest chain,  performing a dual symmetry check: if chain $Ch_k$ is the closest to $Ch_j$, then the closest one to $Ch_k$ should also be $Ch_j$. \Cref{algo:get_closest_chain} (function $get\_closest\_chain$), invoked within $get\_closest\_chain\_logic$, search for the closest candidate chain (described in \Cref{sec:connect_general_logic} item \ref{connect:item_recorrer_cadenas}) that meet conditions described in \Cref{sec:connect_general_logic} item \ref{connect:item_connectiviy_goodness}. In line 2, all chains in the neighborhood (defined by the $Ch_j$ endpoint and the $neighbourhood\_size$ $state$ attribute) are selected. For example, given the endpoint A with an angle of 0 degrees and $neighbourhood\_size=20$, all the chains included in $l\_candidates\_Ch_i$ with endpoint B angle in $[0-20,0]=[340,360]$ are selected and returned in ascending angular order concerning the endpoint of $Ch_j$. Lines 5 to 13 define the main loop. Two conditions determine an exit from the loop: either it exists a chain that satisfies conditions from \Cref{equ:similiraty_criterio} or no chains in the set $l\_sorted\_chains\_in\_neighbourhood$ satisfy the conditions. Line 7 of function \textbf{$connectivity\_goodness\_condition$} implements \Cref{equ:similiraty_criterio}. 

\begin{algorithm}[!htbp]
  \KwIn{$state$,\\
  $Ch_j$, // current chain   \\
  $l\_no\_intersection\_j$, // chains  that no intersect with $Ch_j$, set of candidates  to connect with $Ch_j$\\
  $Ch_i$, // support chain\\
  $location$, // inward o outward position of $Ch_j$ regarding to the support chain\\
  $endpoint$, // $Ch_j$ endpoint \\
  $M$, // intersection matrix\\
  }
\KwOut{closest chain to $Ch_j$ that satisfies the connectivity goodness conditions.\\}
\SetAlgoLined
\LinesNumbered
\BlankLine 
$neighbourhood\_size$ $\leftarrow$ $state.neighbourhood\_size$\\
$l\_sorted\_chains\_in\_neighbourhood$ $\leftarrow$ get\_chains\_in\_neighbourhood($neighbourhood\_size$, $l\_no\_intersection\_j$, $Ch_j$, $Ch_i$, $endpoint$, $location$)\\
$next\_id$ $\leftarrow$ 0\\
$Ch_k$ $\leftarrow$ None\\
\While{$len(l\_sorted\_chains\_in\_neighbourhood) > next\_id$}{
    $candidate\_chain$ $\leftarrow$ l\_sorted\_chains\_in\_neighbourhood[$next\_id$]\\
    $pass\_control$, $radial\_distance$ $\leftarrow$ connectivity\_goodness\_condition($state$, $Ch_j$, $candidate\_chain$, $Ch_i$, $endpoint$) // See  \Cref{algo:connectivity_goodness_condition}\\
    \If{$pass\_control$}{
        $Ch_k$ $\leftarrow$ get\_the\_closest\_chain\_by\_radial\_distance\_that\_does\_not\_intersect($Ch_j$, $endpoint$, $location$, $radial\_distance$, $candidate\_chain$, $M$, $l\_sorted\_chains\_neighbourhood$) // See Figure \ref{fig:connectivityIssue}.a \\
        break\\
    }
    $next\_id$ $\leftarrow$ $next\_id$ + 1\\
}

\KwRet{$Ch_k$}
\caption{Get closest chain}
\label{algo:get_closest_chain}
\end{algorithm}

\begin{figure}
\begin{center}
   \begin{subfigure}{0.47\textwidth}
    \def\svgwidth{\linewidth}
    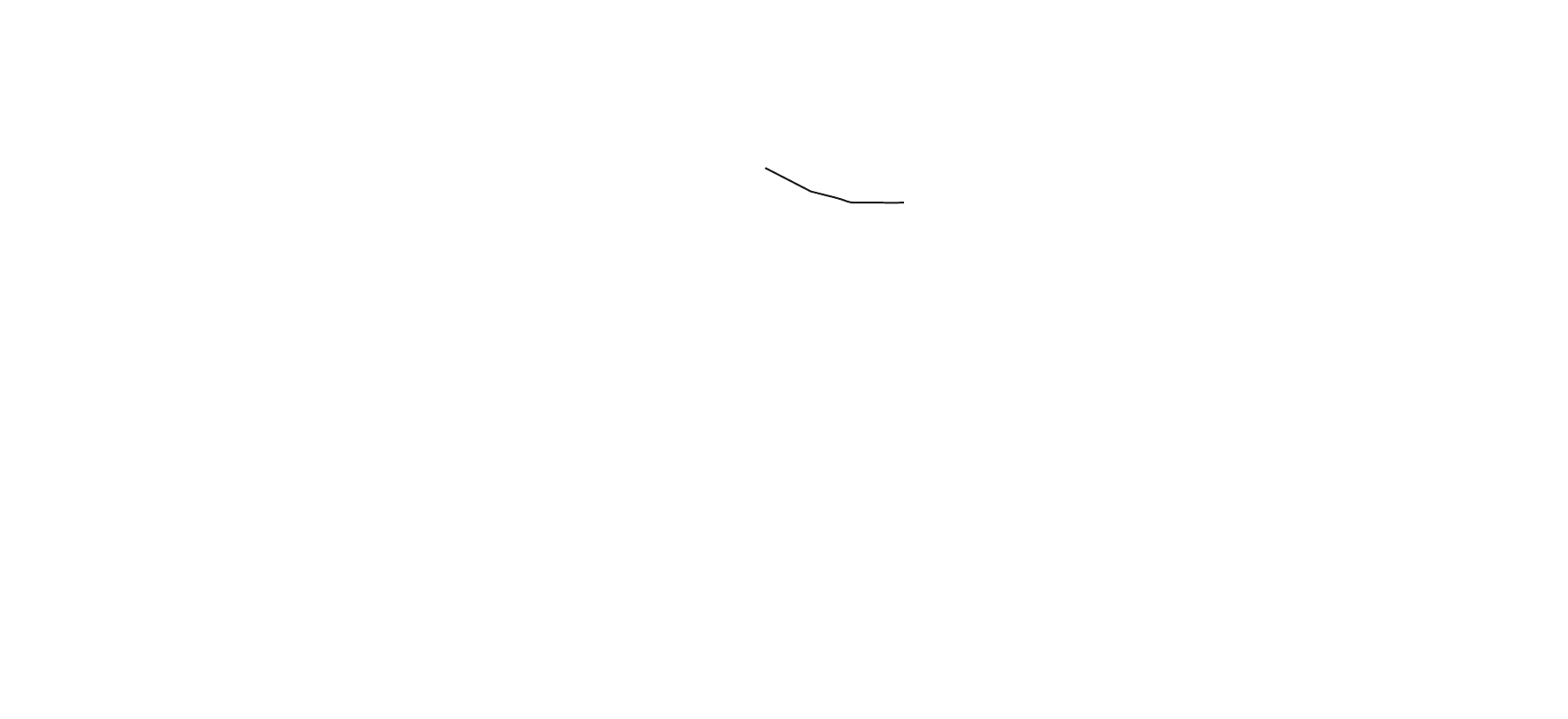
    \caption{} 
    \end{subfigure}
    \begin{subfigure}{0.30\textwidth}
    \def\svgwidth{\linewidth}
    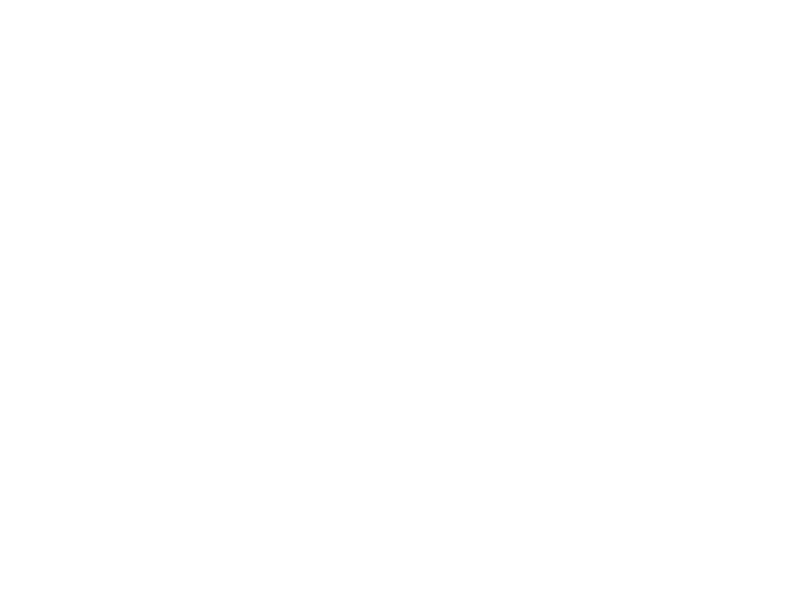
    \caption{}
    \end{subfigure}
    \caption{Chains connection. (a) $Ch_i$ is the support chain. The candidates chains for connection with $Ch_j$, are $Ch_{k}$ and $Ch_{l}$. The angular closest chain to $Ch_j$ is the noisy $Ch_{l}$. $Ch_{k}$ is the radially closest chain to $Ch_{j}$, these means that $\|r_{j}-r_{k}\|<\|r_{j}-r_{l}\|$. Where $r_i$ is the chain's endpoint distance to the support chain. (b) Endpoint condition check. See the text for an explanation.}
   \label{fig:connectivityClosest}
\end{center}
\end{figure}

If $candidate\_chain$ satisfies the conditions, a chain in the subset $l\_no\_intersection\_j$ could be closer to $Ch_j$ in terms of the connectivity goodness conditions but at a greater angular distance. This is tested in line 9 (shown in Figure \ref{fig:connectivityClosest}.a). In angular terms, the closest chain to $Ch_j$ satisfying \Cref{equ:similiraty_criterio} is $Ch_{l}$.
However, $Ch_{k}$ exists, which is more similar but not the closest one in terms of angular distance (\Cref{equ:angular}). To fix this, we get all the chains that intersect $Ch_{l}$ and satisfy \Cref{equ:similiraty_criterio} with $Ch_j$. We sort them by radial proximity to $Ch_j$, \Cref{equ:radial}, and return the best candidate chain as the closer one regarding the radial distance. 

The \textbf{connectivity\_goodness\_condition} function is described in \Cref{algo:connectivity_goodness_condition}. From lines 1 to 4, the parameters (\Cref{tab:connectivityParameters}) are extracted from the $state$ object (an instance from \textbf{SystemStatus} object). In line 6, the chain size condition is verified and saved in $size\_condition$. In line 7, the endpoint condition is verified. Figure \ref{fig:connectivityClosest}.b shows an example of this check where $Ch_i$ is the support chain for $Ch_{i+1}$ and $Ch_{i+2}$. Endpoints $A_{i+1}$ and $B_{i+2}$ from chains $Ch_{i+1}$ and $Ch_{i+2}$ are visible. Connecting endpoints $B_{i+1}$ and $A_{i+2}$ is impossible because they do not belong to the chain support $Ch_i$ angular domain. The \Cref{equ:similiraty_criterio} condition is verified in line 8. A boolean result about the similarity condition and the absolute difference between the mean of the radial distance of the nodes associated with $Ch_j$ and $candidate\_chain$ are returned. The later is defined as $distribution\_distance = \|mean(radials_{ch_j})-mean(radials_{candidate\_chain})\|$. Finally, in line 9, all conditions are verified.
The function returns a boolean value for the conditions (the variable \textit{check}) and the value of $distribution\_distance$.

\begin{algorithm}[!htbp]
  \KwIn{$state$,\\
  $Ch_j$, // current chain\\
  $candidate\_chain$, // chain closer to $Ch_j$\\
  $Ch_i$, // support chain of $Ch_j$ and $candidate\_chain$\\
  $endpoint$, // $Ch_j$ endpoint\\
  }
\KwOut{a boolean indicating if conditions are met, $distribution\_distance$}
\SetAlgoLined
\LinesNumbered
\BlankLine 
\tcc{Parameter extraction, from \Cref{tab:connectivityParameters}}
$th\_radial\_tolerance$ $\leftarrow$ $state.th\_radial\_tolerance$ \\ 
$th\_distribution\_size$$\leftarrow$$state.th\_distribution\_size$ \\
$th\_regular\_derivative$$\leftarrow$$state.th\_regular\_derivative$ \\
$derivative\_from\_center$$\leftarrow$$state.derivative\_from\_center$ \\

\tcc{Condition checks}
$distribution\_distance$ $\leftarrow$  None\\

$size\_condition$ $\leftarrow$ $Ch_j.size$ + $candidate\_chain.size$ $\leq$ Nr\\

$endpoint\_conditions$ $\leftarrow$  check\_endpoints($Ch_i, Ch_j, candidate\_chain, endpoint$) // see  \Cref{fig:connectivityClosest}.b \\

$similarity\_condition$, $distribution\_distance$ $\leftarrow$  similarity\_conditions($state$, $th\_radial\_tolerance$,$th\_distribution\_size$, $th\_regular\_derivative$, $derivative\_from\_center$, $Ch_i$, $Ch_j$, $candidate\_chain$, $endpoint$) // \Cref{equ:similiraty_criterio}\\

$check$ $\leftarrow$ $size\_condition$ and $endpoint\_condition$ and $similarity\_condition$\\ 

\KwRet{$check$, $distribution\_distance$}
\caption{Connectivity goodness condition}
\label{algo:connectivity_goodness_condition}
\end{algorithm}

\subsubsection{Postprocessing}

This last stage aims to complete the remaining chains, relaxing the conditions even more. Many chains are closed (i.e., $size = Nr$), which we call \textit{rings}. There are some non-closed chains (noisy or part of a ring) that, for some reason, are still incomplete. We use the information on the neighborhood chains to close or discard these remaining chains. The ideas on connecting chains are the same as those presented in the previous section.

\begin{figure*}
\begin{center}
   \includegraphics[width=0.3\linewidth]{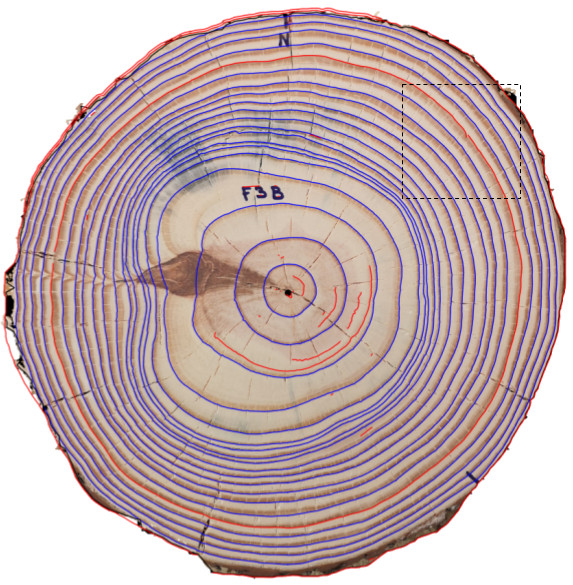}
   \includegraphics[width=0.3\linewidth]{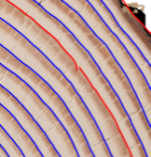}
   \includegraphics[width=0.3\linewidth]{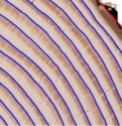}
   \caption{a) F03d disk after the connecting stage. b) There is a ring that cannot be closed because of chain intersection issues. c) The ring is closed after the postprocessing stage.}
   \label{fig:posprocessing_zoom}
\end{center}
\end{figure*}

It can remain some chains belonging to the same ring but not forming a closed chain. Often, this is due to a minor overlapping between chains (see \Cref{fig:posprocessing_zoom}). To solve this problem, we cut the overlapping chains in such a way as to avoid intersections between them and then try to reconnect the resulting chains that respect the connectivity goodness conditions.  

Given two closed chains containing a set of chains between them, suppose the added angular length of the non-overlapping chains between the rings is greater than 180 degrees. In that case, we consider that those chains have enough information about the ring, so we complete it by interpolating between rings and using the location of the existing chains. The chains that become part of the closed chain are the ones that meet the connectivity goodness conditions using the last column of Table \ref{tab:connectivityParameters}.

\subsection{Pith detection}
\label{sec:pith}
The pith position is an input for the method. Users can set it manually or using the method proposed by Decelle et al.,  \cite{ipolPith}, which is at the IPOL site.

\section{Implementation details.}

\label{sec:implementation}

The implementation was made in Python 3.11. The \textit{readme.md} file on the code repository contains all the information needed to run the code.

The demo requires, as input, an image of a tree slice and the pith position.

\Cref{tab:parameter} summarizes the method parameters, which the user can modify if needed. 
As output, the method returns a JSON file with the tree-rings position in Labelme format \cite{wada}.

\begin{table}[]
\centering
\begin{tabular}{|c|c|c|c|c|}
\hline
& stage & Parameter & Description & Default    \\ \hline
Basic & Edges detector & --sigma &Gaussian filtering $\sigma$  & 3       \\ \cline{2-5} 
& Preprocessing & --height & Rezised image height         &  None\\ \cline{3-5}
& & --width & Rezised image width &  None\\ \cline{2-5}
& Filtering, sampling, connect && Pith Position  & Required\\ \hline
Advanced  & Edges detector  & --th\_low &Gradient threshold low & 5\\ \cline{3-5} 
&  & --th\_high & Gradient threshold high & 15 \\ \cline{2-5} 
& Edges filtering  &--alpha & Collinearity threshold ($\alpha$) & 30° \\ \cline{2-5} 
& Sampling  & --nr &Number of rays ($N_r$)  & 360        \\ \cline{3-5} 
&  & --min\_chain\_lenght &Minimum chain length                      & 2          \\ \hline
\end{tabular}
\caption{Method parameters. The user can modify basic parameters in the demo.}
\label{tab:parameter}
\end{table}

\section{Experiments and Results}
\label{sec:ExperimentsAndResults}

\subsection{Datasets}
\label{sec:database}

We use two datasets to evaluate the CS-TRD method. 

The first one (\textbf{UruDendro}) is an online database \cite{UruDendro} with images of cross-sections of fourteen commercially grown \textit{Pinus taeda} trees from northern Uruguay. These trees are 13 to 24 years old and were collected in February 2020. The cross-sections are between 5 and 20 cm thick and were dried at room temperature without further preparation, which resulted in the development of radial cracks and blue fungus stains. Surfaces were smoothed with a handheld planer and a rotary sander. The dataset consists of 64 images of different resolutions, ranging between 1000 and 3000 pixels in width. It contains challenging features for automatic ring detection, including varying illumination and surface preparation, fungal infection (blue stains), knots, missing bark and interruptions in outer rings, and radial cracking. At least two experts annotate the images using the Labelme tool \cite{wada}. Figure \ref{fig:ddbb} shows some images from the dataset. 

The second dataset (\textbf{Kennel}) is proposed in \cite{KennelBS15}, which made available a public set of 7 (1280x1280 pixels) images of Abies alba along with a method for detecting tree rings (the code is not available). We labeled the dataset with the same procedure as the \textbf{UruDendro} dataset as we could not process the annotations given by the authors.

\subsection{Metrics}
\label{sec:metric}
\begin{figure*}
\begin{centering}
    \begin{subfigure}{0.3\textwidth}
   \includegraphics[width=\textwidth]{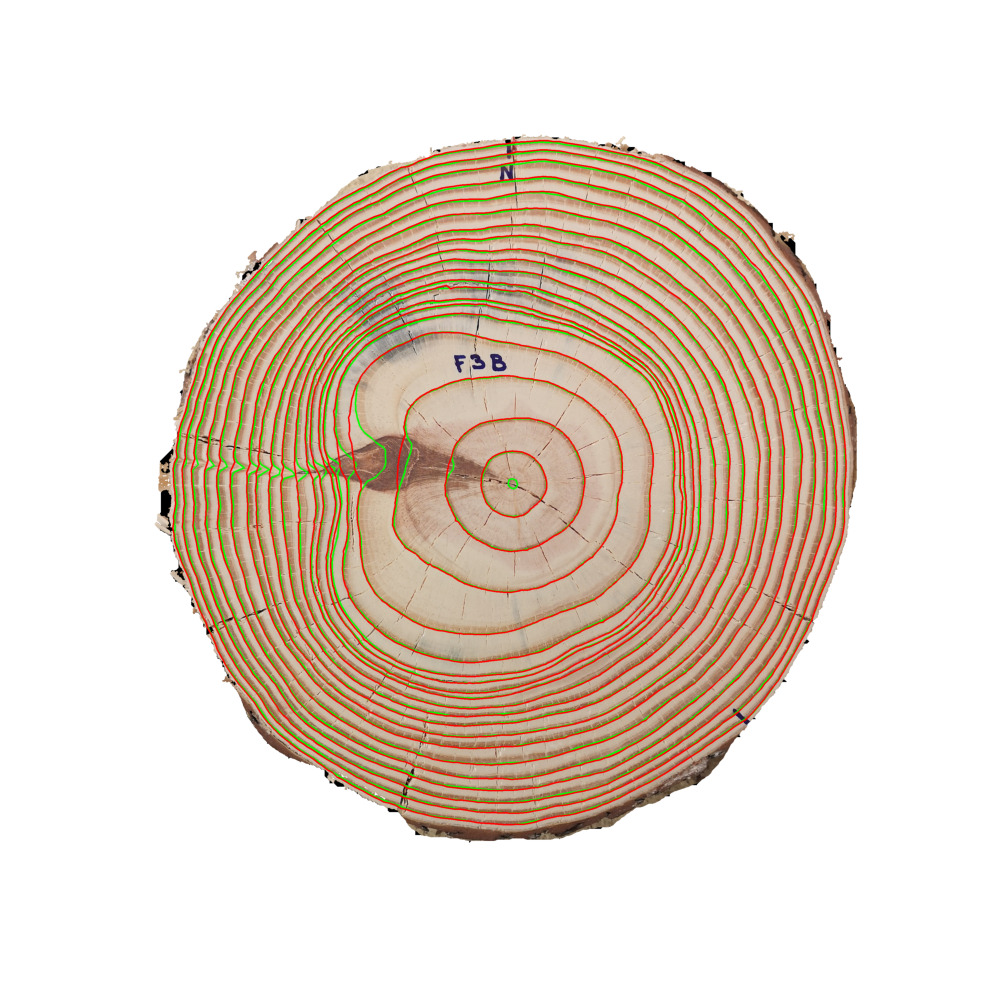}
    \caption{}
    \end{subfigure}
    \begin{subfigure}{0.3\textwidth}
   \includegraphics[width=\textwidth]{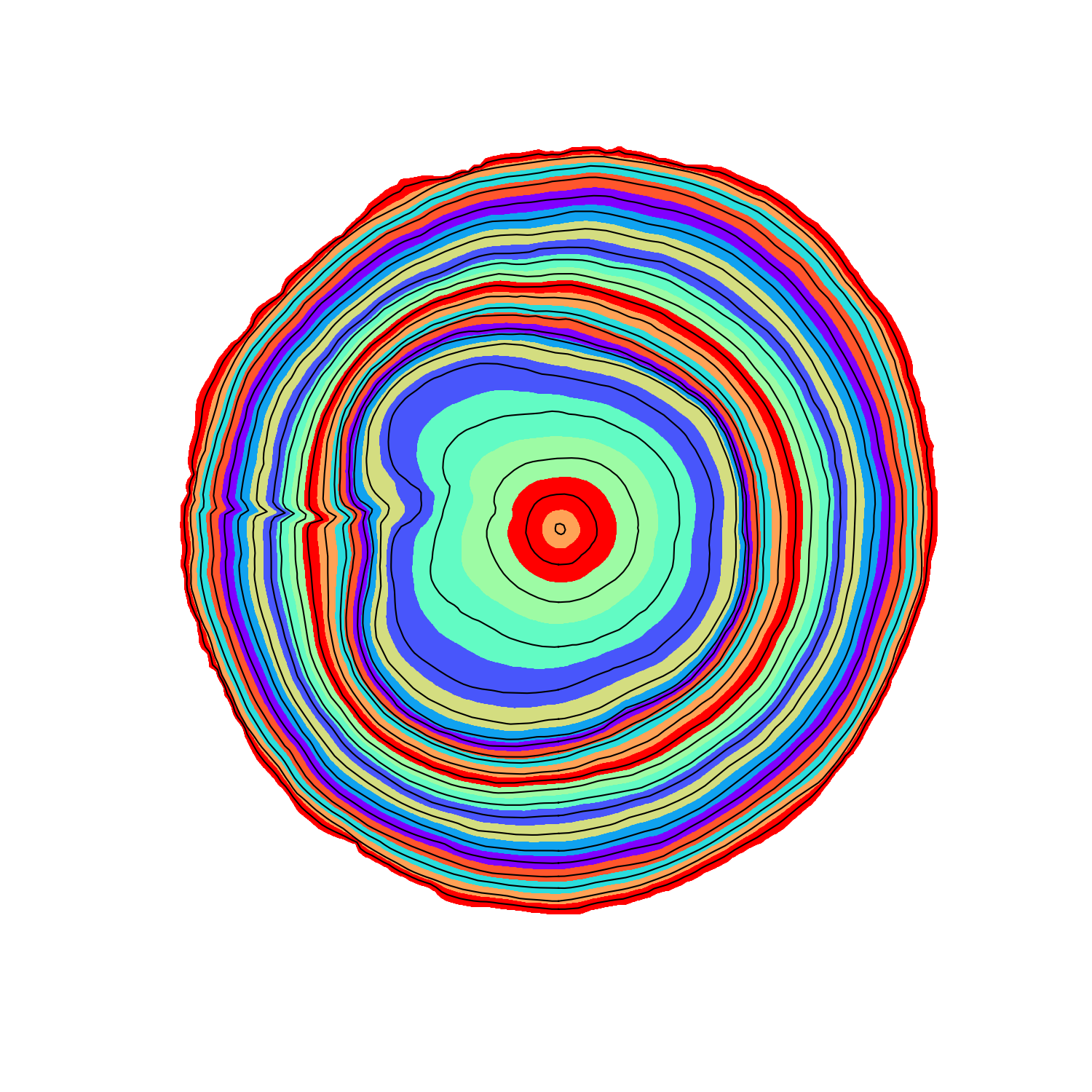}
    \caption{}
    \end{subfigure}
       \begin{subfigure}{0.3\textwidth}
   \includegraphics[width=\textwidth]{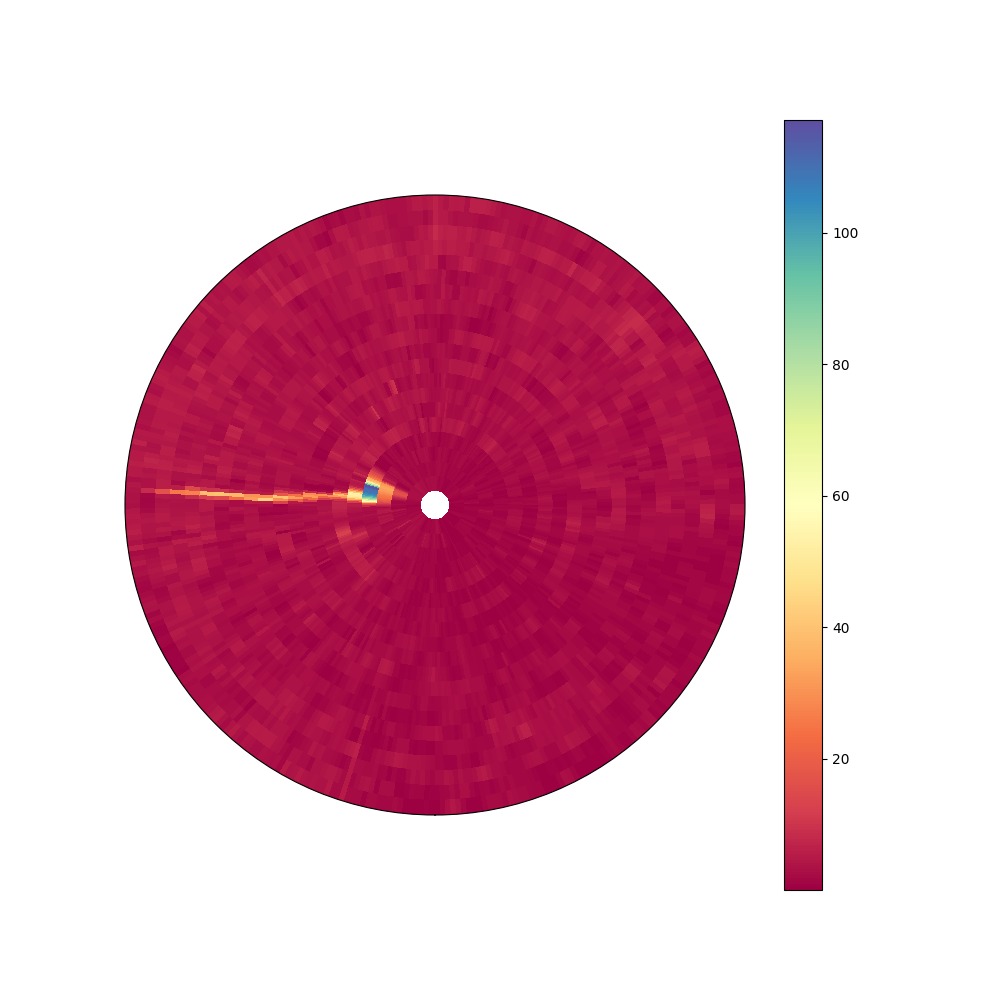}
    \caption{}
    \end{subfigure}
   \caption{Measuring the error of automatic detections for image F03d: (a) In green, the GT; in red, the detections produced by the method. (b) Areas of influence of the GT rings. (c) Absolute error, in pixels, between the detections and the GT.}
   \label{fig:influece_area}
\end{centering}
\end{figure*}

To assess the method, we developed a metric based on the one proposed by Kennel et al. \cite{KennelBS15}. To determine if a ring is detected, we define a ring influence area as the set of pixels closer to that ring. For each ray, the frontier is the midpoint between the nodes of consecutive ground truth (GT) rings. \Cref{fig:influece_area}.b show the influence area for rings in disk F03d.  
\Cref{fig:influece_area}.a shows the detections (in red) and GT marks (in green) for the same image.

The influence region associates a detected curve with a GT ring. In both cases, nodes are associated with the $N_r$ rays. Given a GT ring, we assign it to the nearest detection using:

\begin{equation}
    Dist = \sqrt{\frac{1}{N_r}\sum_{i=0}^{N_r-1}\left(dt_i-gt_i\right)^2}\hspace{0.5cm} 
    \label{equ:rmse}
\end{equation}

Where $i$ represents the ray direction, $dt_i$ is the radial distance (\Cref{equ:radial}) of detected node $i$, and $gt_i$ is the radial distance (\Cref{equ:radial}) of the corresponding GT node $i$. 

The closest detection may be far from the corresponding GT ring. To match a detected curve with a GT ring, it is essential to ensure that the identified chain is the closest one to the ring and that it is sufficiently close. We utilize the influence area of each GT ring (\Cref{fig:influece_area}.b). Upon detecting a curve, if the proportion of nodes within that chain that falls within the influence area of the nearest ring exceeds a specified threshold parameter ($th\_pre=60\%$, see \Cref{sec:precisionTrheshold}), we assign the detected curve to the corresponding GT ring. If it falls below the threshold, the detection is not associated with any GT ring. In other words, for a detected curve to be assigned to a GT ring and considered a true positive, at least 60\% of its nodes must be within the influence area of that GT ring.

We define the absolute error $A\epsilon_i$ (in pixels) for the \textit{node} $i$ as the absolute difference in pixels between the GT ring and the detected ring associated with it:

$$
A\epsilon_i = |r^d_i - r^{GT}_i|
$$

Where $r^d_i$ is the radial distance from the center to node $i$ of the detected ring, and $r^{GT}_i$ is the same for the GT ring. Figure \ref{fig:influece_area}.c shows the absolute error between the GT and the detected rings assigned to them. Red represents a low error, while yellow, green, and blue represent a higher error. As can be seen, the error is concentrated around the knot, affecting the precise detection of some rings.

Once all the detected chains are matched with the GT rings, we calculate the following values:
\begin{enumerate}
    \item True Positive (TP): when the identified closed chain and the GT ring match.
    \item False Positive (FP): when the identified closed chain doesn't match any GT ring.
    \item False Negative (FN): when a GT ring doesn't match any detected closed chain.
\end{enumerate}

Precision is given by $P=\frac{TP}{TP+FP}$, Recall by
$R=\frac{TP}{TP+FN}$ and the F-Score by $F=\frac{2PR}{P+R}$.

\subsection{Experiments}
\label{sec:experiments}

This section presents some experiments to help us better understand the method and its limitations. 

\Cref{tab:best_res} presents the metric results for the optimal values of $\sigma$ and image resolution on both datasets. For example, CS-TRD fails to detect one ring in the image \textit{F03d} ($FN=1$), and the other rings are correctly detected, leading to the displayed values of $P$, $R$, and $F$. The table also compares the mean execution time by image and the RMSE error in pixels (\Cref{equ:rmse}) between the detected and GT rings. All experiments used an Intel Core i5 10300H workstation with 16GB of RAM. These are very good results, considering the data diversity and the presence of perturbations.

\begin{table}[!ht]
\centering
\begin{tabular}{|l|c|c|c|c|c|c|c|}
\hline
Dataset                                          & \multicolumn{1}{l|}{Image Size (pixels)} & \multicolumn{1}{l|}{$\sigma$} & \multicolumn{1}{l|}{P} & \multicolumn{1}{l|}{R} & \multicolumn{1}{l|}{F} & \multicolumn{1}{l|}{RMSE} & \multicolumn{1}{l|}{Execution Time (sec.)} \\ \hline
UruDendro & 1500x1500 & 3.0 & 0,95 & 0,86 & 0,89 & 5.27 & 17.3 \\ \hline
Kennel & 1500x1500  & 2.5 & 0,97 & 0,97 & 0,97 & 2.4  & 11.1   \\ \hline
\end{tabular}
\caption{Mean performance and execution time for both datasets' at the optimal $\sigma$ and image resolution.}
\label{tab:best_res}
\end{table}

\subsubsection{Edge detector optimization stage}
\label{exp:size}
\begin{figure*}[ht]
\begin{centering}
\begin{subfigure}{0.4\textwidth}
    \includegraphics[width=\textwidth]{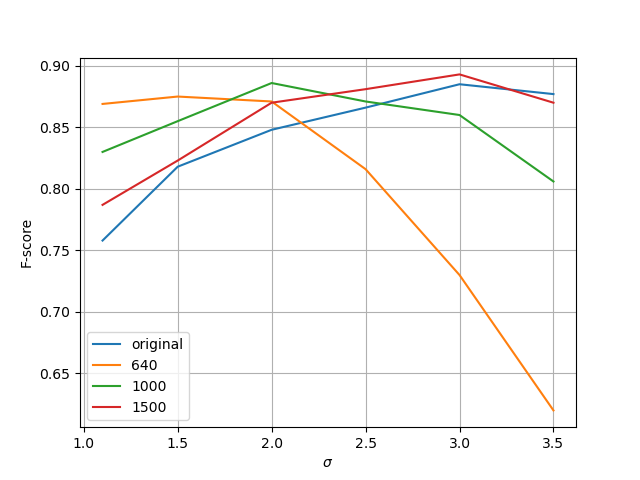}
    \caption{}
    \end{subfigure}
    \begin{subfigure}{0.4\textwidth}
    \includegraphics[width=\textwidth]{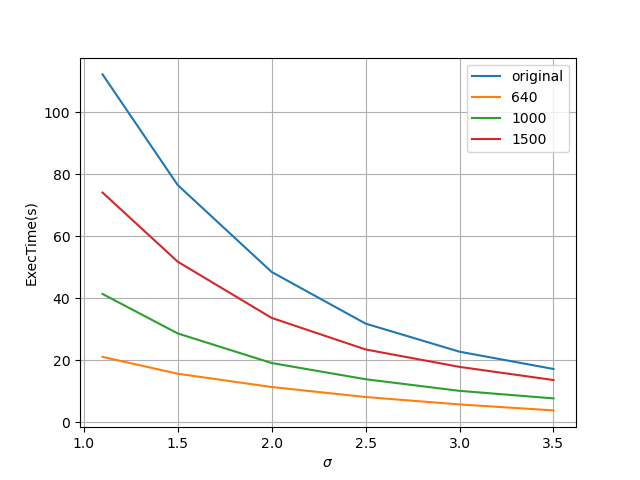}
    \caption{}
    \end{subfigure}
    \begin{subfigure}{0.4\textwidth}
    \includegraphics[width=\textwidth]{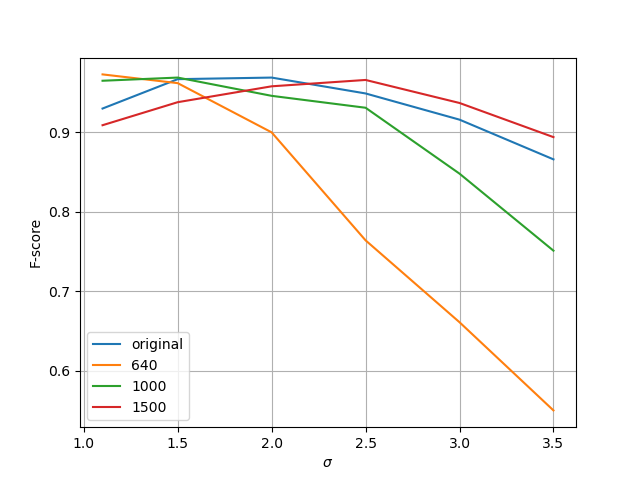}
    \caption{}
    \end{subfigure}
    \begin{subfigure}{0.4\textwidth}
    \includegraphics[width=\textwidth]{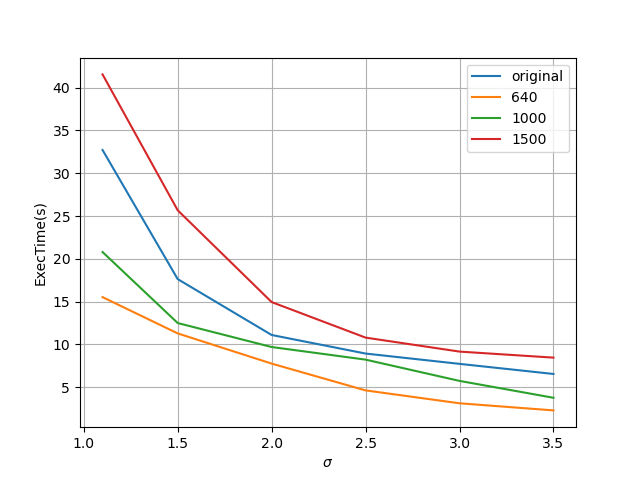}
    \caption{}
    \end{subfigure}
   \caption{ Influence of the image size and edge detector $\sigma$ parameter experiment. Each curve represents a different image resolution: $640\times640$, $1000\times1000$, $1500\times1500$, and the original resolution (in blue). (a) Average F1 vs. $\sigma$ for different image sizes over the UruDendro dataset. (b) Average execution time (in seconds) vs $\sigma$ for different image sizes over the UruDendro dataset. (c) Average F1 vs. $\sigma$ for different image sizes over the Kennel dataset. (d) Average execution time (in seconds) vs $\sigma$ for different image sizes over the Kennel dataset.
   }
   \label{fig:sigma-results_ac}
\end{centering}
\end{figure*}

The algorithm heavily relies on the edge detector stage. In the first experiment, we test different $\sigma$ values for the Canny Devernay edge detector to get the one that maximizes the F-Score for the UruDendro dataset. This dataset shows significant variations in image resolution, allowing us to study the overall performance with different input image dimensions. The results are presented in Figures \ref{fig:sigma-results_ac}.a and \ref{fig:sigma-results_ac}.b. We compute the average F-Score for the original image sizes and then scale all images in the dataset to $640\times640$, $1000\times1000$, and $1500\times1500$ pixels. The best result was achieved for the size $1500\times1500$ with $\sigma=3.0$. The execution time varies with the image size. The average execution time for the $1500\times1500$ size is 17 seconds. The execution time decreased as $\sigma$ increased because fewer edge chains were detected. Results for the same experiment on the Kennel dataset are shown in Figures \ref{fig:sigma-results_ac}.c and \ref{fig:sigma-results_ac}.d. The best F-Score is achieved for the $1500\times1500$ resolution with $\sigma=2.5$. The lower optimal $\sigma$ value can be attributed to the Kennel dataset having images with more rings on the disk, averaging 30 rings. In comparison, the UruDendro one has 19 rings per disk on average. Table \ref{tab:best_res} summarizes the results of this experiment.

\subsubsection{Pith position sensibility}
\begin{figure*}
\begin{centering}
   \begin{subfigure}{0.28\textwidth}
   \includegraphics[width=\textwidth]{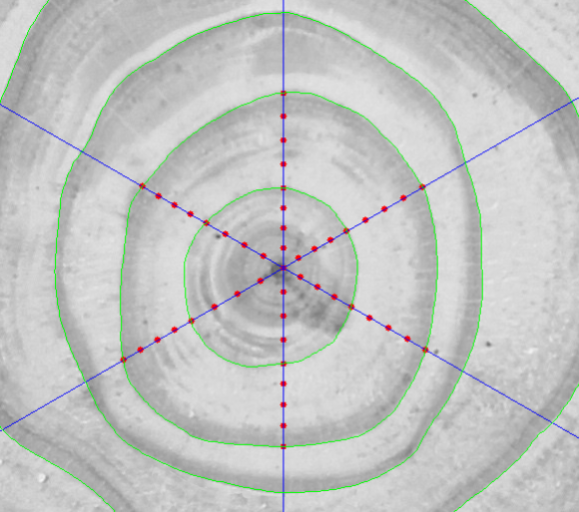}
    \caption{Pith position grid}
    \end{subfigure}
    \begin{subfigure}{0.35\textwidth}
    \includegraphics[width=\textwidth]{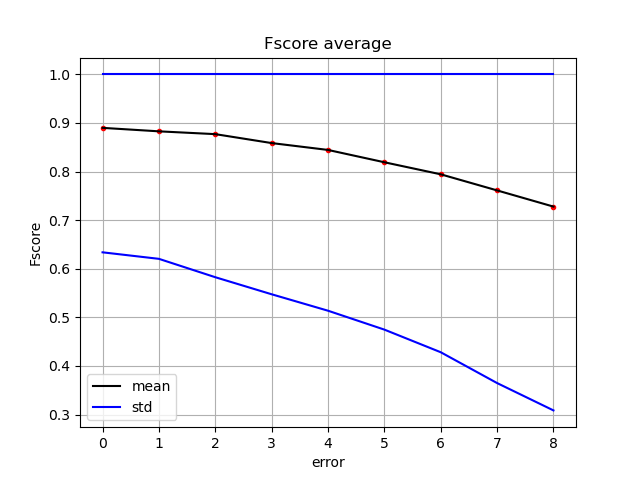}
    \caption{Average F-Score over the UruDendro dataset.}
    \end{subfigure}
    \begin{subfigure}{0.35\textwidth}
    \includegraphics[width=\textwidth]{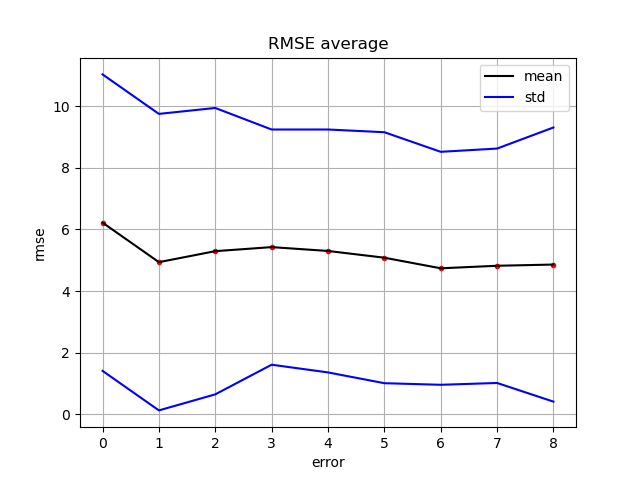}
    \caption{Average RMSE over the  UruDendro dataset.}
    \end{subfigure}
   \caption{Pith position experiment. (a) Eight different pith positions are marked given six ray directions. We executed the method for each marked pith position. GT rings are in green. (b and c) For each disk of the UruDendro dataset, we run the method using the 48 different pith positions. Results are averaged over the six rays' directions per error position. The X-axis values correspond to increments of 25 \% in error, i.e., the value 2 implies an error of 50 \% of the first ring equivalent diameter in the pith position concerning the GT.}
   \label{fig:average_pith_position_exp}
\end{centering}
\end{figure*}

The next experiment assesses the method's sensitivity to errors in the pith estimation. \Cref{fig:average_pith_position_exp}.a shows 48 different pith positions used in this experiment (eight different pith positions across six rays). These radially displaced pith positions are selected as follows:
\begin{itemize}
    \item We define an error step along a ray as 25\% of the first ring equivalent diameter.
    \item Three positions are marked inside ring 1, with errors  25\%, 50\%, and 75\% off the GT center in the ray direction.
    \item One position is marked on ring 1.
    \item Three positions are marked between the first and second rings, with the same increasing errors in the ray direction.
    \item Finally, another position is marked on ring 2.
\end{itemize}

We executed the algorithm for each disk and pith position of the UruDendro dataset (size of $1500\times1500$ and $\sigma=3.0$), resulting in 48 outcomes. We calculated the average RMSE and F-Score measurements for the six-ray directions for each radially displaced pith position. This produced two eight-coordinate vectors, one for RMSE and one for the F-Score. Figures \ref{fig:average_pith_position_exp}.b and \ref{fig:average_pith_position_exp}.c illustrates the average F-Score and RMSE for each error position across the dataset, respectively. The F-Score decreases as the error in the pith estimation increases, while the RMSE is less sensitive to pith errors.
 
\subsubsection{Metric precision threshold}
\label{sec:precisionTrheshold}
\begin{figure*}[ht]
\begin{centering}
    \begin{subfigure}{0.35\textwidth}
    \includegraphics[width=\textwidth]{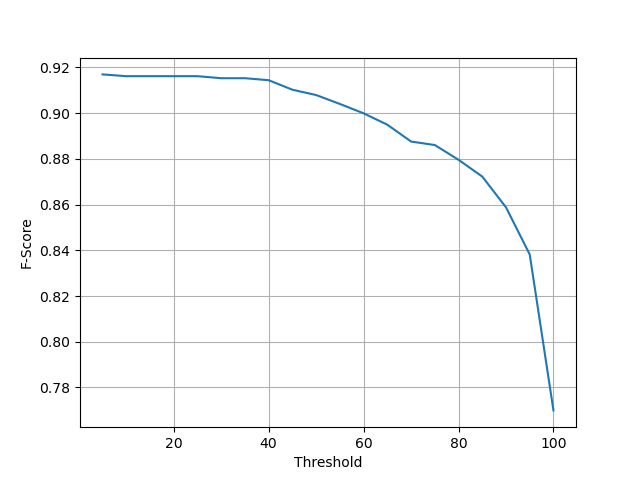}
    \end{subfigure}
    \begin{subfigure}{0.35\textwidth}
    \includegraphics[width=\textwidth]{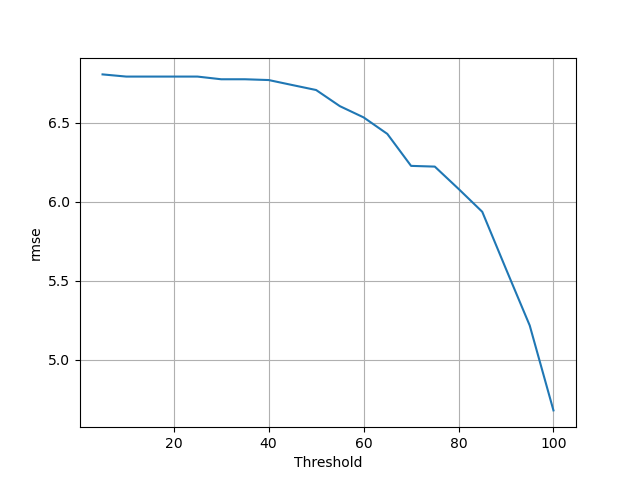}
    \end{subfigure}
    \begin{subfigure}{0.35\textwidth}
    \includegraphics[width=\textwidth]{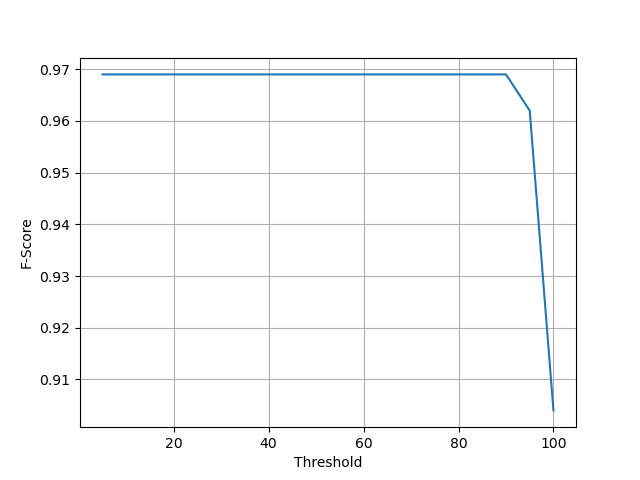}
    \end{subfigure}
    \begin{subfigure}{0.35\textwidth}
    \includegraphics[width=\textwidth]{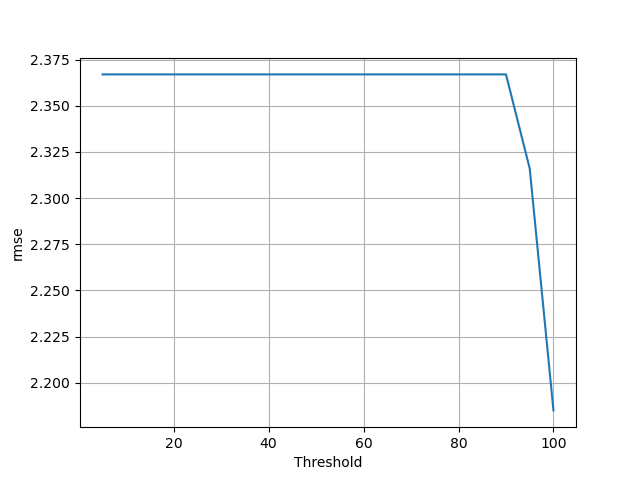}
    \end{subfigure}
   \caption{Performance metrics were computed for different values of the $th\_pre$ parameter. The first row shows the results for the Urudendro dataset, and the second row for the Kennel dataset. The left column shows the average F-Score, and the right column shows the average RMSE.}
   \label{fig:metric_our_database_ac}
\end{centering}
\end{figure*}

In this experiment, we examine how performance changes with different values of the $th\_pre$ parameter, which determines the number of ring nodes within the influence area to be considered in the detection-to-ground-truth assignation step. Figure \ref{fig:metric_our_database_ac} displays the results for the UruDendro and Kennel datasets. As anticipated, higher precision leads to lower RMSE but lower F-Score. Based on these findings, we fixed $th\_pre = 60\%$ as the default value, which appears to be a good compromise.

\subsection{Results}
\label{sec:results}
\begin{figure*}[h!]
\begin{centering}
    \begin{subfigure}{0.3\textwidth}
    \includegraphics[width=\textwidth]{F02b_output}
    \caption{F02b}
    \end{subfigure}
    \begin{subfigure}{0.3\textwidth}
    \includegraphics[width=\textwidth]{F07b_output}
    \caption{F07b}
    \end{subfigure}
    \begin{subfigure}{0.3\textwidth}
    \includegraphics[width=\textwidth]{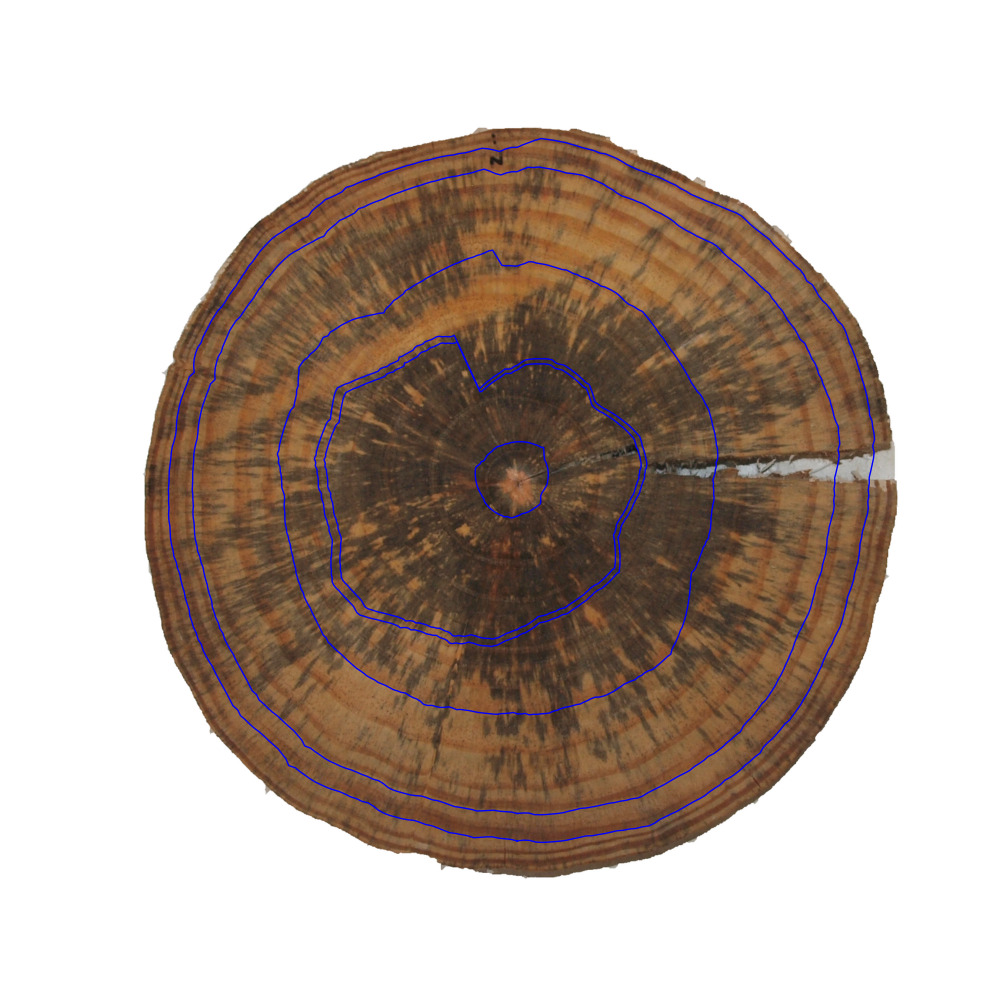}
    \caption{L02b}
    \end{subfigure}
   \caption{The CS-TRD method predicted detections (shown as blue curves) in some images of the UruDendro dataset.}
   \label{fig:ddbb_output}
\end{centering}
\end{figure*}

This section provides a quantitative analysis of the CS-TRD method performance on challenging images in the datasets. \Cref{fig:ddbb_output} illustrates some results of the CS-TRD over the UruDendro dataset. Note the successful performance of CS-TRD on disks with cracks (F02b), knots (L03c), and fungus (L02b). The mean F-Score for the whole dataset is 0.89 (\Cref{tab:best_res}), indicating the successful detection of rings in complex images containing knots, fungus, and cracks.  Results of the CS-TRD algorithm on other samples of the Urudendro dataset are in the supplemental material. %

\begin{figure*}[ht!]
\begin{centering}
\begin{subfigure}{0.3\textwidth}
\includegraphics[width=\linewidth]{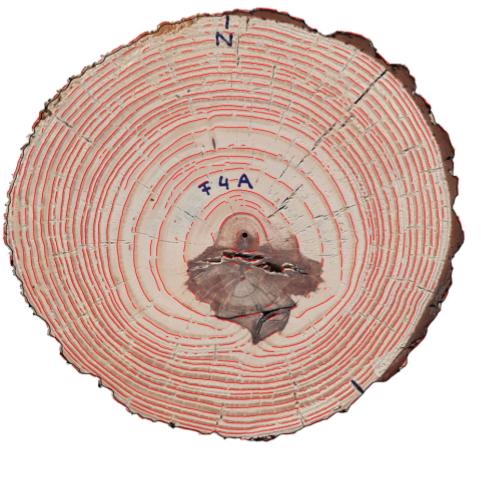}
\caption{Chains}
\end{subfigure}
\begin{subfigure}{0.3\textwidth}
\includegraphics[width=\linewidth]{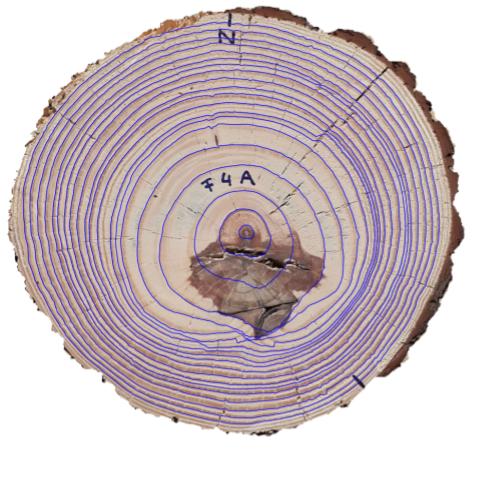}
\caption{Output}
\end{subfigure}
\begin{subfigure}{0.3\textwidth}
\includegraphics[width=\linewidth]{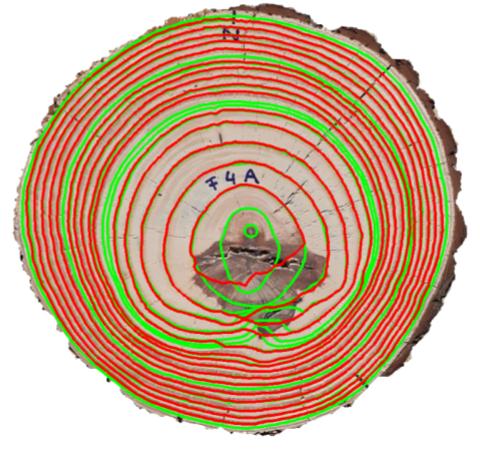}
\caption{Gt and Dt}
\end{subfigure}
\begin{subfigure}{0.3\textwidth}
\includegraphics[width=\linewidth]{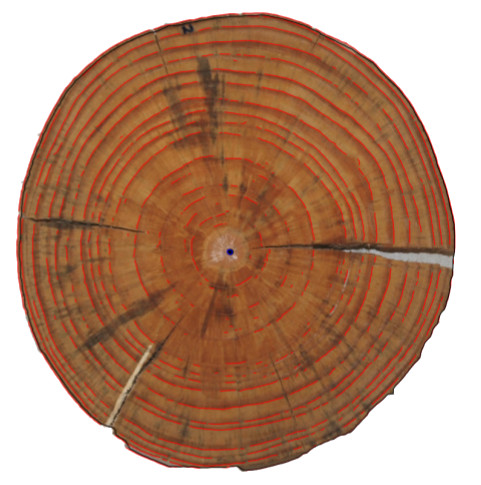}
\caption{Chains}
\end{subfigure}
\begin{subfigure}{0.3\textwidth}
\includegraphics[width=\linewidth]{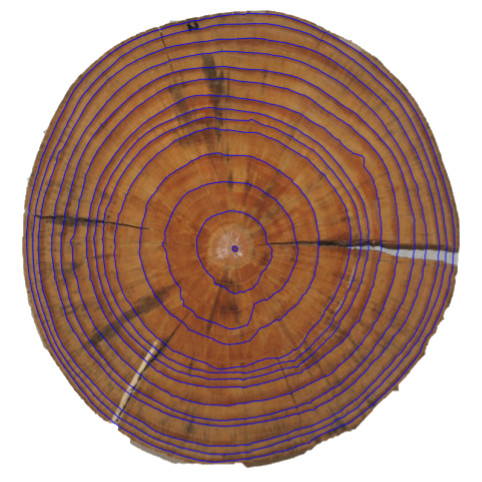}
\caption{Output}
\end{subfigure}
\begin{subfigure}{0.3\textwidth}
\includegraphics[width=\linewidth]{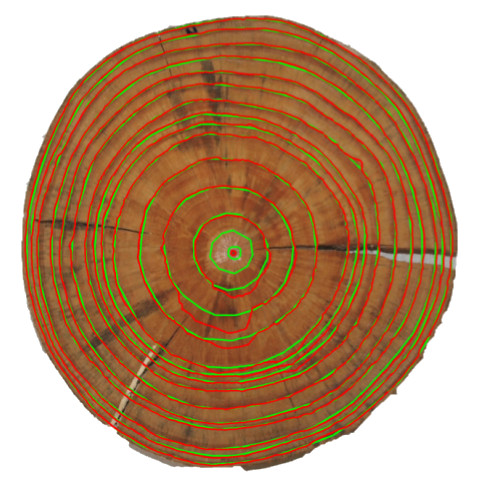}
\caption{Gt and Dt}
\end{subfigure}
\caption{Upper row: CS-TRD result for disk F04c. Note the impact of the knot on the edge detection step. (a) chains, (b) detected rings, (c) GT rings in green, and detected rings in red. Lower row: CS-TRD result for disk L09e. The method successfully detects almost all the rings (FN=2 and FP=0) despite cracks and fungus stains. (d) chains, (e) detected rings, (f) GT rings in green, and detected rings in red.}
\label{fig:analysis}
\end{centering}
\end{figure*}

The upper row of Figure \ref{fig:analysis}
illustrates how the method performs in the presence of knots. The disk F04c fails to detect the first and third rings, identifies a false ring over the knot, and misses the last ring. Despite these errors, the method detected 19 rings, with only one false detection, and missed two rings, resulting in an F1-Score of 90\%. The lower row of \Cref{fig:analysis} shows the CS-TRD results for disk L09e. Despite two significant cracks and several fungus stains, the method successfully detects 13 out of 15 rings, with one false detection. As a result, it achieves an F1-Score of 90\%.

\begin{figure*}[ht!]
\begin{centering}
    \begin{subfigure}{0.3\textwidth}
    \includegraphics[width=\textwidth]{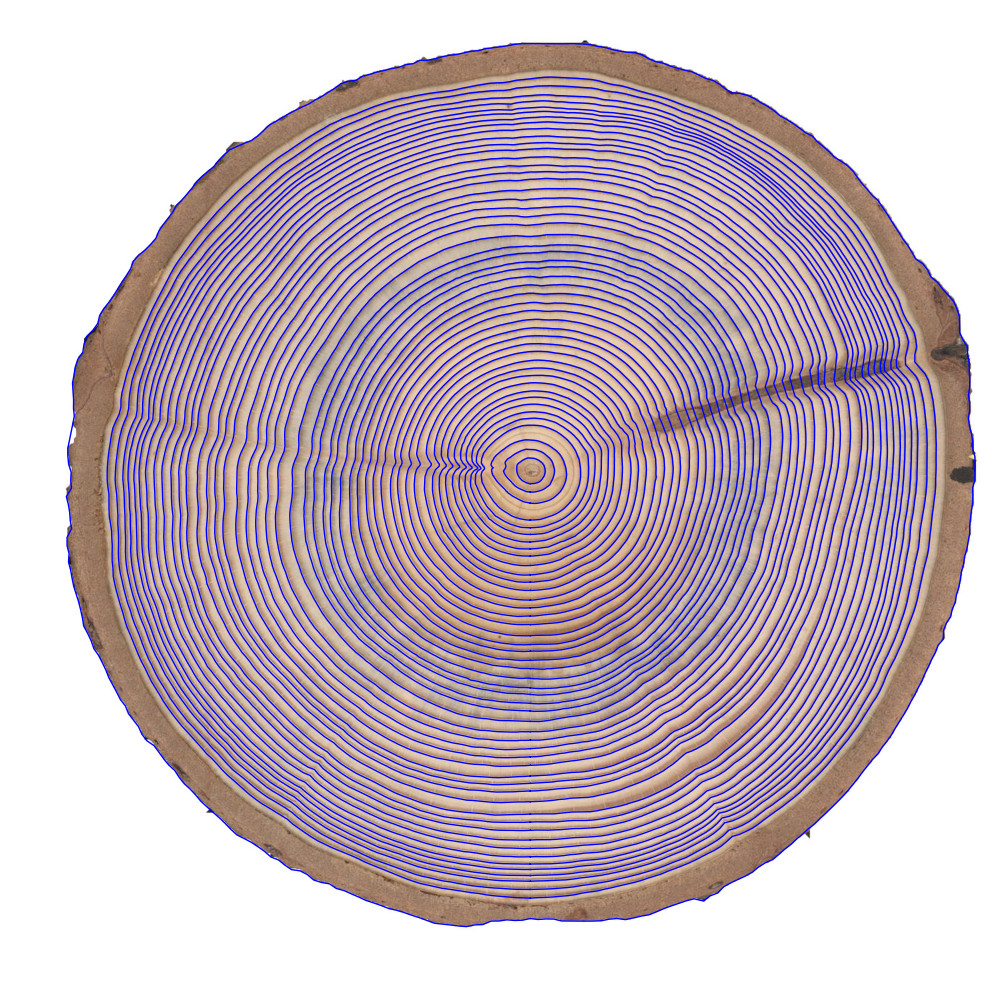}
    \caption{AbiesAlba1}
    \end{subfigure}
    \begin{subfigure}{0.3\textwidth}
    \includegraphics[width=\textwidth]{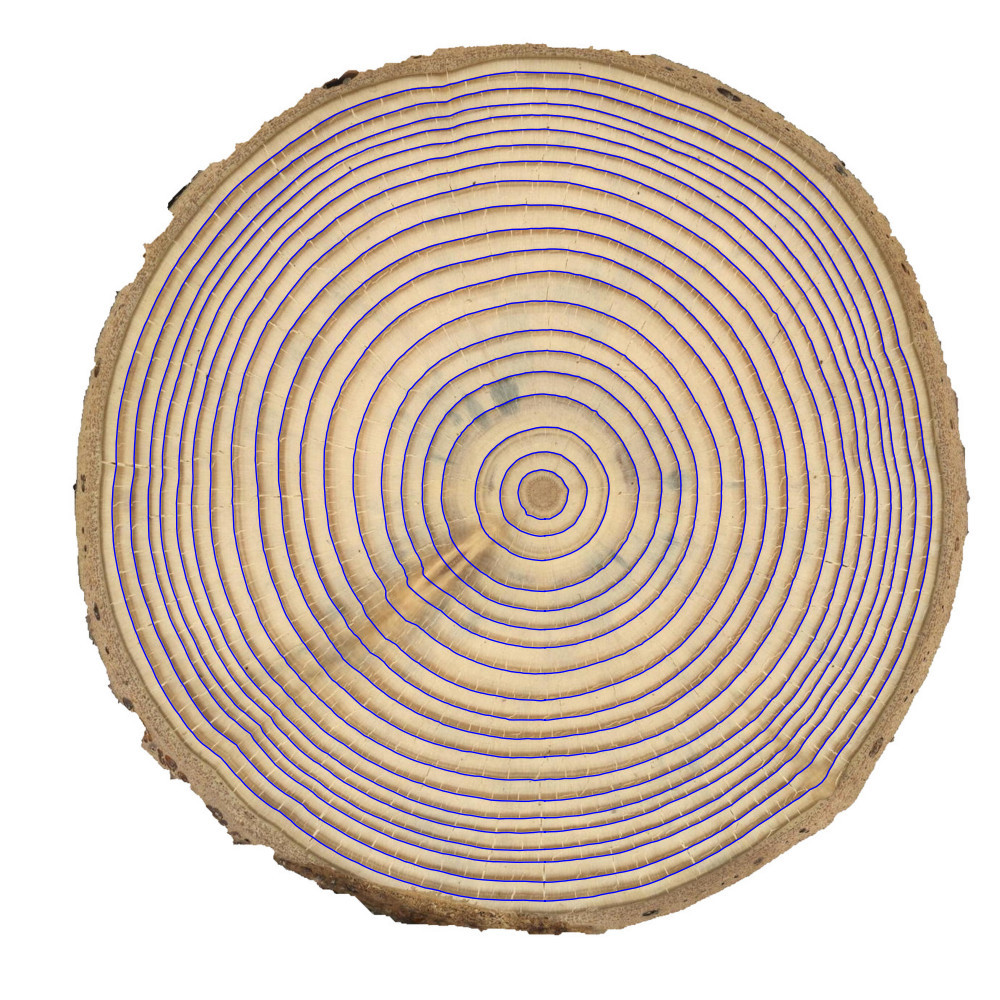}
    \caption{AbiesAlba2}
    \end{subfigure}
    \begin{subfigure}{0.3\textwidth}
    \includegraphics[width=\textwidth]{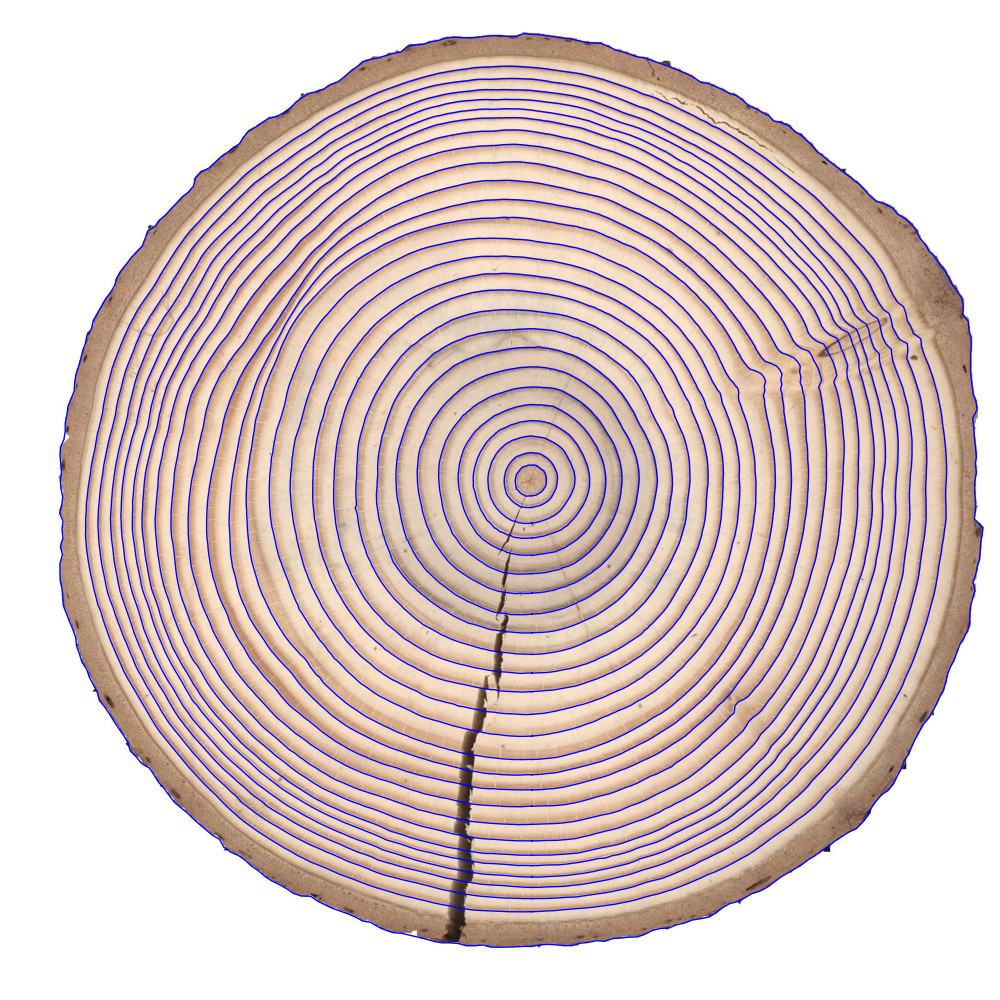}
    \caption{AbiesAlba3}
    \end{subfigure}
   \caption{CS-TRD results for images from the Kennel dataset with $1500\times1500$ image size and $\sigma=2.5$.}
   \label{fig:ddbb_ac_output}
\end{centering}
\end{figure*}

\begin{figure*}[ht]
\begin{centering}
\begin{subfigure}{0.3\textwidth}
\begin{centering}
\includegraphics[width=\linewidth]{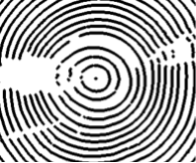}
\caption{Filter}
\label{fig:ac1_filter}
\end{centering}
\end{subfigure}
\begin{subfigure}{0.3\textwidth}
\begin{centering}
\includegraphics[width=\linewidth]{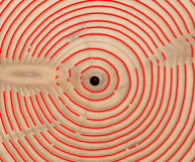}
\caption{Chains}
\label{fig:ac1_chains}
\end{centering}
\end{subfigure}
\begin{subfigure}{0.3\textwidth}
\begin{centering}
\includegraphics[width=\linewidth]{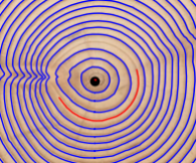}
\caption{Postprocessing}
\label{fig:ac1_post}
\end{centering}
\end{subfigure}

\caption{CS-TRD result for AbiesAlba1 image from the Kennel Dataset (zoom over pith center). a) Filter stage output, b) Chain stage output, c) Postprocessing stage output. In a) and b), we can see that the method fails to detect edges for the pith. The $\sigma$ threshold may be too high to detect things at this resolution. In c), we can see that the red chain was not closed due to a size smaller than the $information\_threshold$ value (180)}.

\label{fig:ac1_analysis}
\end{centering}
\end{figure*}

Some results for the Kennel dataset are shown in \Cref{fig:ddbb_ac_output}. The results for the whole dataset can be seen in the supplementary material. The mean F-Score is 0.97 (\Cref{tab:best_res}). At most, three rings are not detected per disk. In the worst case, one ring is mistakenly detected per image, usually the last one (sometimes incomplete) or the core.  In \Cref{fig:ac1_analysis}, we illustrate the example of disk \textit{AbiesAlba1}. The edge parameter is set too high ($\sigma=2.5$), causing the edge detector to fail to detect the pith. Additionally, the red chain in \Cref{fig:ac1_analysis}.c is not closed because its size is smaller than 180 degrees (the $information\_threshold$ parameter is fixed once and for all). However, the method effectively detects the rings over the knot. 

\subsection{Comparison between CS-TRD and INBD methods.}
\label{sec:inbd}

The only other method with available code for automatically detecting tree rings in wood cross-section images is the Iterative Next Boundary Detection Instance Segmentation (INBD) proposed by Gillert et al. \cite{inbd}. Note that this method was designed for Microscopy Images of Shrub Cross Sections, which differ from the ones we are working with regarding image resolution and species characteristics. In the INBD method, the image is segmented into the background, ring boundaries, and the pith region. Then, the circular image is transformed into polar coordinates using the pith's center as the origin. Iterative image patches are then extracted, and rings are segmented individually from the inner to the outer rings. Both stages employ a U-NET network.

Furthermore, the ground truth pith location is used as input in the experiments for the CS-TRD. To ensure a fair comparison, the second stage of the INBD method is modified to take the pith boundary as input.

To train the INBD model with the UruDendro dataset, we randomly divided it into train, validation, and test sets with 40, 12, and 12 images, respectively. Guillert et al. trained their model with the EH dataset, which consists of 82 images with 949 rings. Our UruDendro dataset comprises 64 images and 1123 rings. For training the INBD model, each image is divided into patches determined by successive rings in polar coordinates, ensuring that each patch includes an entire ring. Even though the EH dataset has more images, the UruDendro dataset has more rings per image. Considering the total number of tree rings, both datasets are comparable.

The INBD method relies on two important hyperparameters: the number of iterations at each epoch (n) and the image size factor (downsample). We seek the best-performing INBD model by exploring a grid with $n \in \{1, 2, 3, 4\}$ and $downsample \in \{0, 2, 4\}$. The model was trained using the UruDendro training and validation sets. The model that exhibits the best performance on the validation set was chosen. For training, we utilized the ClusterUy infrastructure, as described in ClusterUy \cite{clusterUy}, equipped with an Nvidia Tesla P100 GPU with 12GB of RAM. The hyperparameters $n=3$ and $downsample=0$ yield the best performance on the validation set.

\begin{table}[htbp]
\scriptsize
\centering
\caption{Comparative results between INBD and CS-TRD methods over the test set of images from the UruDendro dataset. To compare the execution times the same HW was used. $\uparrow$ ($\downarrow$) indicates higher (lower) values are better.}
\begin{tabular}{|c|c|c|c|c|c|}
\hline
\textbf{Method} & \textbf{P $\uparrow$}  & \textbf{R $\uparrow$} & \textbf{F $\uparrow$} & \textbf{RMSE (pixels) $\downarrow$} & \textbf{Time CPU (seconds) $\downarrow$ }\\ \hline
INBD            & 0.75       & 0.84       & 0.79       & 5.7           & \textbf{7.5}                                                                  \\ \hline
CS-TRD          & \textbf{0.94 }      & \textbf{0.88 }      & \textbf{0.91}       & \textbf{3.0}           & 18                                                                     \\ \hline
\end{tabular}
\label{tab:inbd}
\end{table}

Table \ref{tab:inbd} compares the performance of both methods using the test set from the UruDendro dataset. When using the same CPU hardware, we observe that the INBD  execution time is better, running at an average of 7.5 seconds compared to 18 seconds for CS-TRD. In terms of performance, CS-TRD outperforms the INBD model in precision, recall, and F-score. When considering the RMSE metric, CS-TRD  performs slightly better than INBD, with a difference of 2.7 pixels. As a point of reference, the RMSE difference between the human annotators on the UruDendro dataset is around 2.5 pixels.

\begin{figure*}[t]
\begin{subfigure}{\textwidth}
\begin{center}
   \begin{subfigure}{0.23\textwidth}
   \includegraphics[width=1\linewidth]{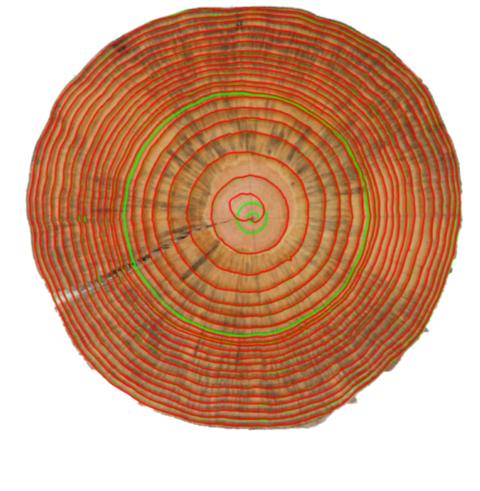}
   \caption{}
   \end{subfigure}   
   \begin{subfigure}{0.23\textwidth}
   \includegraphics[width=1\linewidth]{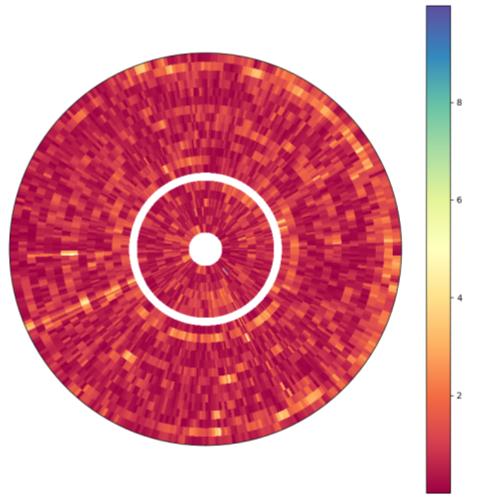}
   \caption{}
   \end{subfigure}
   \begin{subfigure}{0.23\textwidth}
   \includegraphics[width=1\linewidth]{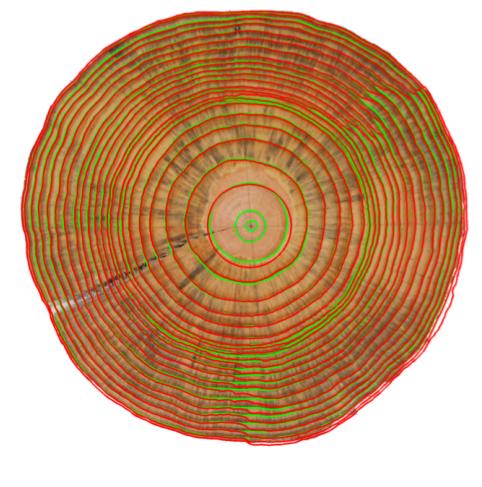}
   \caption{}
   \end{subfigure}   
   \begin{subfigure}{0.23\textwidth}
   \includegraphics[width=1\linewidth]{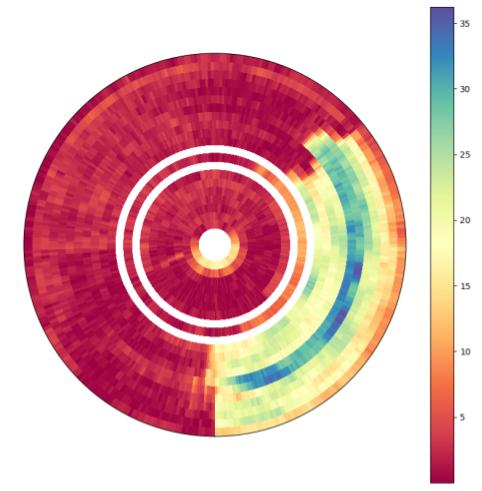}
   \caption{}
   \end{subfigure}
\end{center}
\end{subfigure}
\begin{subfigure}{\textwidth}
\begin{center}
    \begin{subfigure}{0.23\textwidth}
   \includegraphics[width=1\linewidth]{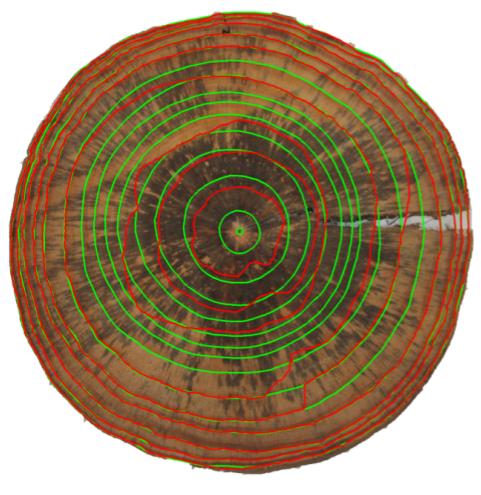}
  \caption{}
  \end{subfigure}   
  \begin{subfigure}{0.23\textwidth}
   \includegraphics[width=1\linewidth]{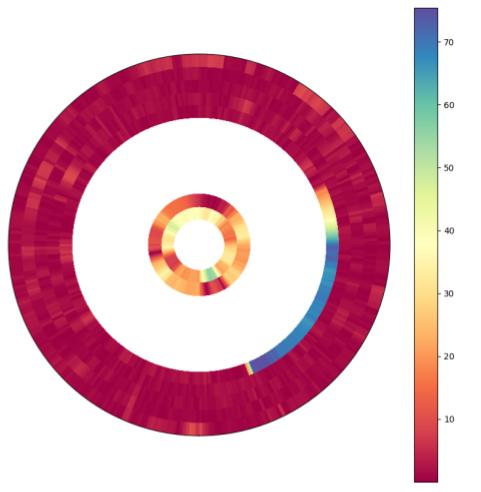}
   \caption{}
   \end{subfigure}
   \begin{subfigure}{0.23\textwidth}
   \includegraphics[width=1\linewidth]{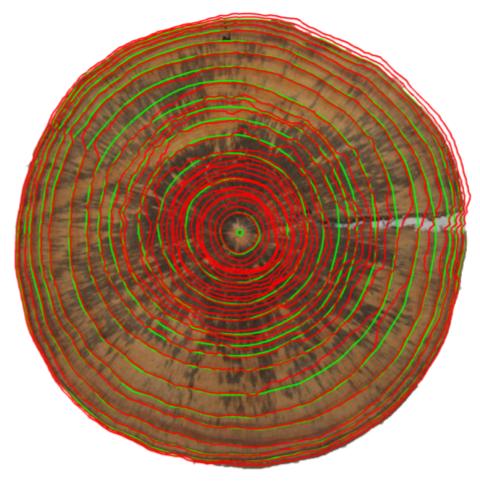}
  \caption{}
  \end{subfigure}   
  \begin{subfigure}{0.23\textwidth}
   \includegraphics[width=1\linewidth]{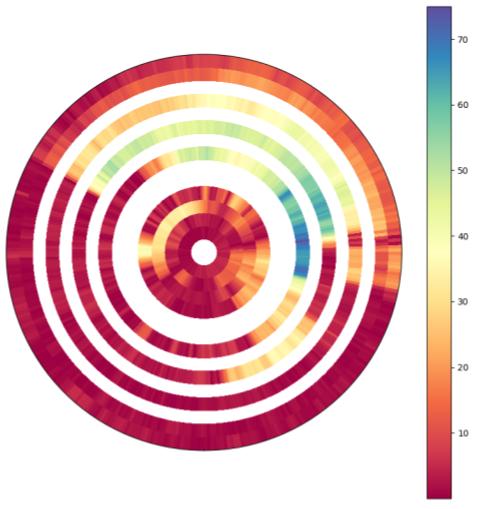}
   \caption{}
   \end{subfigure}
\end{center}
\end{subfigure}
\caption{Upper row, results of CS-TRD (left) and INBD (right) methods for disk F03b of UruDendro. In (a) and (c), the ring detections are shown in red, and the GT rings in green. In (b) and (d), the absolute radial error between the ring detections and the GT rings is displayed. (The red indicates a low error, and the blue indicates a high error.  A white band indicates no detection for a given ground truth ring. The results of the CS-TRD (left) and INBD (right) methods for disk L02b of UruDendro are shown in the lower row. In (e) and (g), the ring detections are shown in red, and the GT rings in green. In (f) and (h), the absolute radial error between the ring detections and the GT rings is displayed (read indicates a low error, and blue indicates a high error. A white band indicates no detection for a given ground truth ring.}
   \label{fig:CSTRDvsINBD}
\end{figure*}

\Cref{fig:CSTRDvsINBD} illustrates a qualitative comparison between both methods for two disks of the UruDendro dataset. Subfigures a, c, e, and g superpose the detected rings in red and the ground truth rings in green, while subfigures b, d, f, and h depict the absolute radial error between the automatic detections and the GT disks. In the upper row, the results for disk F03b are shown. CS-TRD produced 20 True Positives, 1 False Positive, and 3 False Negatives, compared to 19 True Positives, 3 False Positives, and 4 False Negatives for INBD. The INBD method is iterative, and an error produced in a certain ring is propagated to the next rings outward, as seen in subfigure d. This significantly increases the RMSE error (9.2 for the INBD and 1.1 for the CS-TRD). The lower row of \Cref{fig:CSTRDvsINBD} illustrates the results for a disk with a high amount of fungus stains (L02b), which produces an important number of false detections (20) by the INBD method.

\section{Conclusions and future work}
\label{sec:conclusions}

We presented an automatic method for Tree Ring Detection of cross-section wood images. It achieves an F-Score of 97\% in the Kennel dataset and 89\% in the more complex UruDendro dataset. The CS-TRD method outperforms deep learning state-of-the-art methods such as the INBD \cite{inbd} in the Pinus taeda species. It performs well even in the presence of fungus, cracks, and knots and in two different species (Abies alba and Pinus taeda).

The method runs at an average execution time of 17 seconds in the UruDendro dataset and 11 seconds in the Kennel dataset on an Intel Core i5 10300H workstation with 16 GB of RAM (without GPU). Compared with the time each annotator needs to manually delineate every disk, 3 hours on average, is a vast improvement. CS-TRD can be fully implemented in C++ to accelerate the execution time compared to the Python implementation\footnote{https://medium.com/agents-and-robots/the-bitter-truth-python-3-11-vs-cython-vs-c-performance-for-simulations-babc85cdfef5}. This will allow using the method in real-time applications.

In the future, we plan to include the  automatic pith detection, 
 extend the method to other tree species, and explore machine-learning techniques to improve the results.

\section*{Image Credits}
\includegraphics[height=2em]{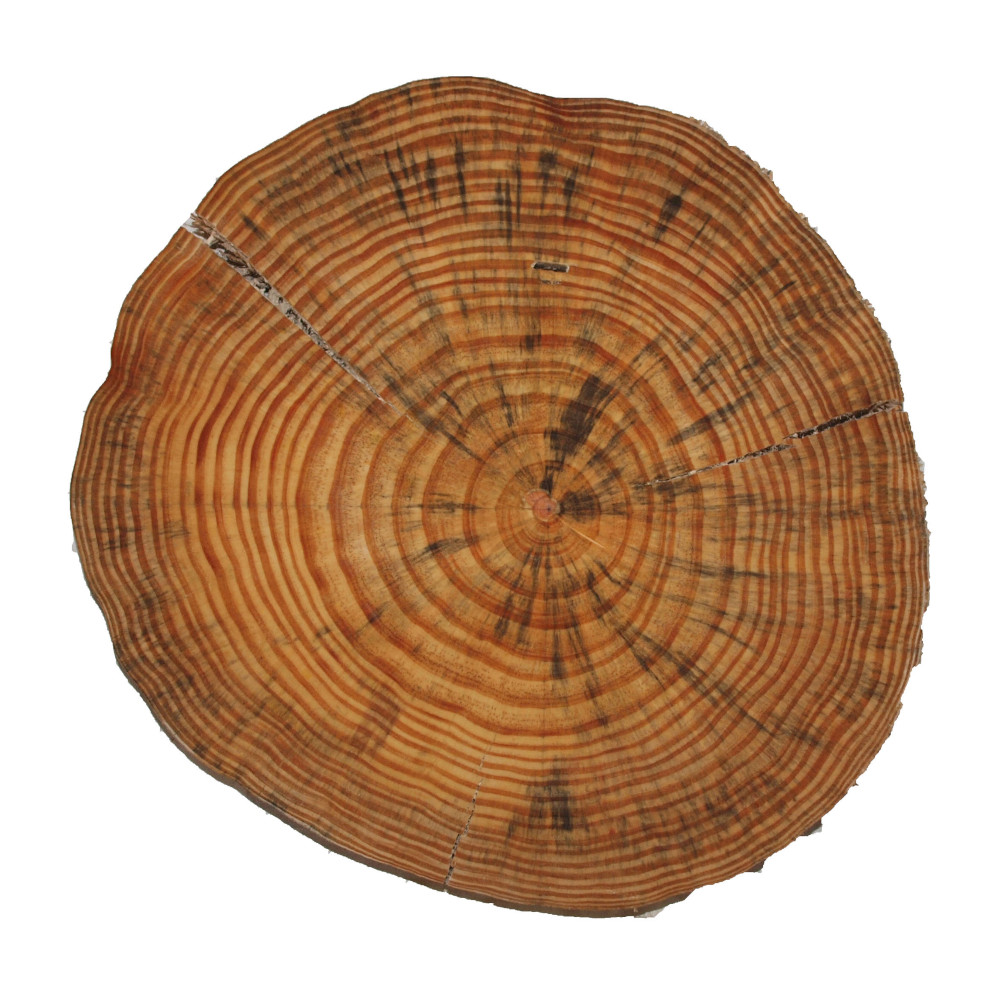}
\includegraphics[height=2em]{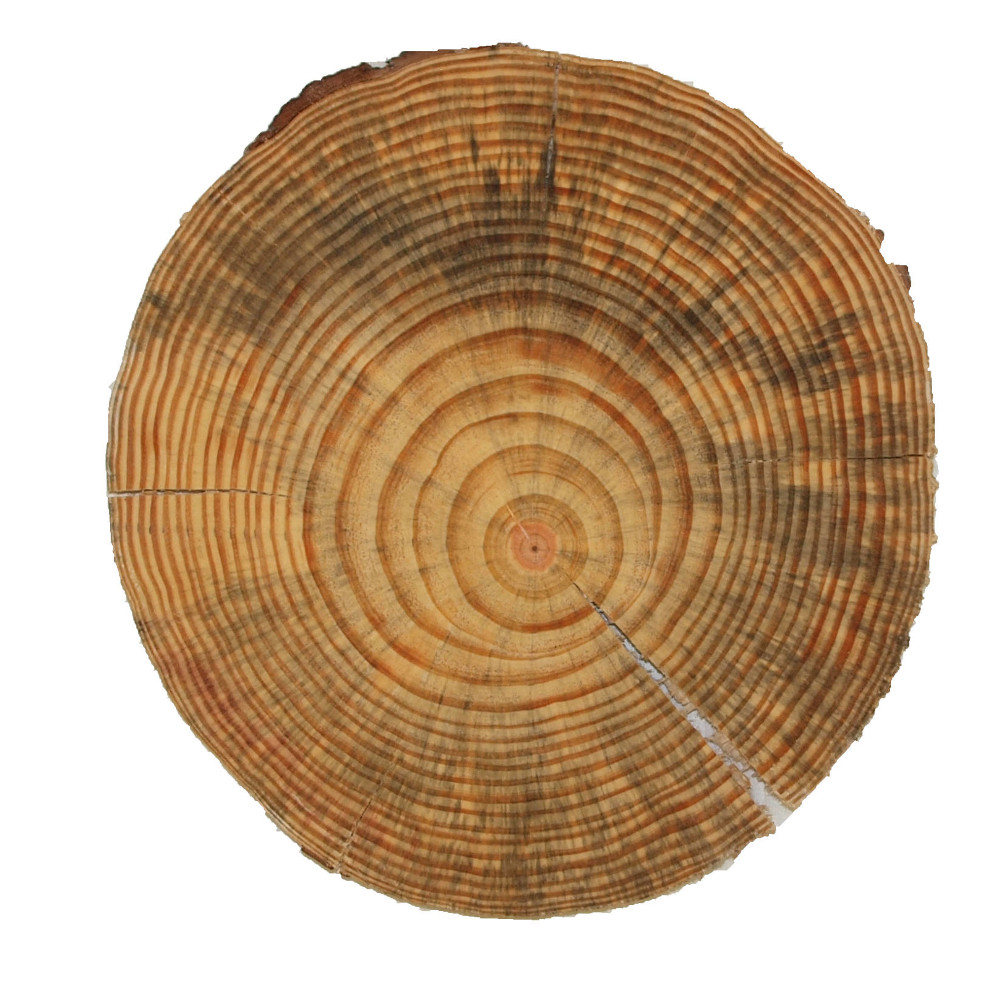}
\includegraphics[height=2em]{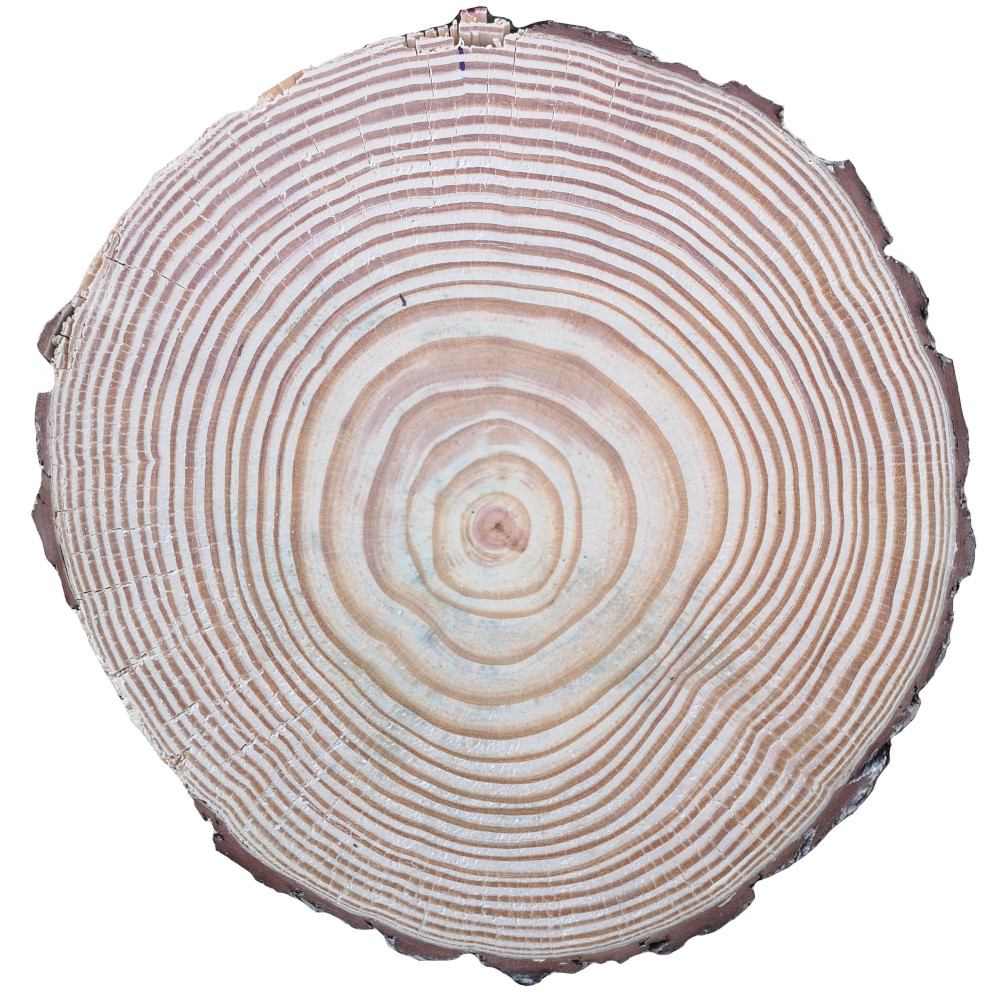}
\includegraphics[height=2em]{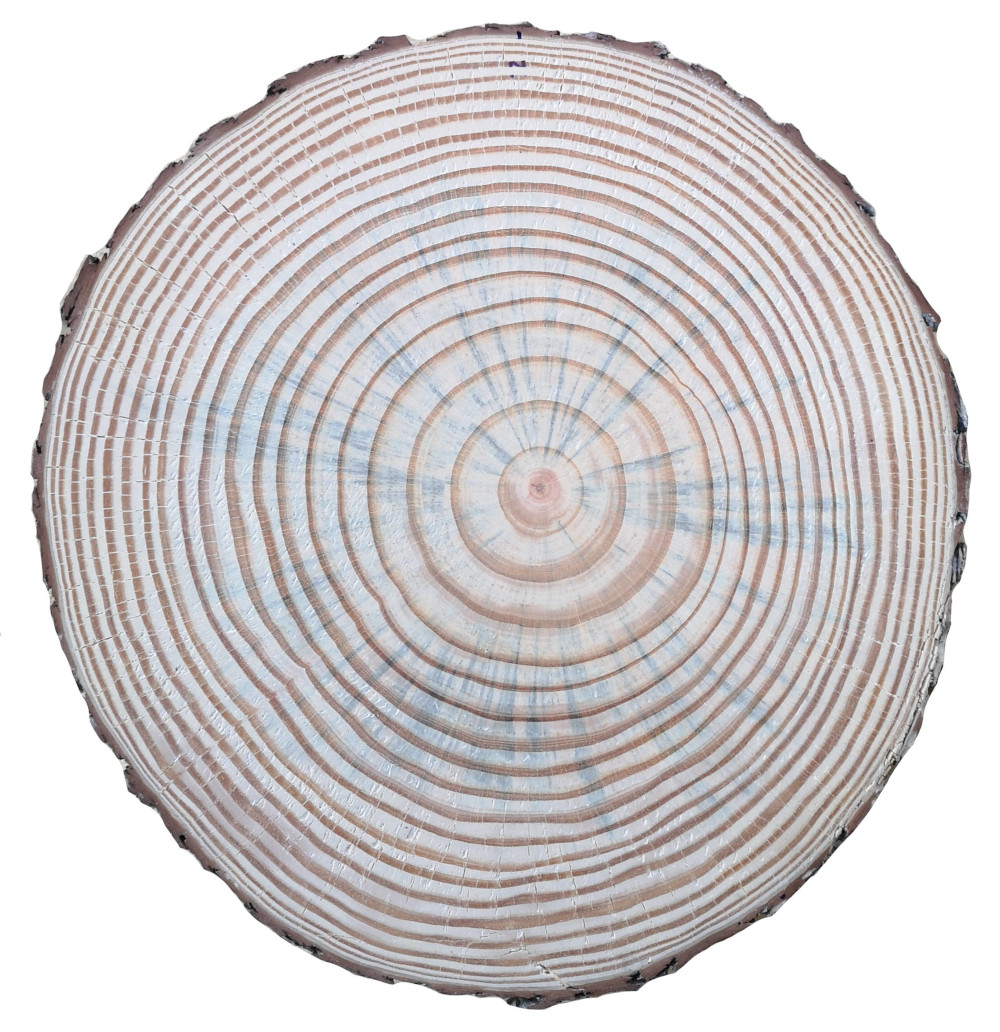}
\includegraphics[height=2em]{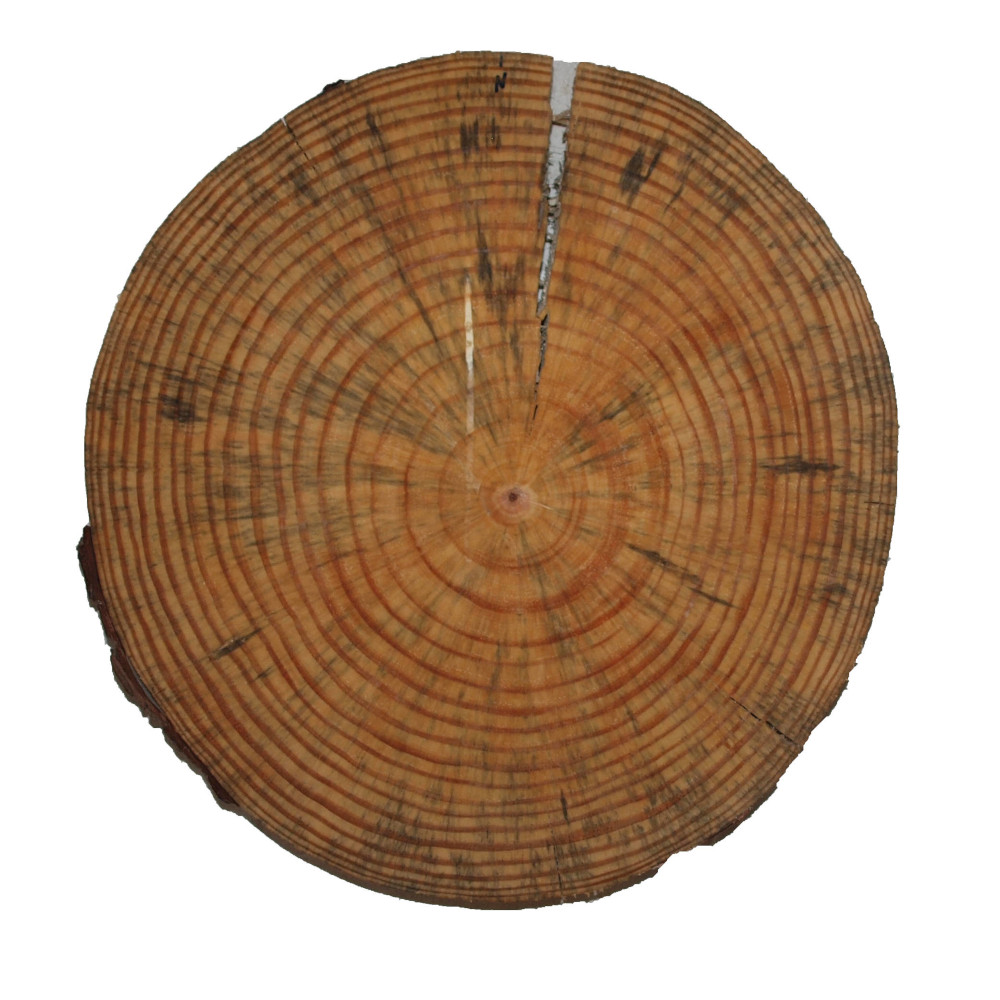}
\includegraphics[height=2em]{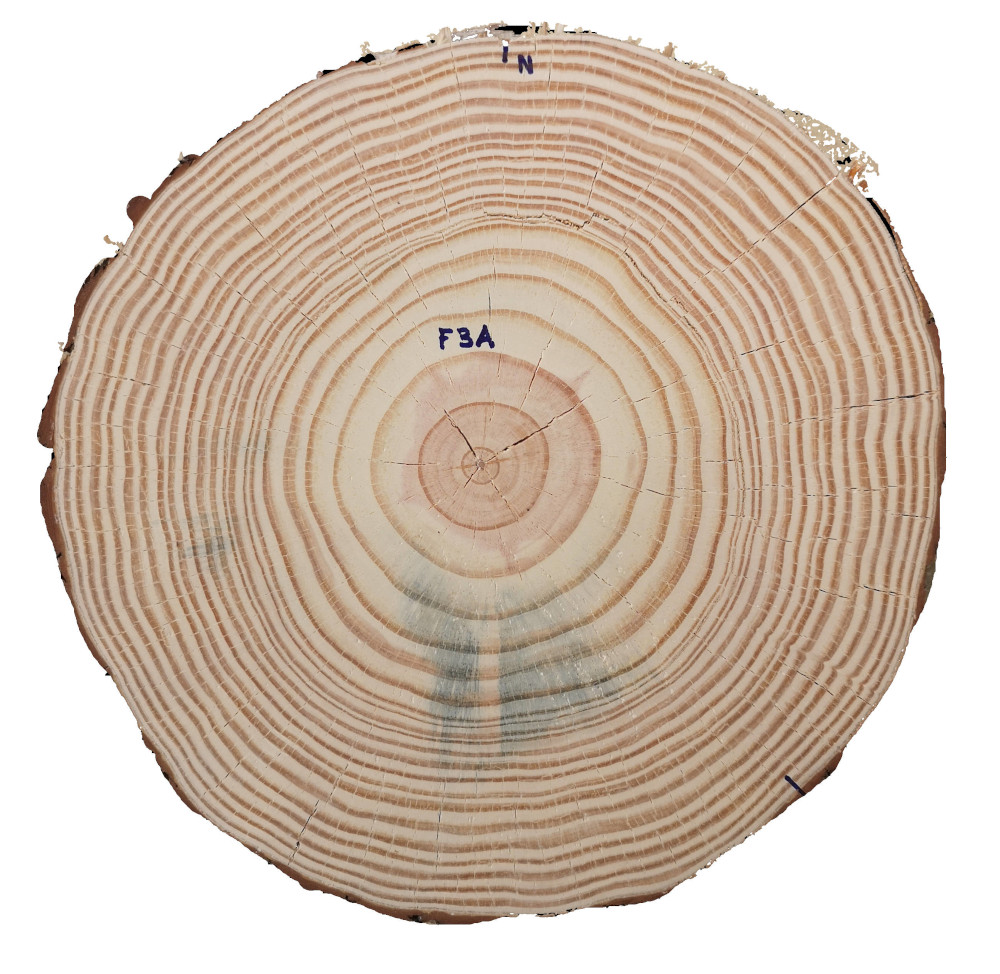}
\includegraphics[height=2em]{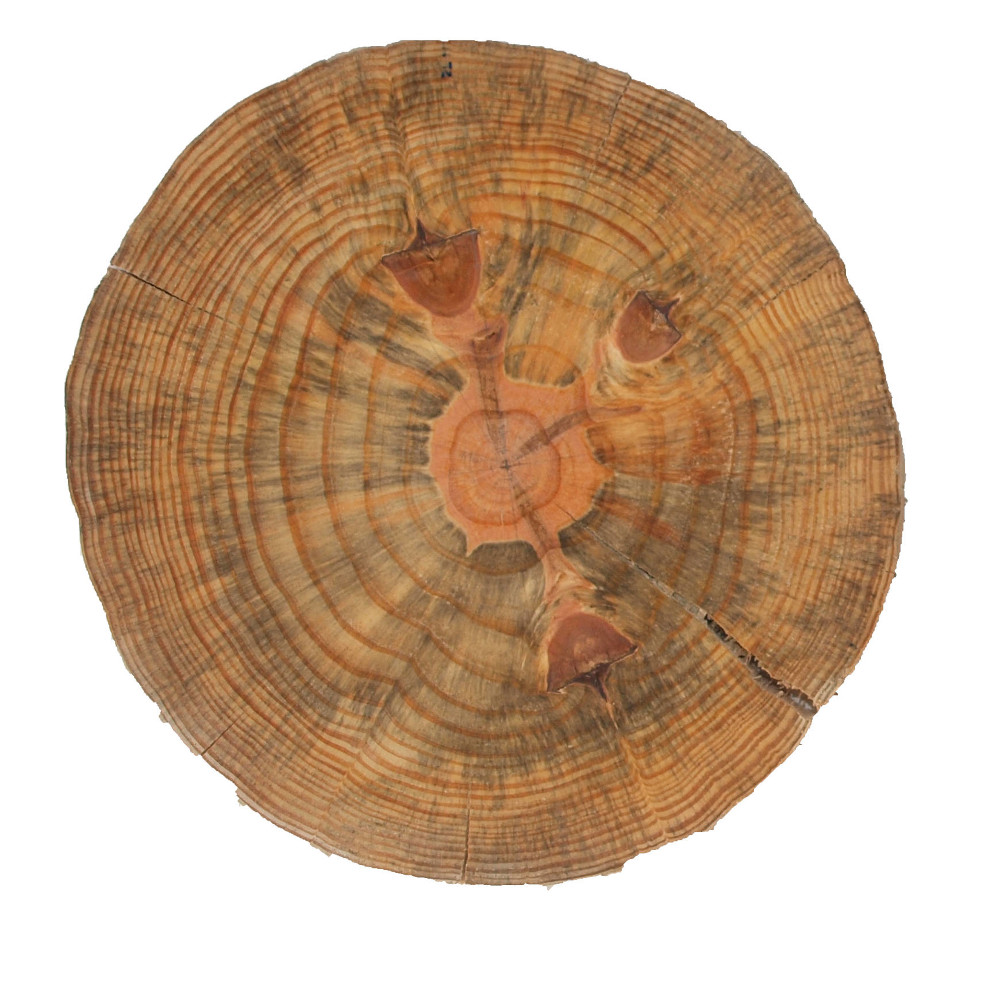}
\includegraphics[height=2em]{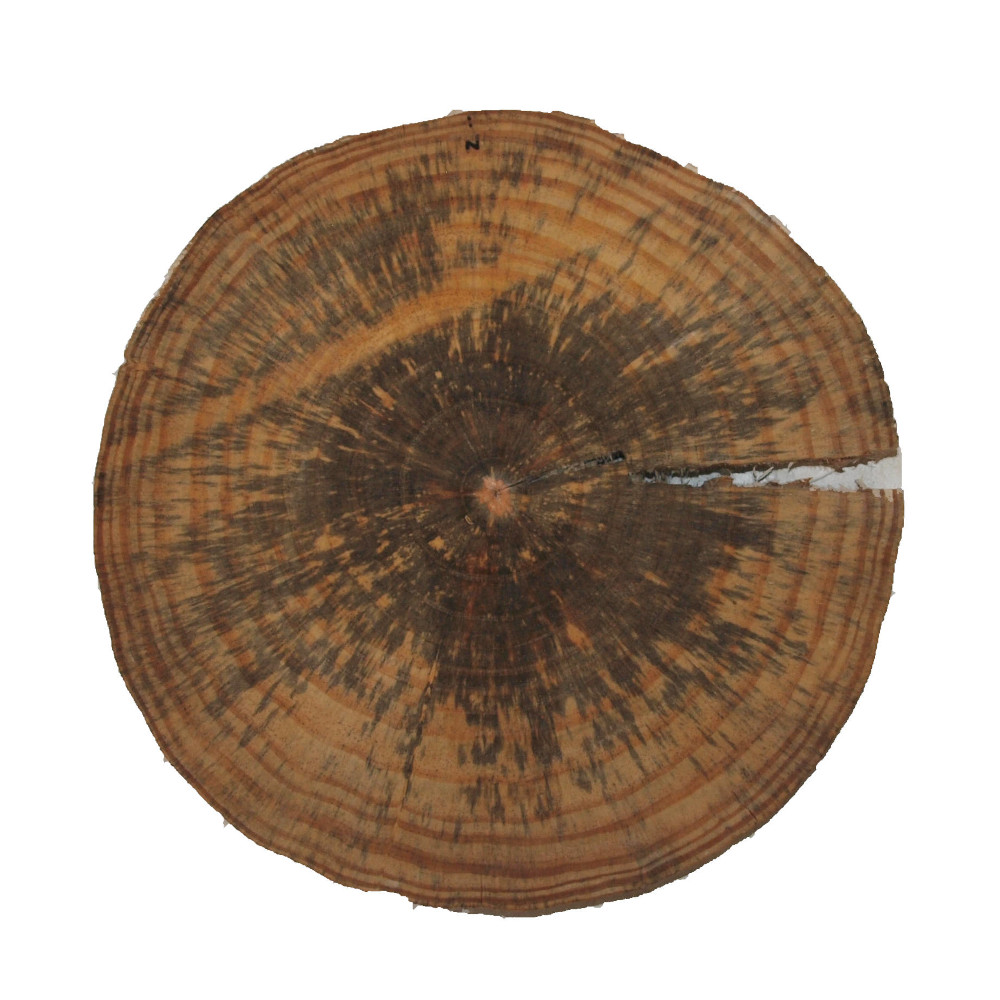}
\includegraphics[height=2em]{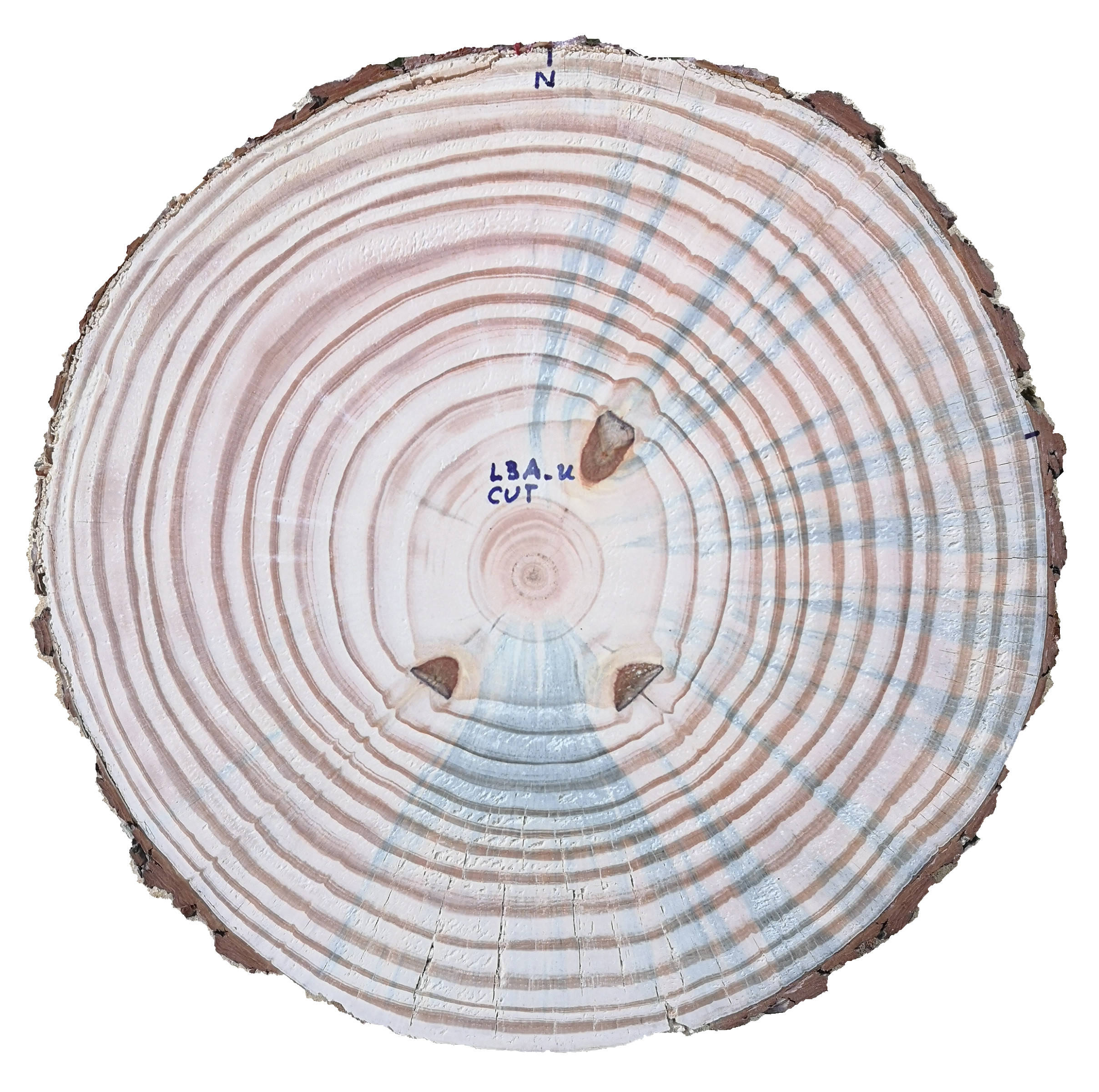}
\includegraphics[height=2em]{F03d_segmentation.jpg}
Images from the UruDendro dataset.
\\
\includegraphics[height=0.5em]{picea_04r.png}
\includegraphics[height=0.5em]{larix_19r.png}
\includegraphics[height=0.5em]{abies_04r.png}
Images taken from a \cite{FABIJANSKA2017279}
\\
\includegraphics[height=2em]{ac1_output.jpg}
\includegraphics[height=2em]{ac2_output.jpg}
\includegraphics[height=2em]{ac3_output.jpg}
original images from the Kennel dataset.


{\small
\bibliographystyle{siam}
\bibliography{TreeRingDetection}
}

\end{document}


\maketitle 

\section{Showcasing CS-TRD examples}
\subsection{UruDendro}

Some other examples of the ring delineation in the dataset are illustrated in~\Cref{fig:ddbb_ac_output_sup}. Over these sets of disks, the results are really good. The CS-TRD algorithm generally works fine, even if it has problems with some images. Let's discuss several examples, such as images L02b, F07e, and L02d.

\begin{figure*}
\begin{centering}
    \begin{subfigure}{0.3\textwidth}
    \begin{centering}
    \includegraphics[width=\textwidth]{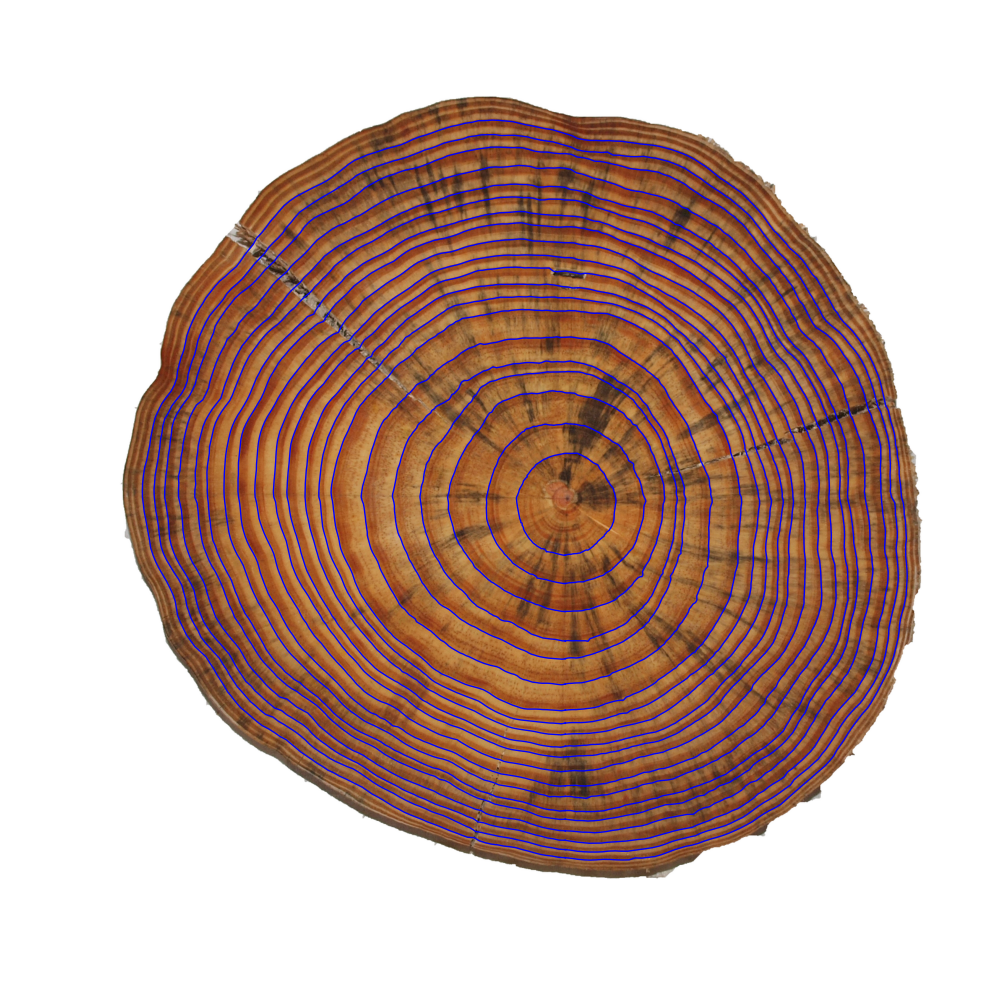}
    \label{fig:ddbb-F02a_output}
    \caption{F02a}
    \end{centering}
    \end{subfigure}
    \hfill
    \begin{subfigure}{0.3\textwidth}
    \begin{centering}
    \includegraphics[width=\textwidth]{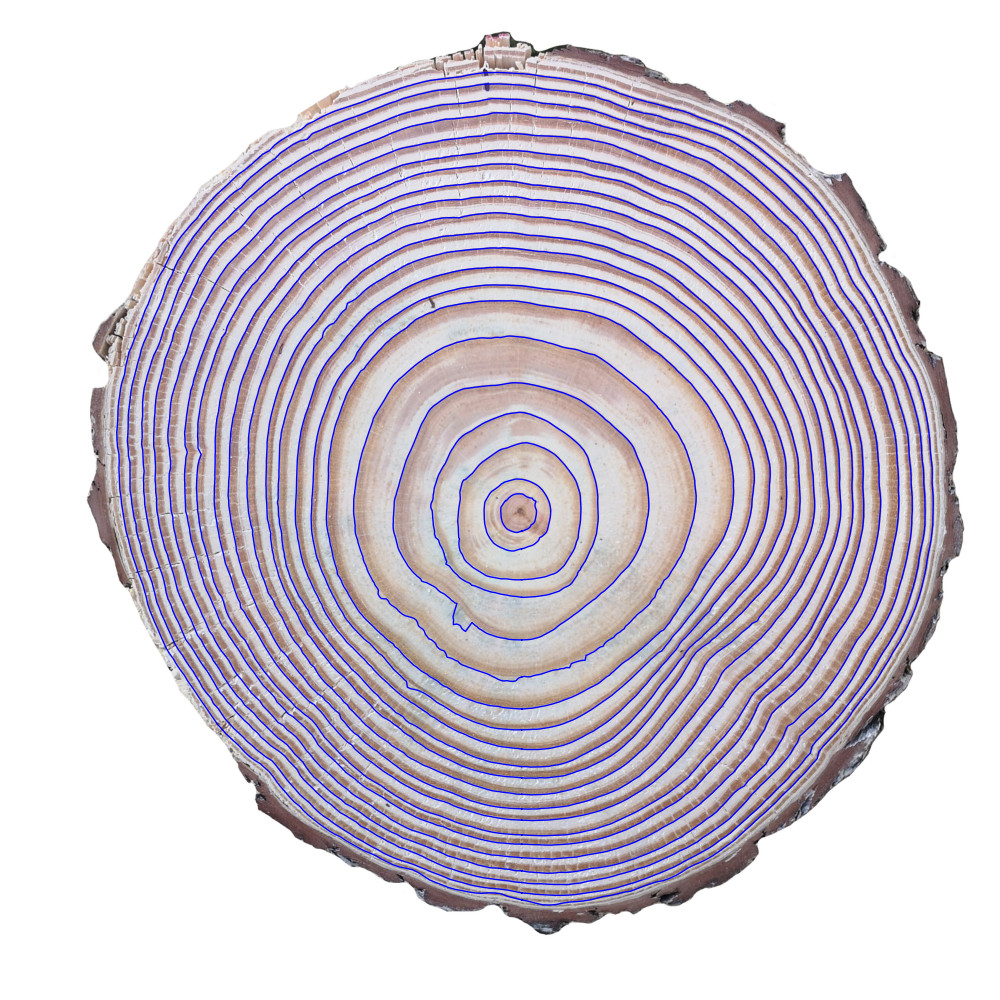}
    \label{fig:ddbb-F02c_output_output}
    \caption{F02c}
    \end{centering}
    \end{subfigure}
    \hfill
    \begin{subfigure}{0.3\textwidth}
    \begin{centering}
   \includegraphics[width=\textwidth]{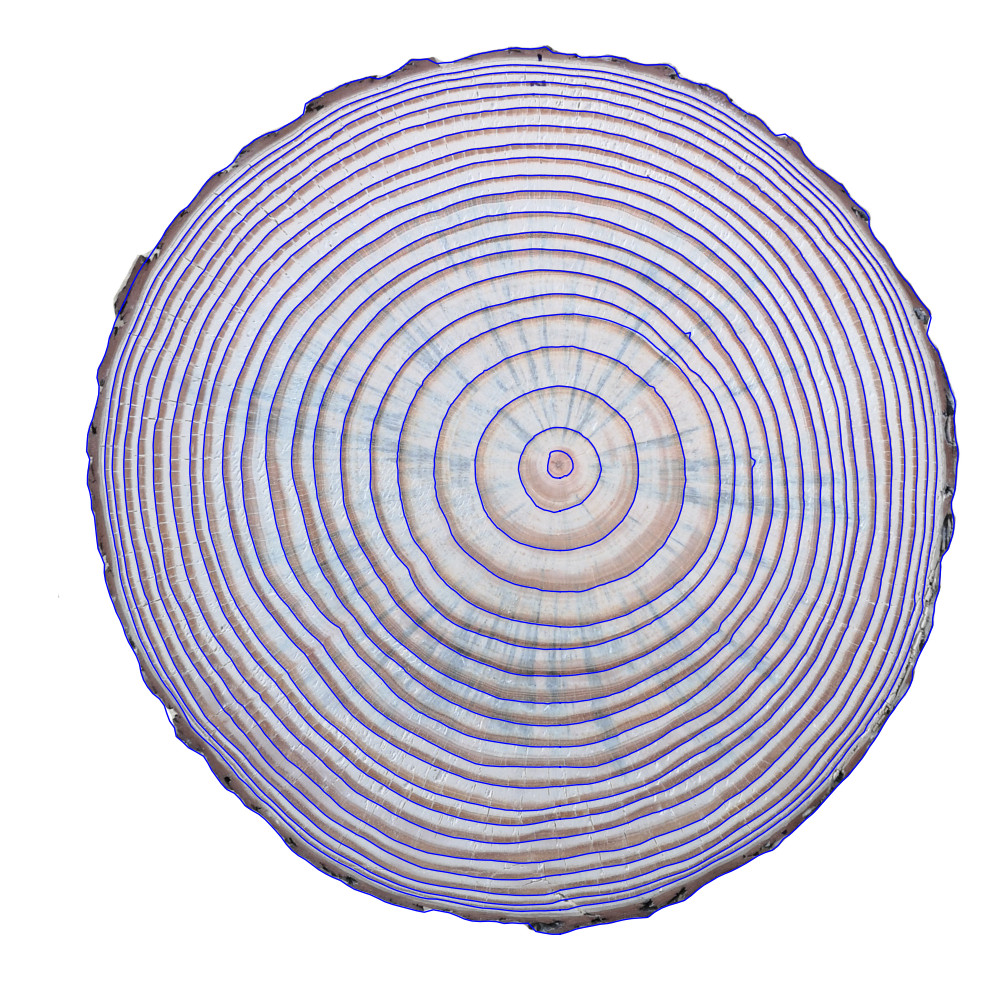}
    \label{fig:ddbb-F02d_output}
    \caption{F02d}
    \end{centering}
    \end{subfigure}
    \hfill
    \begin{subfigure}{0.3\textwidth}
    \begin{centering}
   \includegraphics[width=\textwidth]{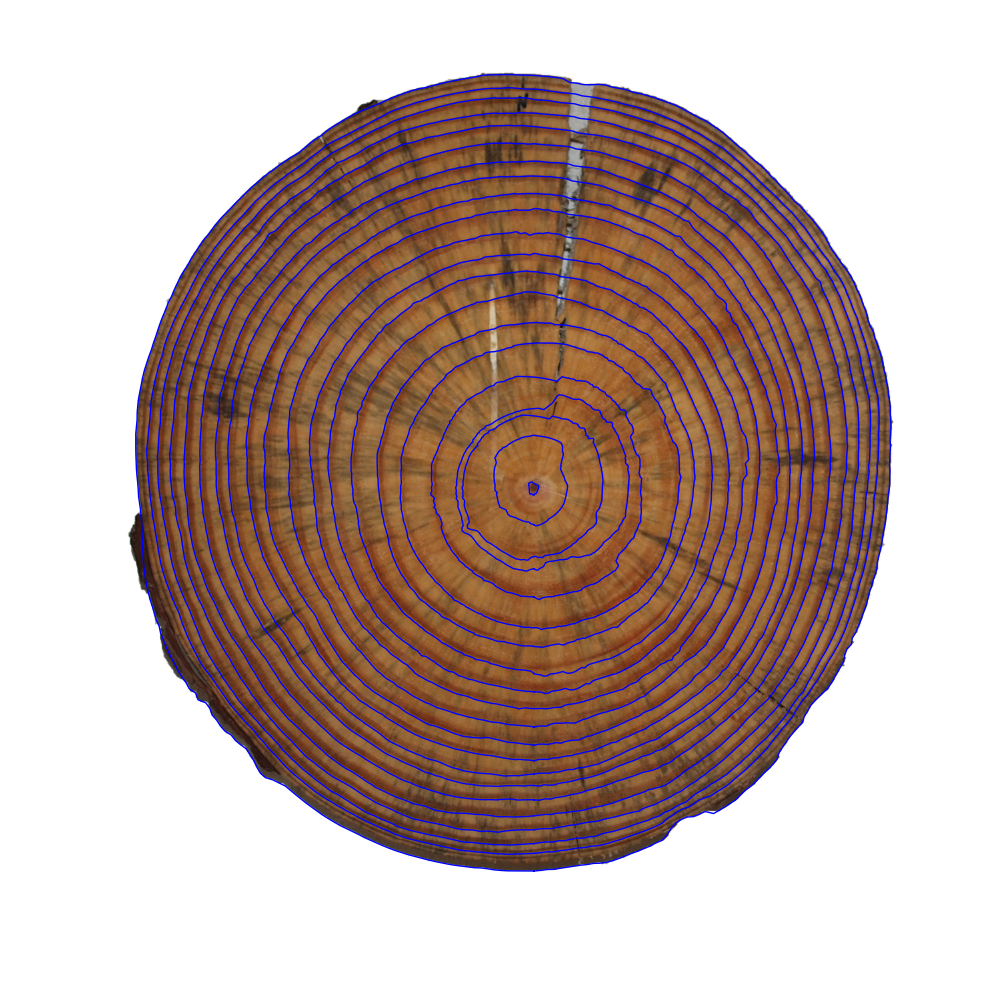}
    \label{fig:ddbb-F02e_output}
    \caption{F02e}
    \end{centering}
    \end{subfigure}
    \hfill
    \begin{subfigure}{0.3\textwidth}
    \begin{centering}
   \includegraphics[width=\textwidth]{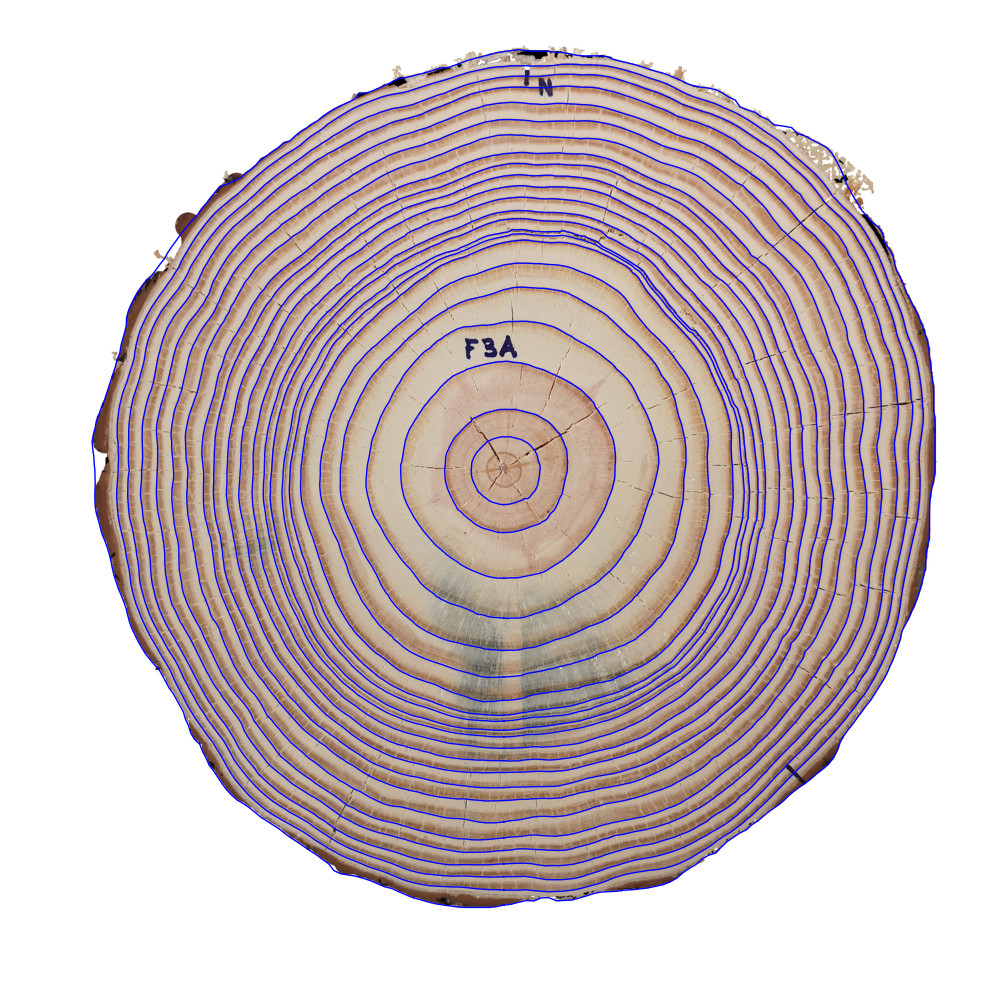}
    \label{fig:ddbb-F03c_output}
    \caption{F03c}
    \end{centering}
    \end{subfigure}
    \hfill
    \begin{subfigure}{0.3\textwidth}
    \begin{centering}
   \includegraphics[width=\textwidth]{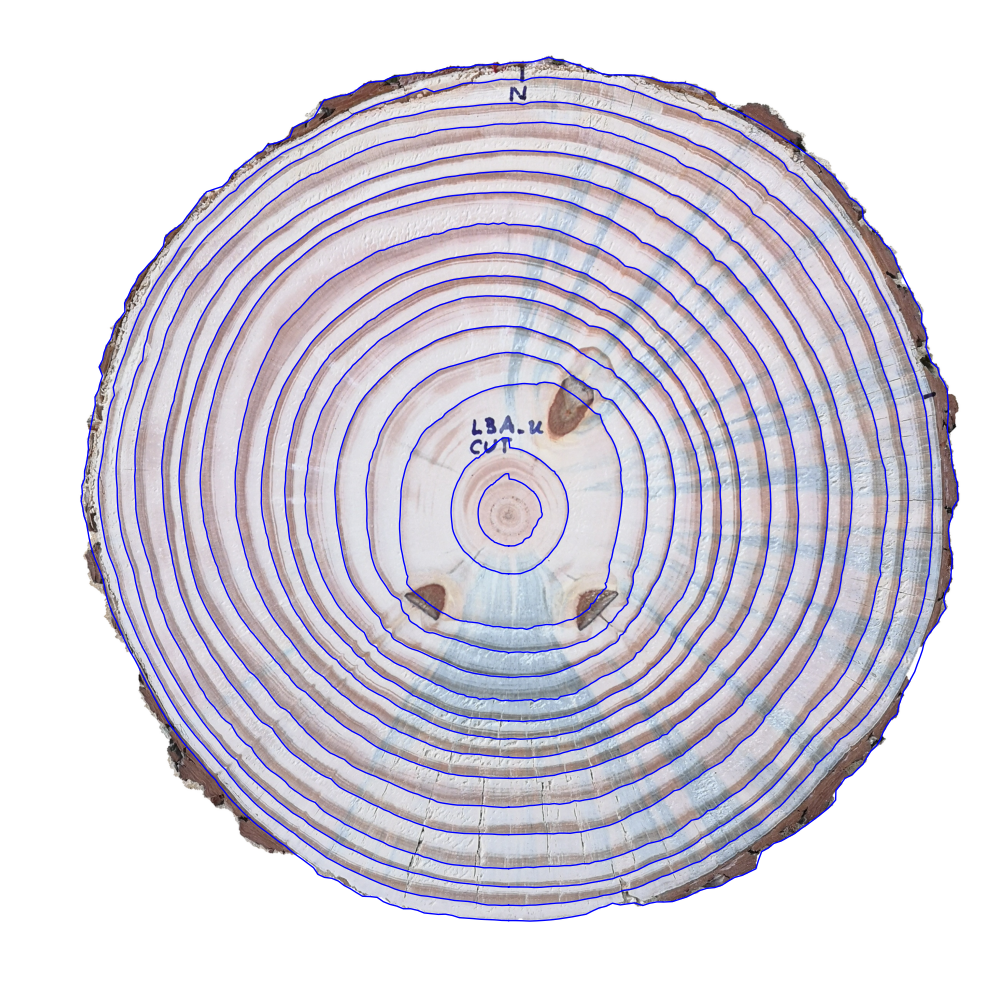}
    \label{fig:ddbb-L03c_output}
    \caption{L03c}
    \end{centering}
    \end{subfigure}
    \hfill
   \caption{Some results for the UruDendro dataset.}
   \label{fig:ddbb_output_sup}
\end{centering}
\end{figure*}

\begin{figure*}
\begin{centering}
\begin{subfigure}{0.3\textwidth}
\begin{centering}
\includegraphics[width=\linewidth]{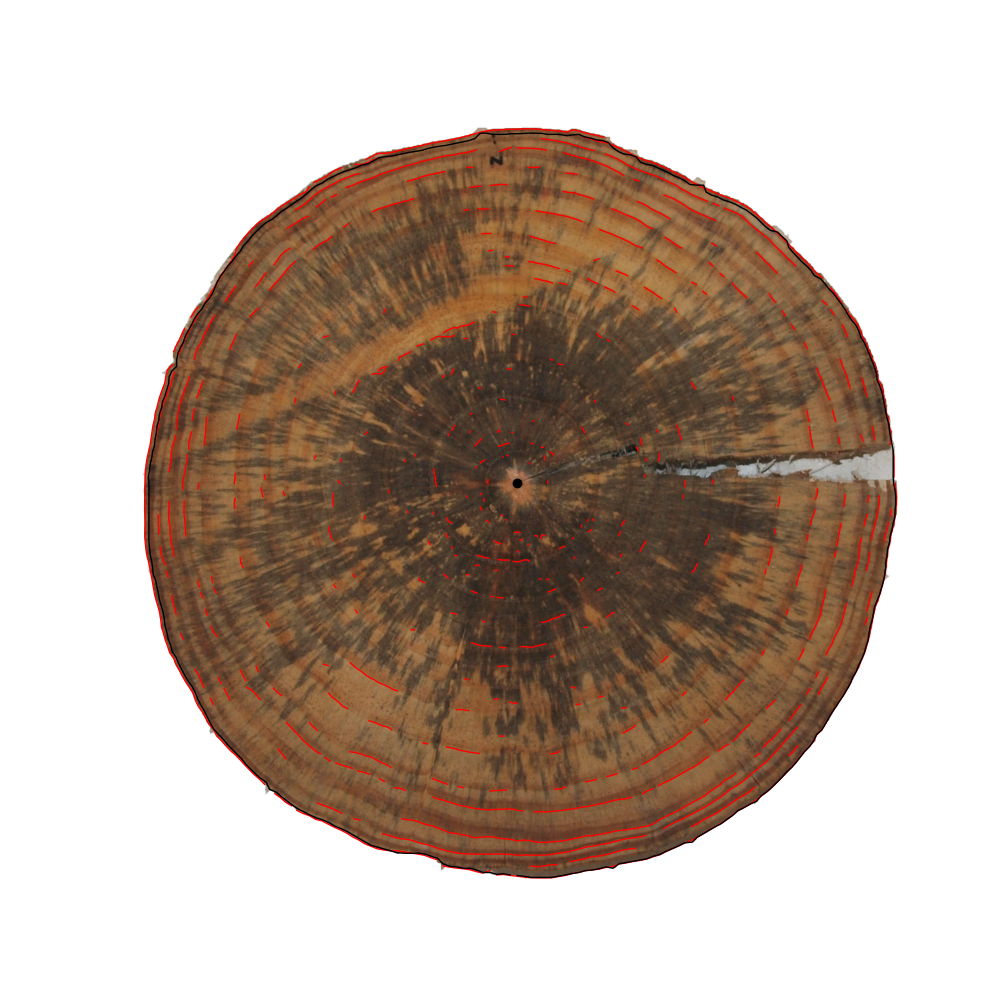}
\caption{Chains}
\label{fig:L02b_chains}
\end{centering}
\end{subfigure}
\begin{subfigure}{0.3\textwidth}
\begin{centering}
\includegraphics[width=\linewidth]{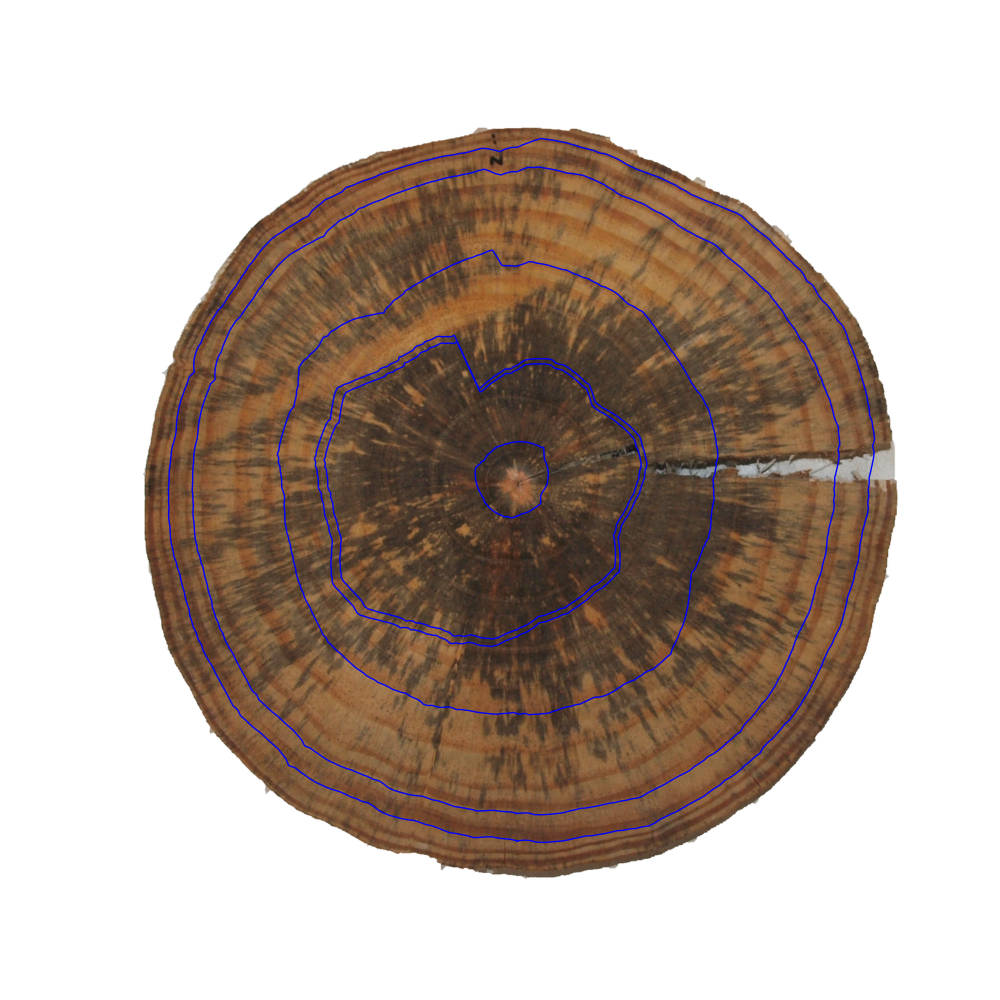}
\caption{Output}
\label{fig:L02b_output}
\end{centering}
\end{subfigure}
\begin{subfigure}{0.35\textwidth}
\begin{centering}
\includegraphics[width=\linewidth]{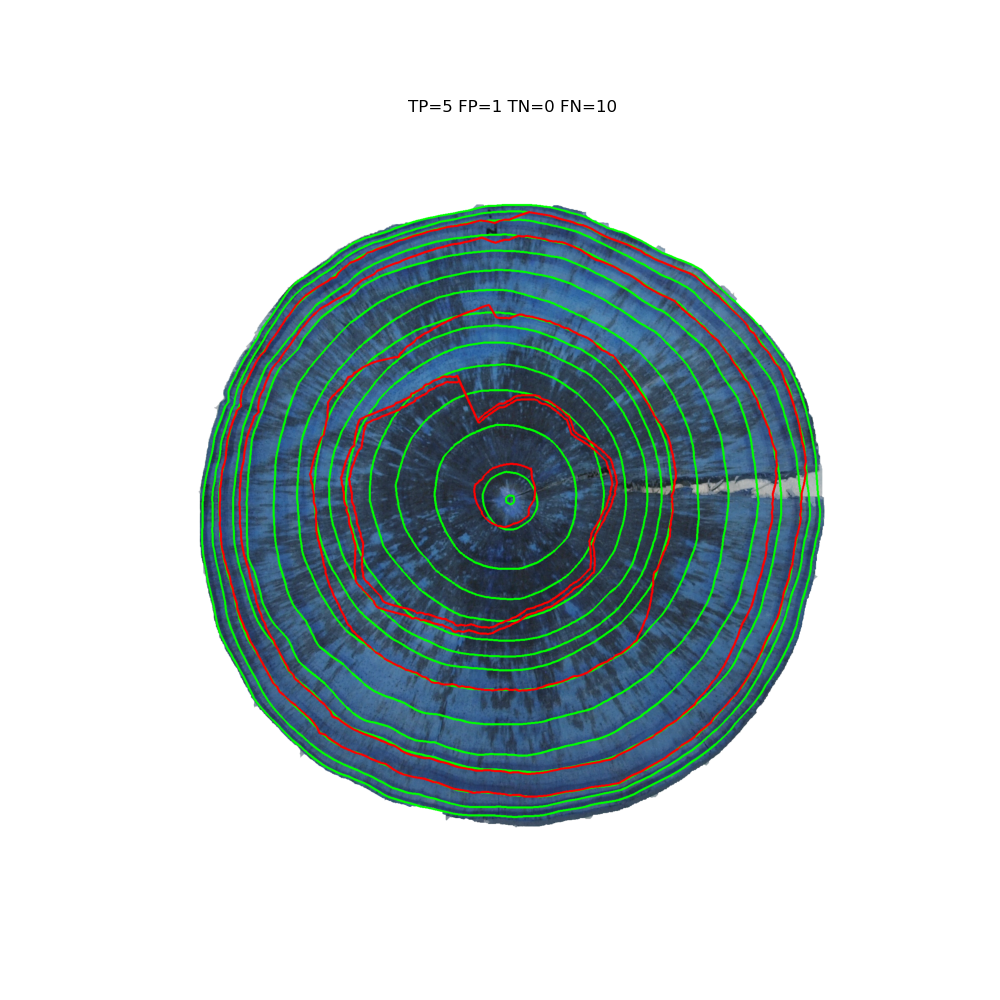}
\caption{Gt and Dt}
\label{fig:L02b_gt_and_dt}
\end{centering}
\end{subfigure}
\caption{Method result for disk L02b. Note how the fungus stain perturbs the edge detection step.}
\label{fig:L02b_analysis}
\end{centering}
\end{figure*}

\Cref{fig:L02b_analysis} illustrates the results for disk L02b.  \Cref{fig:L02b_gt_and_dt}, shows the detected rings in red and the GT in green. Four detections are closed curves and determined as correct (TP), while two are determined as incorrect (FP). Counting from the center to the border, the first detection is correct, and the next two are bad, corresponding to the second and third rings. Analyzing the chains step output shown in \Cref{fig:L02b_chains}, it seems clear that there is not enough edge information to see the rings due to the fungus stain. 

\begin{figure*}
\begin{centering}
\begin{subfigure}{0.3\textwidth}
\begin{centering}
\includegraphics[width=\linewidth]{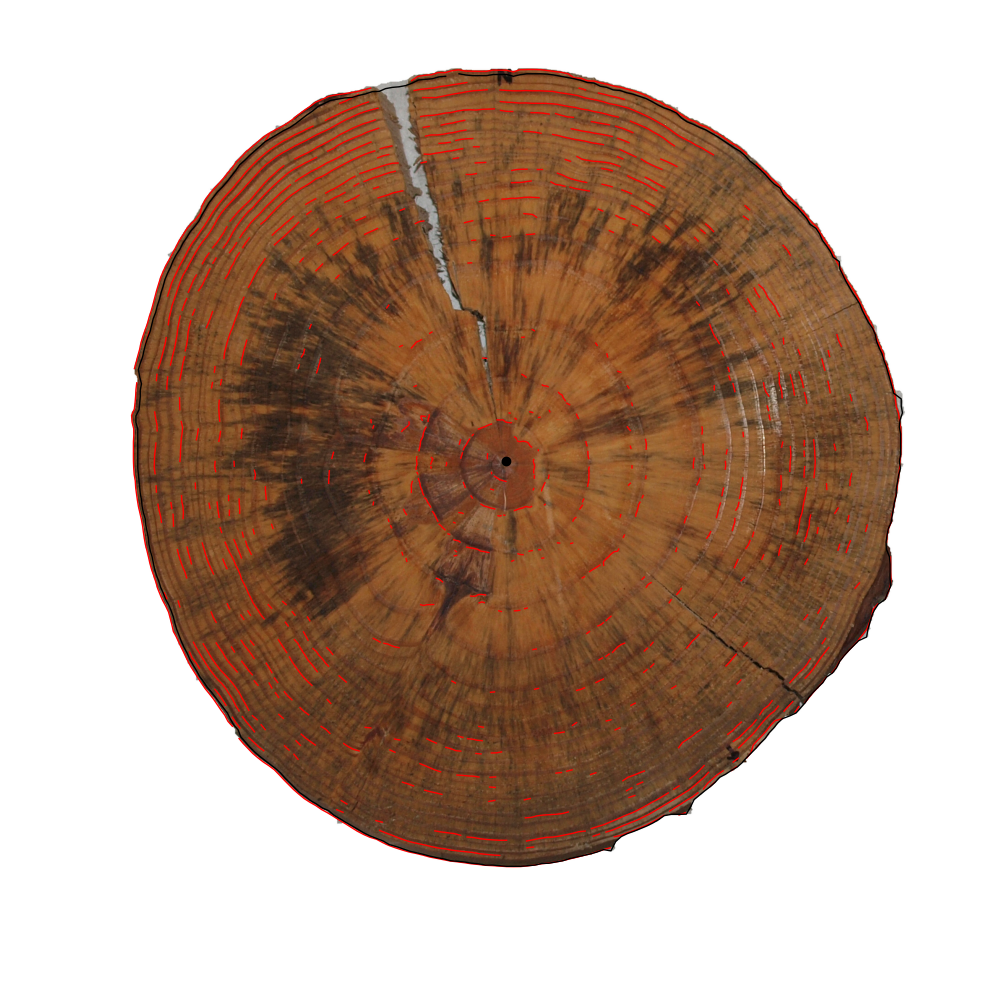}
\caption{Chains}
\label{fig:F07e_chains}
\end{centering}
\end{subfigure}
\begin{subfigure}{0.3\textwidth}
\begin{centering}
\includegraphics[width=\linewidth]{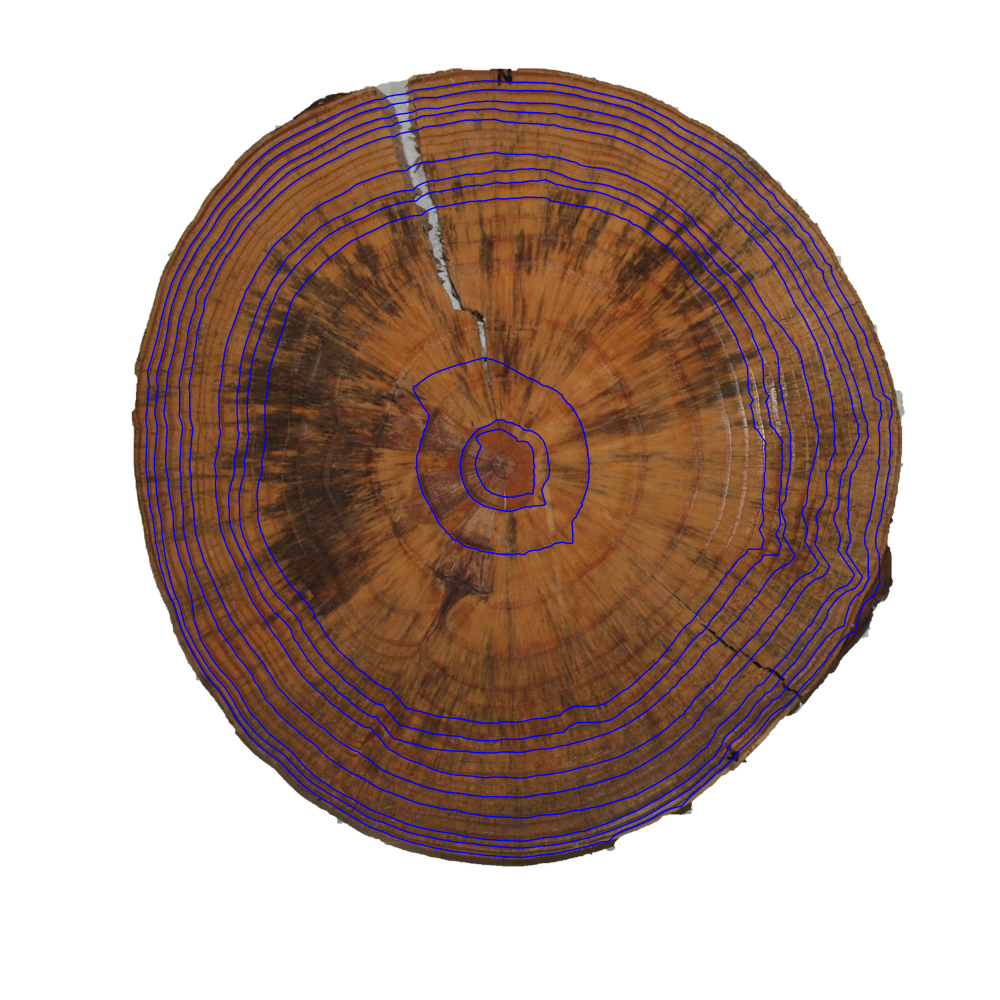}
\caption{Output}
\label{fig:F07e_output}
\end{centering}
\end{subfigure}
\begin{subfigure}{0.35\textwidth}
\begin{centering}
\includegraphics[width=\linewidth]{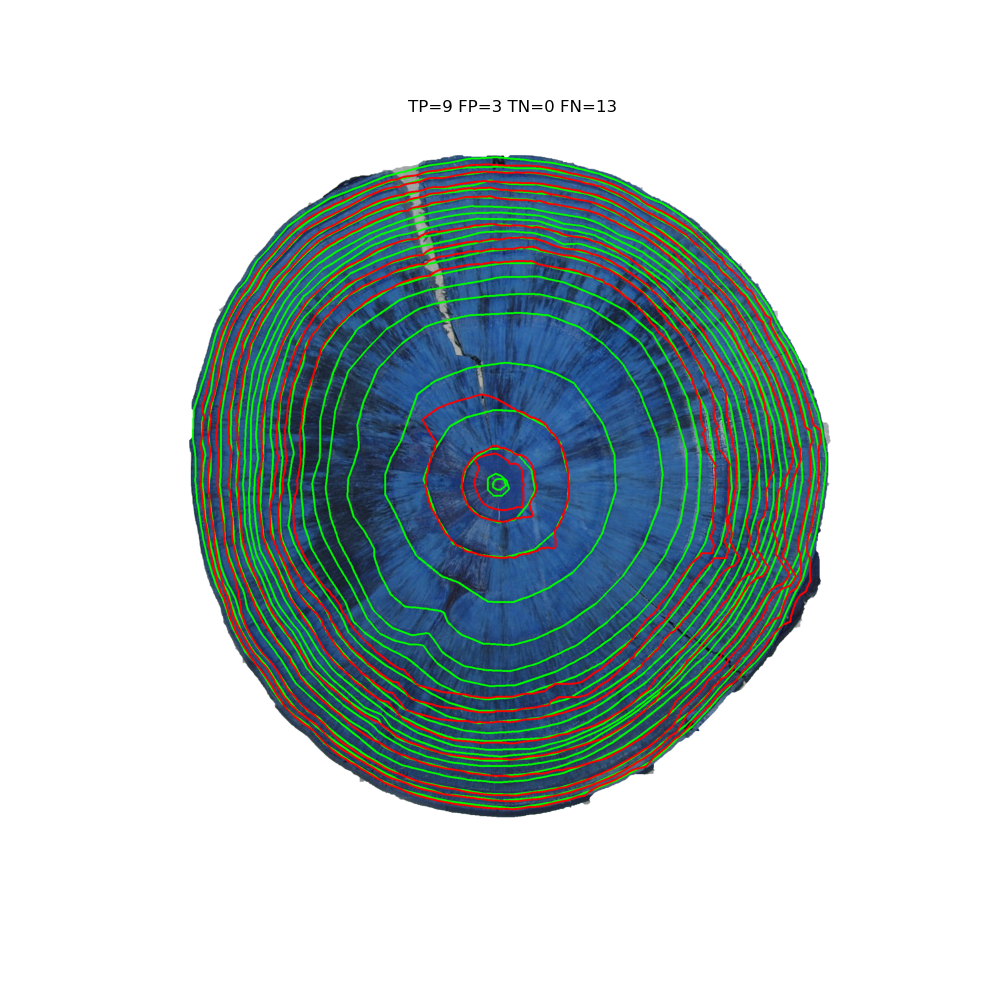}
\caption{Gt and Dt}
\label{fig:F07e_gt_and_dt}
\end{centering}
\end{subfigure}
\caption{Method result for disk F07e. Note how the fungus stain perturbs the edge detection step.}
\label{fig:F07e_analysis}
\end{centering}
\end{figure*}

\begin{figure*}
\begin{centering}
\begin{subfigure}{0.3\textwidth}
\begin{centering}
\includegraphics[width=\linewidth]{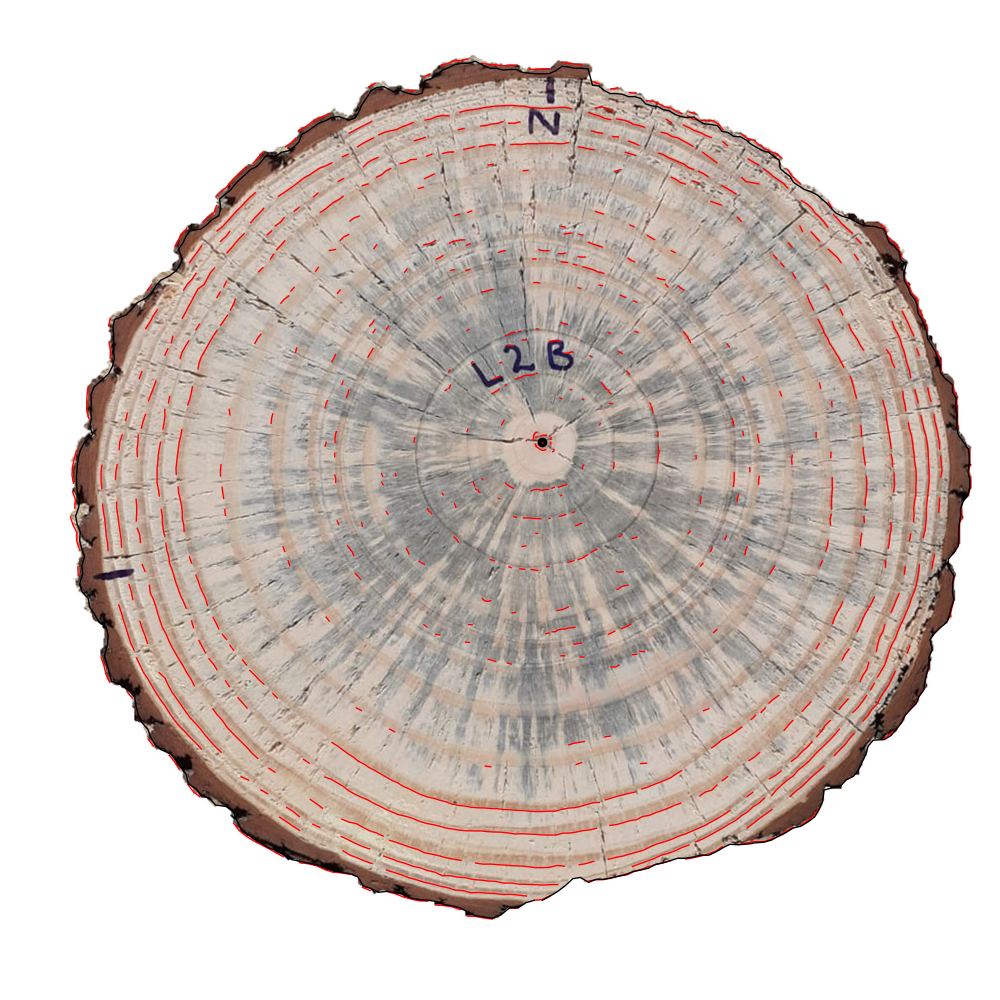}
\caption{Chains}
\label{fig:L02d_chains}
\end{centering}
\end{subfigure}
\begin{subfigure}{0.3\textwidth}
\begin{centering}
\includegraphics[width=\linewidth]{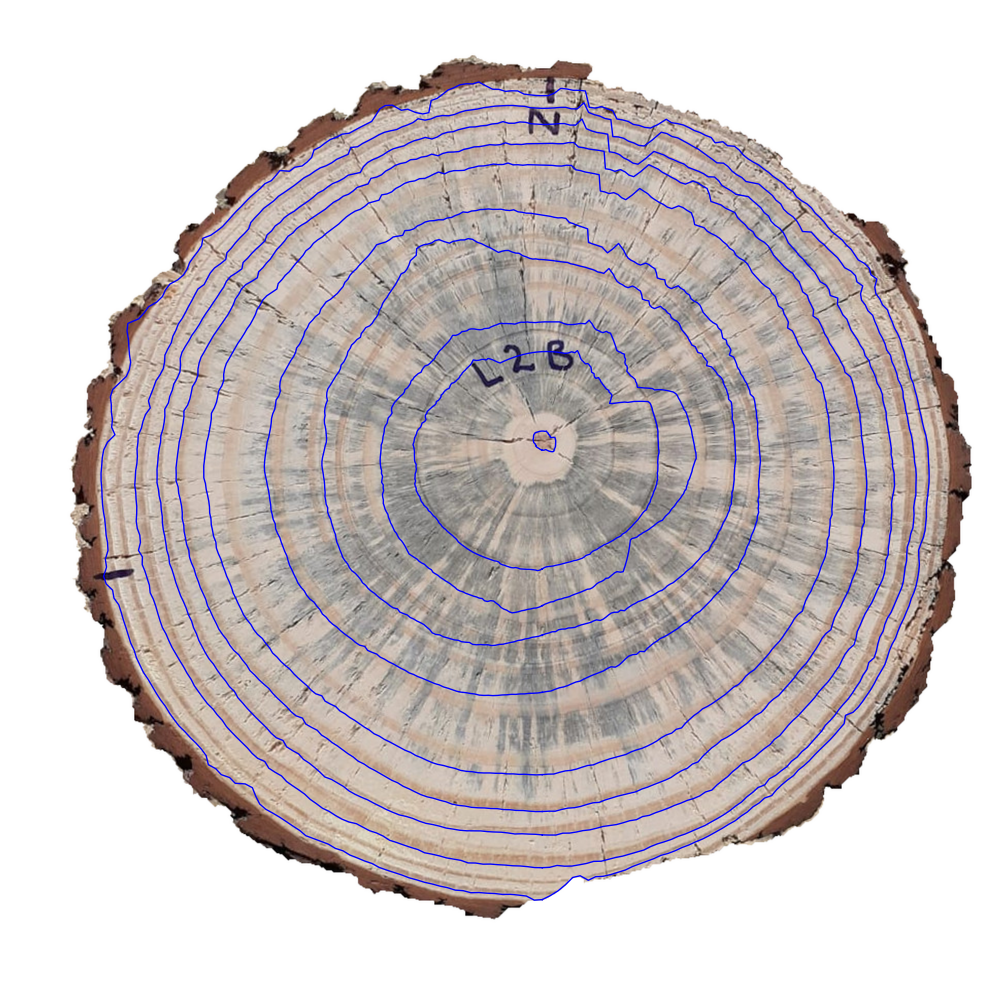}
\caption{Output}
\label{fig:L02d_output}
\end{centering}
\end{subfigure}
\begin{subfigure}{0.35\textwidth}
\begin{centering}
\includegraphics[width=\linewidth]{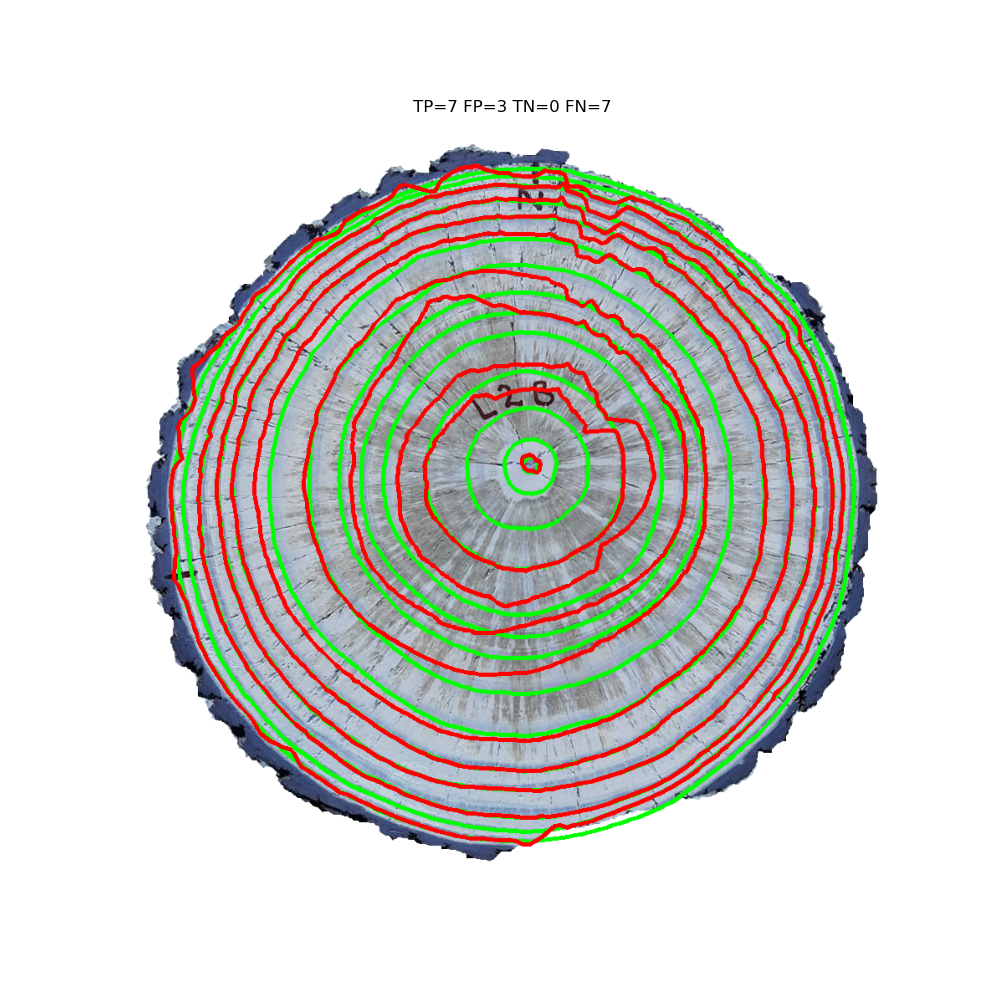}
\caption{Gt and Dt}
\label{fig:L02d_gt_and_dt}
\end{centering}
\end{subfigure}
\caption{Method result for disk L02d. Note how the fungus stain perturbs the edge detection step.}
\label{fig:L02d_analysis}
\end{centering}
\end{figure*}

A similar situation happens for disk F07e, \Cref{fig:F07e_analysis}. There is a strong fungus stain presence, meaning some rings do not have enough edges to form a closed curve. 

The method results are slightly better for disk L02d with an F-Score of 58\% in the presence of the same fungus stain issue that former disks. \Cref{fig:L02d_chains} illustrates this case and how the fungus perturbs the edge detection step in the middle of the disk.

~\Cref{Tab:results_nuestra-discos_presentados_en_el_articulo} illustrates the metric result over all the disks presented in the article. 

\begin{table}[htbp]
\scriptsize
\centering
\caption{Results over the UruDendro dataset (size 1500x1500 and $\sigma =3$) with  $th\_pre=60\%$. }
\begin{tabular}{|l|l|l|l|l|l|l|l|l|l|} 
\hline
\textbf{Name} & \textbf{TP} & \textbf{FP} & \textbf{TN} & \textbf{FN} & \textbf{P} & \textbf{R} & \textbf{F} & \textbf{RMSE} & \textbf{Time (sec.)} \\ 
\hline
F03d    & 19 & 1  & 0  & 2  & 0.95 & 0.91 & 0.93 & 7.81          & 11.26       \\
\hline
F02b    & 21 & 1  & 0  & 1  & 0.96 & 0.96 & 0.96 & 4.02          & 13.26       \\ 
\hline
F07b    & 17 & 3  & 0  & 6  & 0.85 & 0.74 & 0.79 & 7.99          & 45.74       \\ 
\hline
L02b    & 4  & 2  & 0  & 11 & 0.67 & 0.27 & 0.38 & 9.41          & 21.90       \\ 
\hline
F04c    & 18 & 1  & 0  & 3  & 0.95 & 0.86 & 0.90 & 5.60          & 26.88       \\ 
\hline
L09e    & 13 & 0  & 0  & 2  & 1.00 & 0.87 & 0.93 & 4.14          & 14.66       \\ 
\hline
F03b    & 20 & 0  & 0  & 3  & 1.00 & 0.87 & 0.93 & 2.15          & 13.95       \\ 
\hline
F02a    & 20 & 0  & 0  & 3  & 1.00 & 0.87 & 0.93 & 1.50          & 15.29       \\ 
\hline
F02c    & 21 & 0  & 0  & 1  & 1.00 & 0.96 & 0.98 & 3.76          & 12.26       \\ 
\hline
F02d    & 20 & 1  & 0  & 0  & 0.95 & 1.00 & 0.98 & 2.11          & 8.34        \\
\hline
F02e    & 20 & 1  & 0  & 0  & 0.95 & 1.00 & 0.98 & 7.62          & 23.31       \\ 
\hline
F03c    & 23 & 0  & 0  & 1  & 1.00 & 0.96 & 0.98 & 10.69         & 7.34        \\ 
\hline
L03c    & 15 & 1  & 0  & 1  & 0.94 & 0.94 & 0.94 & 9.01          & 8.28        \\ 
\hline
F07e    & 8  & 4  & 0  & 14 & 0.67 & 0.36 & 0.47 & 8.17          & 40.29       \\ 
\hline
L02d    & 7  & 3  & 0  & 7  & 0.70 & 0.50 & 0.58 & 5.66          & 28.80       \\ 
\hline
\end{tabular}
\label{Tab:results_nuestra-discos_presentados_en_el_articulo}

\end{table}

\subsection{Kennel}

\begin{figure*}[ht!]
\begin{centering}
    \begin{subfigure}{0.3\textwidth}
    \begin{centering}
   \includegraphics[width=\textwidth]{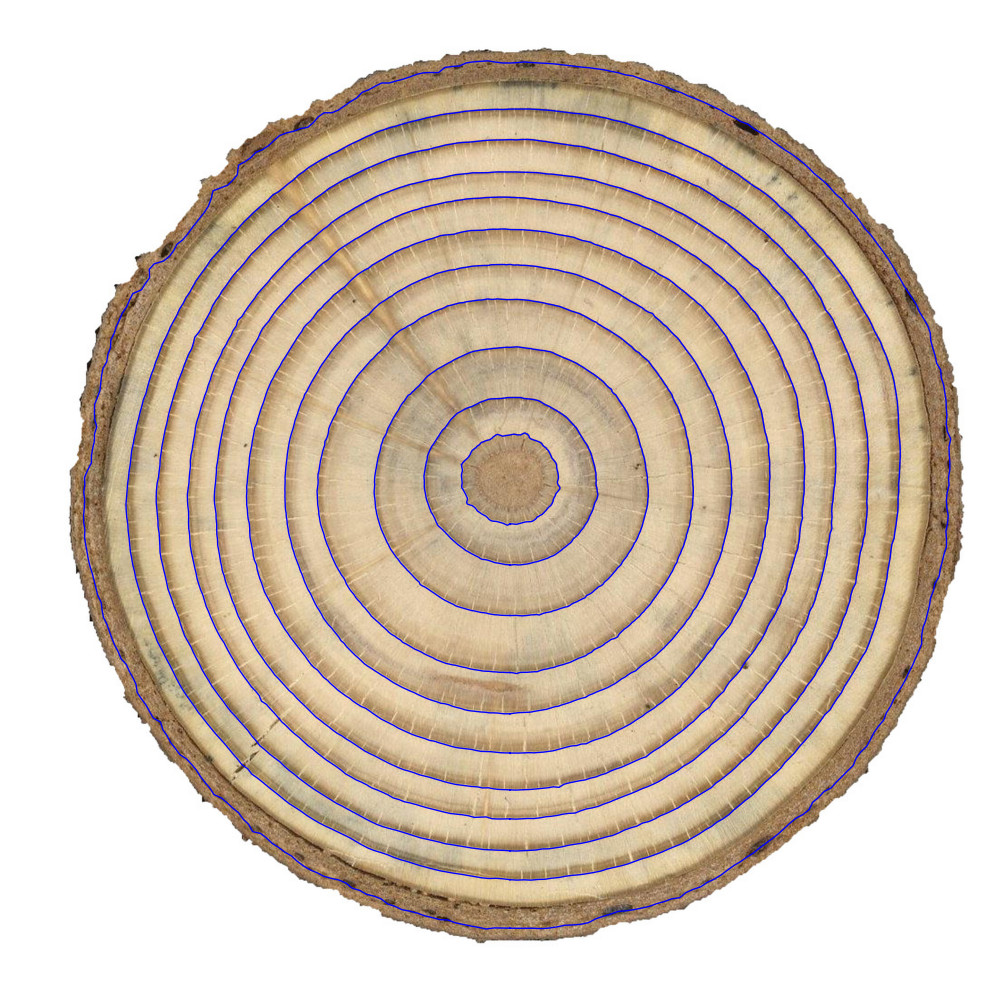}
    \label{fig:ddbb-ac4_output}
    \caption{AbiesAlba4}
    \end{centering}
    \end{subfigure}
    \hfill
    \begin{subfigure}{0.3\textwidth}
    \begin{centering}
   \includegraphics[width=\textwidth]{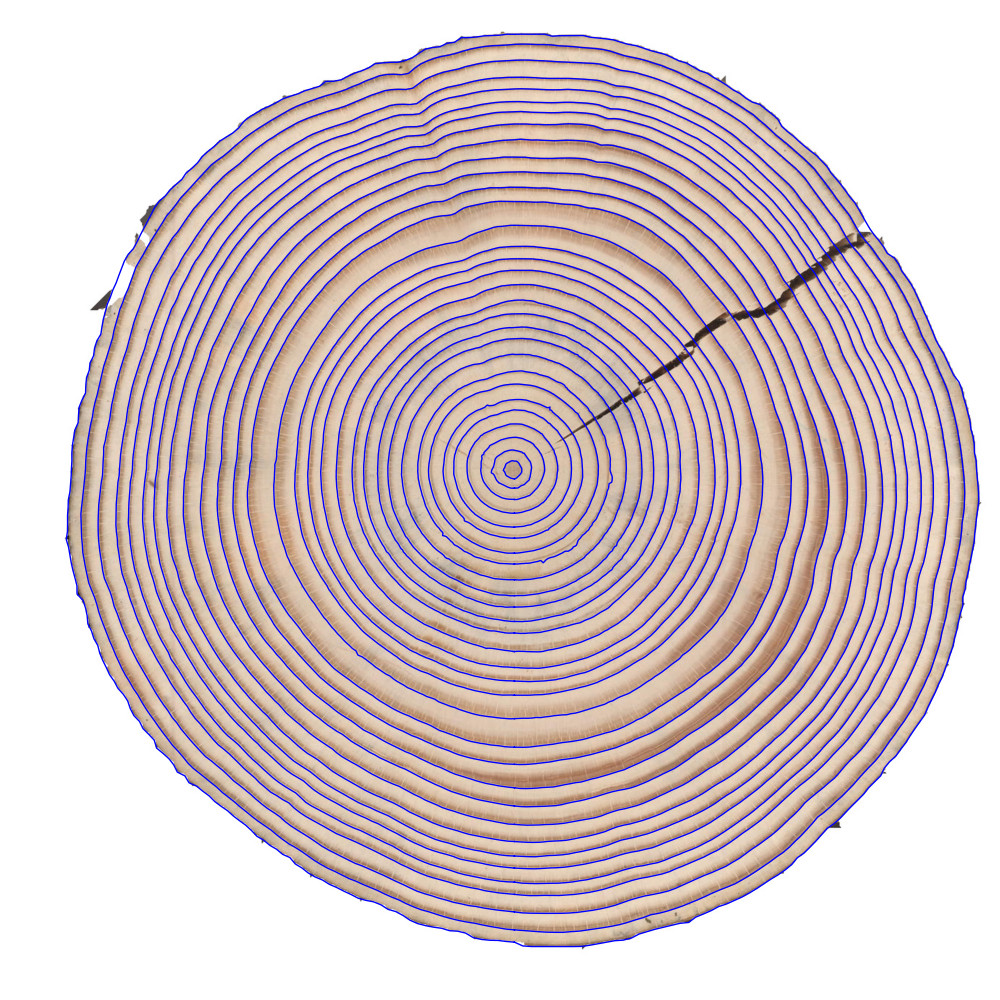}
    \label{fig:ddbb-ac5_output}
    \caption{AbiesAlba5}
    \end{centering}
    \end{subfigure}
    \hfill
    \begin{subfigure}{0.3\textwidth}
    \begin{centering}
   \includegraphics[width=\textwidth]{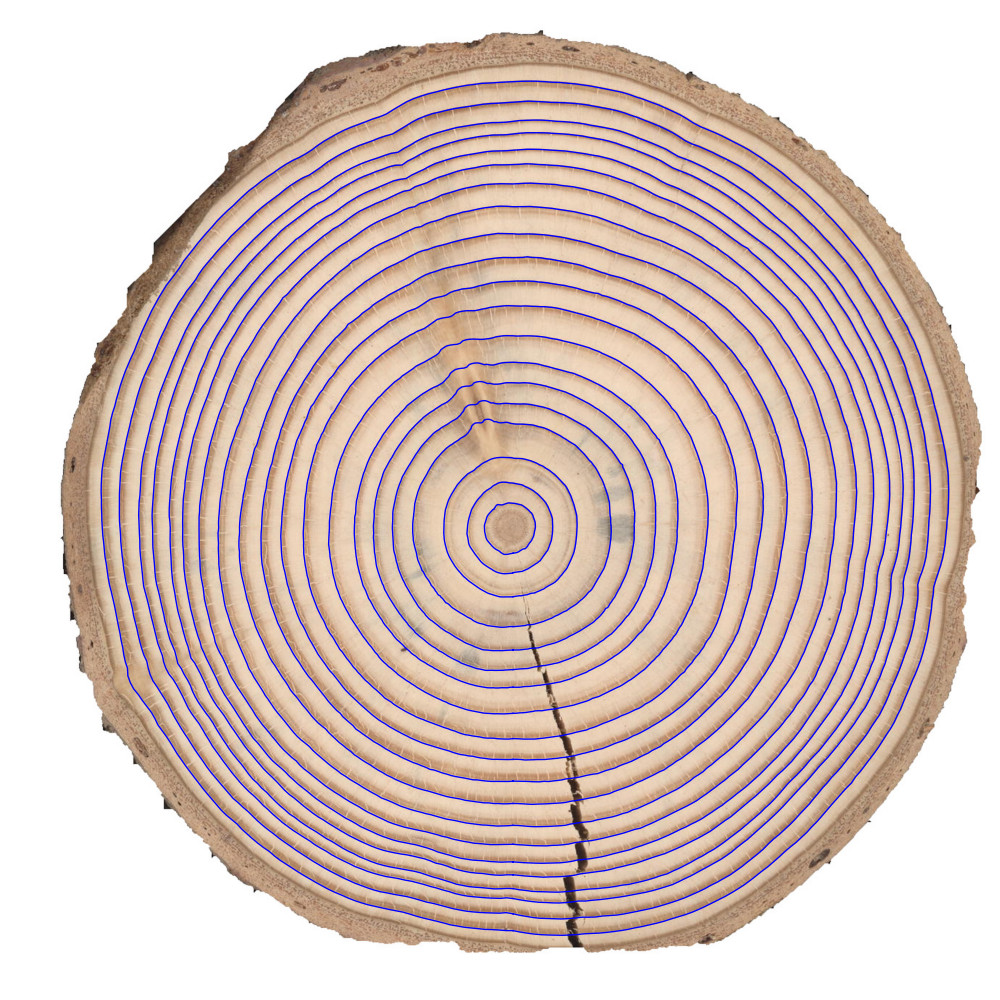}
    \label{fig:ddbb-ac6_output}
    \caption{AbiesAlba6}
    \end{centering}
    \end{subfigure}
    \hfill
    \begin{subfigure}{0.3\textwidth}
    \begin{centering}
   \includegraphics[width=\textwidth]{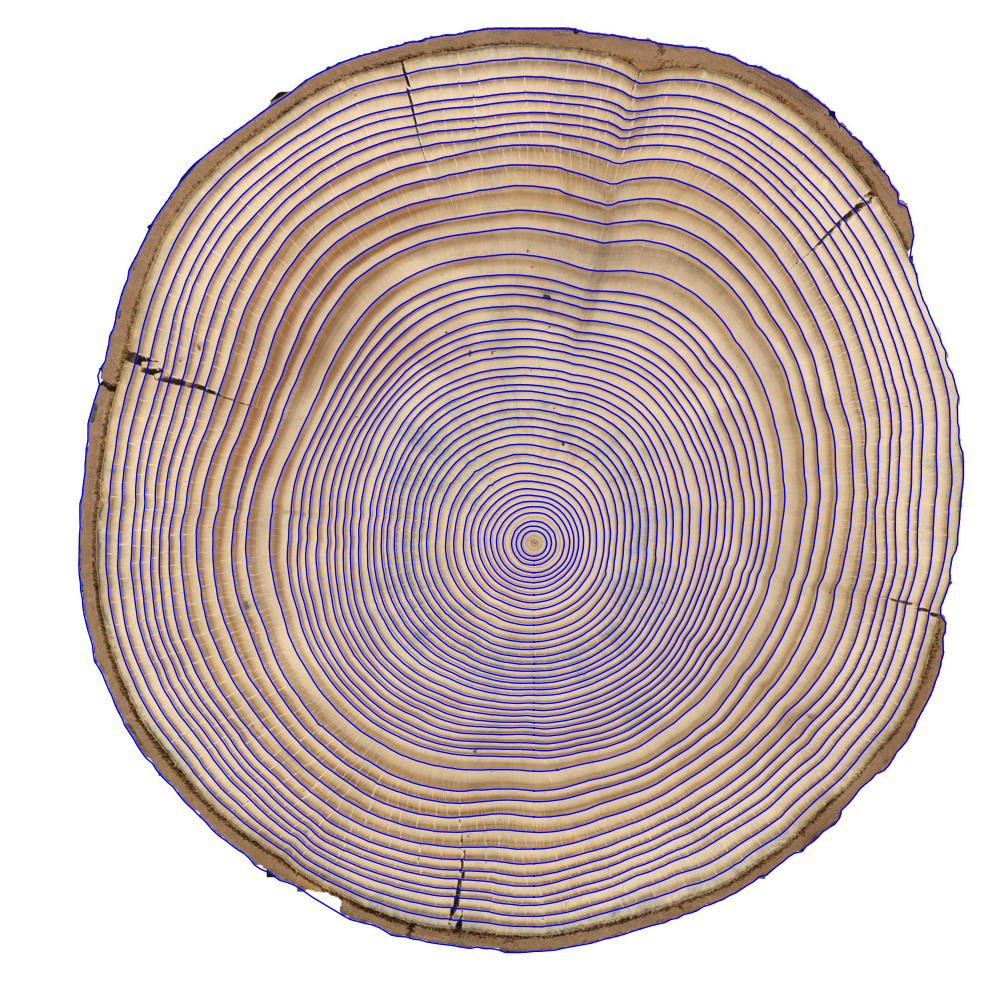}
    \label{fig:ddbb-ac7_output}
    \caption{AbiesAlba7}
    \end{centering}
    \end{subfigure}
   \caption{Results over images from the Kennel dataset with 1500x1500 image size and $\sigma=2.5$.}
   \label{fig:ddbb_ac_output_sup}
\end{centering}
\end{figure*}

In this section, we present the results of the CS-TRD method applied to the entire Kennel dataset. The first three disks of the dataset were shown in the article section, \Cref{fig:ddbb_ac_output_sup} show the results for the other four. The main difference with the UruDendro dataset is the age of the trees (number of annual rings); for example, the disk \textit{AbiesAlba7} is 48 years old, while in the UruDendro dataset, the oldest sample is 24 years old. As explained in the article, the results of the method in this dataset are really good: F-score of 97\%.

\begin{table}[htbp]
\scriptsize
\centering
\caption{Results on the Kennel dataset (size 1500x1500 and $\sigma=2.5$) with  $th\_pre=60$.}
\begin{tabular}{|c|c|c|c|c|c|c|c|c|c|}
\hline
\textbf{Name} & \textbf{TP} & \textbf{FP} & \textbf{TN} & \textbf{FN} & \textbf{P} & \textbf{R} & \textbf{F} & \textbf{RMSE} & \textbf{Time (sec.)} \\ \hline
AbiesAlba1 & 49 & 1  & 0  & 3  & 0.98 & 0.94 & 0.96 & 3.66 & 18.01      \\ \hline
AbiesAlba2 & 20 & 0  & 0  & 2  & 1.00 & 0.91 & 0.95 & 0.95 & 9.21       \\\hline
AbiesAlba3 & 26 & 1  & 0  & 1  & 0.96 & 0.96 & 0.96 & 1.30 & 8.93       \\\hline
AbiesAlba4 & 11 & 0  & 0  & 1  & 1.00 & 0.92 & 0.96 & 5.88 & 8.96       \\\hline
AbiesAlba5 & 30 & 1  & 0  & 0  & 0.97 & 1.00 & 0.98 & 1.29 & 9.06       \\\hline
AbiesAlba6 & 20 & 0  & 0  & 1  & 1.00 & 0.95 & 0.98 & 1.26 & 7.63       \\\hline
AbiesAlba7 & 45 & 0  & 0  & 3  & 1.00 & 0.94 & 0.97 & 3.58 & 13.78 \\\hline
Average & & & &  &0.99& 0.95& 0.97& 2.56 & 10.80 \\\hline
\end{tabular}
\label{tab:resultsKennel}
\end{table}

\section{Algorithms}

This section describes several algorithms not included in the main article due to space constraints.

\paragraph{Preprocessing}~\Cref{algo:preprocessing},~\Cref{algo:resize} and~\Cref{algo:equalization} describes the preprocessing step.

\begin{algorithm}[!htbp]
  \KwIn{$Im_{in}$, // input image without background. Background pixel set to white (255)  \\
  Parameter: \\
  $height_{output}$, $width_{output}$, // output image size, in pixels\\
  $cy$, $cx$, // pith's coordinates, in pixels\\
  }
  \KwOut{Preprocessed Image, pith coordinates scaled to the size of the preprocessed image}
  \SetAlgoLined
  \LinesNumbered
  \BlankLine 
   \If{None in $[height_{output}$, $width_{output}]$ }{
        $Im_r, cy_{output}, cx_{output}$ $\leftarrow$ $(Im_{in}, cy, cx)$
   }
   \Else{
        $Im_r, cy_{output}, cx_{output}$ $\leftarrow$ resize( $Im_{in}, height_{output}, width_{output}, cy, cx$) // See \Cref{algo:resize}\\
   }
  
   $Im_{g}$ $\leftarrow$ rgb2gray($Im_{r}$) \\ 
   $Im_{pre}$ $\leftarrow$  equalize($Im_{g}$) // See \Cref{algo:equalization} \\
   \KwRet{$[Im_{pre}, cy_{output}, cx_{output}]$}
\caption{preprocessing}
\label{algo:preprocessing}
\end{algorithm}

ACA

The size of acquired images can vary widely, and this has an impact on the performance. On one side, the bigger the image, the slower the algorithm, as more data must be processed. On the other hand, if the image is too small, the relevant structures will be challenging to detect. \Cref{algo:preprocessing} shows the pseudo-code of the preprocessing stage. The first step is resizing the input image to a standard size of 1500x1500 pixels. The input image size can vary, so zoom in or zoom out is applied to the input image to a fixed image size for the rest of the processing. Pith coordinates must be resized as well. This step can be turned off by the user in the demo. 

From lines 1 to 6, the former logic is implemented. The resize function (Line 5) is shown in \Cref{algo:resize}. The dimensions of the input image can vary, so image resize  (Line 1, \Cref{algo:resize}) is applied using the function \textit{resize} from Pillow library\footnote{version 10.1.0} \cite{pillow}. The method involves filtering to avoid aliasing if the flag \textit{image.Resampling.LANCZOS} is set. The center coordinates must be modified accordingly as well. To this aim, we use the following equations:

\begin{equation}
    cy_{output} = cy * \frac{height_{output}}{height} \hspace{0.5cm}cx_{output} = cx * \frac{width_{output}}{width} 
    \label{equ:center_resize}
\end{equation}

Where ($height$, $width$) is the input image dimensions, ($height_{output}$,  $width_{output}$) is the output image dimension, and (cy, cx) is the (original resolution) disk pith location coordinates in pixels.

In line 7, the RGB image is converted to grayscale using the OpenCV \cite{opencv} function: 
$$cv2.cvtColor(img, cv2.COLOR\_BGR2GRAY)$$

Finally, in line 8, a histogram equalization step is applied to enhance contrast. The method is described in Algorithm \ref{algo:equalization}. The first step (Line 1)  changes the background pixels to the mean grayscale value to avoid undesirable background effects during equalization. Both equalized and masked background images are returned.  Then, in Line 2, the Contrast Limited Adaptive Histogram Equalization (CLAHE) \cite{CLAHE} method for equalizing images is applied. We use the OpenCV implementation\footnote{https://www.geeksforgeeks.org/clahe-histogram-equalization-opencv/} \cite{opencv}. The threshold for contrast limiting is set to 10 using the $clipLimit$ parameter, while the size of the tile in which the histogram is computed is set to 8 by default ($tileSize$ parameter). Finally, in Line 3, the background of the equalized image is set to white (255).

\begin{algorithm}[!htbp]
  \KwIn{$Im_{in}$, // the input image  \\
  Parameter: \\
  $height_{output}$,  $width_{output}$: // output image size, in pixels\\
  $cy$,  $cx$: // pith's coordinates, in pixels\\
  }
  \KwOut{Resize image using Pillow Library}
  \SetAlgoLined
  \LinesNumbered
  \BlankLine 
    $Im_{r}$ $\leftarrow$  resize\_image\_using\_pil\_lib($Im_{in}, height_{output}, width_{output}$)
    
    $height, width$ $\leftarrow$  get\_image\_shape($Im_{in}$)\\
    $cy_{output}, cx_{output}$ $\leftarrow$  convert\_center\_coordinate\_to\_output\_coordinate($cy, cx, height, width, height_{output}, width_{output}$)// See \Cref{equ:center_resize}
    
\KwRet{$[Im_{r}, cy_{output}, cx_{output}]$}
\caption{resize}
\label{algo:resize}
\end{algorithm}

\begin{algorithm}[!htbp]
\KwIn{$Im_{g}$, // gray scale input image. The background is white (255)  \\
  }
\KwOut{Equalized image using OpenCV CLAHE \cite{CLAHE} method}
\SetAlgoLined
\LinesNumbered
\BlankLine
$Im_{pre}, mask$ $\leftarrow$ change\_background\_intensity\_to\_mean($Im_{g}$)

$Im_{pre}$ $\leftarrow$ equalize\_image\_using\_clahe($Im_{pre}$)

$Im_{pre}$ $\leftarrow$ change\_background\_to\_value($Im_{pre}, mask,255$)

\KwRet{[$Im_{pre}$]}
\caption{equalize}
\label{algo:equalization}
\end{algorithm}

\paragraph{Filtering the edge chains} The filter edges method is shown at  \Cref{algo:ChainFilter}. It gets as input the Deverney edges $m\_ch_e$, the pith center, the image gradient components in the form of two matrices $G_x$ and $G_y$ and the preprocessed image $Im_{pre}$. As a parameter, it needs the threshold $\alpha$. From lines 1 to  5, it computes the angle between vector $\vec{c p_i}$ and the gradient $\vec{G_{p_i}}$ at point $p_i$. We use the Python \textit{numpy} library matrix operations to speed up computation. 
In line 1, we change the edges reference axis relative to the pith location, pixel ($c_y$, $c_x$). 

Each edge gradient is saved in matrix $G$ (line 2), keeping the same edge order of the matrix $m\_ch_e$; this means that 
\begin{equation*}
   p_i = m\_ch_e[i] \rightarrow \vec{G_{p_i}} = G[i] 
\end{equation*}
Where $p_i$ is the i-row of matrix $m\_ch_e$ and $\vec{G_{p_i}}$ is the i-row of matrix $G[i]$.

In Lines 3 and 4, the matrices $Xb.T$($Xb$ transposed matrix) and $G$ are normalized, dividing the vector by the norm as shown in  \Cref{equ:filter_angle_normalized}:

\begin{equation}
\delta(R_i,\vec{G_{p_i}}) =\arccos\left( \frac{\vec{c p_i} \times \vec{G_{p_i}}}{\|\vec{c p_i}\|\|\vec{G_{p_i}}\|}\right) =  \arccos\left( \frac{\vec{c p_i}}{\|\vec{c p_i}\|} \times \frac{\vec{G_{p_i}}}{\|\vec{G_{p_i}}\|}\right)  = \arccos\left( \vec{c p_i}_{unit} \times \vec{G_{p_i}}_{unit}\right)  \label{equ:filter_angle_normalized}   
\end{equation}

In line 5, \Cref{equ:filter_angle_normalized} is computed in matrix form, and the angle between normalized vectors  $\vec{G_{p_i}}_{unit}$ and  $\vec{c p_i}_{unit}$ is returned in degrees in matrix $\theta$. The edge filtering is applied in line 6. If the  edge point $p_i$ is filtered out, then  $X_{edges\_filtered}[i]=[-1,-1]$. The edges are converted to  \textbf{curve} objects in line 7. We say that two edge pixels belong to the same edge if,  between them, there does not exist a row in matrix $X_{edges\_filtered}$ with values [-1,-1]. The object \textbf{Curve} inherits the properties of the class \textbf{LineString} from the \textbf{Shapely} library\footnote{version 1.7} \cite{shapely}, which is used in the \textit{sampling edges} stage. Finally, in lines 8 and 9, the curve belonging to the border edges is computed and added to the curve list, $l\_ch_f$. In this context, the limits of the segmented image concerning the background are named $border$. 

The function $get\_border\_curve$ is shown in \Cref{algo:get_border}. We use a simple method to compute the border edges. First, we generate a mask, an image of the same dimensions as $Im_{pre}$, enlarged by $3$ lines and $3$ columns before the first and after the last line and column to avoid filtering border effects. The mask image has two values: $0$ for the region of the wood slice and $255$ for the background. Lines 1 to 4  calculate the mask image. In line 1
we threshold $Im_{pre}$ masking all the pixels with a value equal to 255, particularly the background. Some internal pixels can also have a value equal to 255. To avoid those pixels in the mask, we blur the mask (line 2), using a Gaussian Kernel with a high $\sigma$ (in our implementation $\sigma = 11$), and we set to 255 all the pixels with a value higher than 0 (lines 2 and 3). In line 4, the mask is padded with $pad = 3$.  Finally, in line 5, we apply an OpenCV finding contour method to get the border contour on the mask. The OpenCV implementation returns all the contours it finds, including the image's border. We select the contour for which the enclosed area is closest to half the image. This is a criterion that works fine for this purpose.  In line 6, the contour object is converted to a \textbf{Curve} object.

\begin{algorithm}[!htbp]
  \KwIn{$m\_ch_{e}$, // matrix of edge \textit{curves}  \\ 
        $cy$, $cx$, // pith's coordinates, in pixels \\
        $G_x$, $G_y$, // X and Y components of the gradient, a matrix\\
        $Im_{pre}$, // Preprocessed image\\
        Parameter:\\
        $\alpha$, //Threshold edge filter\\
   }
  \KwOut{A list $l\_ch^{k}_{f}$, $k=1,2, \cdots, N$, where each element is a filtered  edge \textit{curve}}
  \SetAlgoLined
  \LinesNumbered
  \BlankLine 
    $Xb$ $\leftarrow$  change\_reference\_axis($m\_ch_e$, $c_y$, $c_x$)// New origin is the pith center \\
    $G$ $\leftarrow$  get\_gradient\_vector\_for\_each\_edge\_pixel($m\_ch_e$, $G_x$, $G_y$)\\
    $Xb_{normalized}$ $\leftarrow$  normalized\_row\_matrix(Xb.T)\\
    $G_{normalized}$ $\leftarrow$  normalized\_row\_matrix(G)\\
    $\theta$ $\leftarrow$  compute\_angle\_between\_gradient\_and\_edges($Xb_{normalized}$, $G_{normalized}$)\\
    $X_{edges\_filtered}$ $\leftarrow$  filter\_edges\_by\_threshold($m\_ch_e$, $\theta$, $\alpha$)\\
    $l\_ch_f$ $\leftarrow$  convert\_masked\_pixels\_to\_curves($X_{edges\_filtered}$)\\
    $border\_curve$ $\leftarrow$  get\_border\_curve($Im_{pre}$, $l\_ch_f$)// See  \Cref{algo:get_border}\\
    $l\_ch_f$ $\leftarrow$  $l\_ch_f$ + $border\_curve$\\
    \KwRet{$l\_ch_f$}
\caption{filter\_edges}
\label{algo:ChainFilter}
\end{algorithm}



\begin{algorithm}[!htbp]
\KwIn{$Im_{pre}$, // Preprocessed image \\
$l\_ch_f$, //list of object Curves
}
\KwOut{$border\_curve$}
\SetAlgoLined
\LinesNumbered
\BlankLine 
$mask$ $\leftarrow$  mask\_background($Im_{pre}$)\\
$mask$ $\leftarrow$  blur($mask$)\\
$mask$ $\leftarrow$  thresholding($mask$)\\
$mask$ $\leftarrow$  padding\_mask($mask$)\\
$border\_contour$ $\leftarrow$  find\_border\_contour($mask$, $Im_{pre}$)\\
$border\_curve$ $\leftarrow$  contour\_to\_curve($border\_contour$, len($l\_ch_f$))\\
\KwRet{$border\_curve$}
\caption{get\_border\_curve}
\label{algo:get_border}
\end{algorithm}

\paragraph{Sampling edges} ~\Cref{algo:samplingEdges} described the module logic. 

\begin{algorithm}[!htbp]
  \KwIn{
  $l\_ch_{f}$, // list of \textit{curves} \\ 
  $cy$, $cx$, // pith's coordinates, in pixels\\
  $Im_{pre}$ // preprocessed image  \\
  Parameters: \\
$min\_chain\_length$, //  minimum length of a chain \\
$nr$ // number of total rays\\
  }
  \KwOut{A list $l\_ch^{k}_{s}$, $k=1,2, \cdots, N$, where each element is a \textit{chain}; \\
  $l\_nodes^{k}_{s}$, $k=1,2, \cdots, N_n$, where each element is a \textit{Node}\\}
  \SetAlgoLined
  \LinesNumbered
  \BlankLine 
$height$, $width$ $\leftarrow$ $Im_{pre}$.shape\\
$l\_rays$ $\leftarrow$ build\_rays($nr$, $height$, $width$, $cy$, $cx$)\\
 $l\_ch_{s}$,$l\_nodes_s$ $\leftarrow$ intersections\_between\_rays\_and\_devernay\_curves($cy$, $cx$, $l\_rays$ , $l\_ch_f$, $min\_chain\_length$, $nr$, $height$, $width$) \\
\tcc{Virtual center chain is added in the input list $l\_ch_s$. Additionally, its nodes are added in  $l\_nodes_s$}
generate\_virtual\_center\_chain($cy$, $cx$, nr, $l\_ch_s$,  $l\_nodes_s$, $height$, $width$)\\
   \KwRet{$l\_ch_{s}$,$l\_nodes_s$}
\caption{sampling\_edges}
\label{algo:samplingEdges}
\end{algorithm}

\paragraph{Connect chains} in addition to the corresponding section in the paper, ~\Cref{algo:connect},~\Cref{algo:update_system},~\Cref{algo:get_next_chain},~\Cref{algo:fill_chain_if_there_is_no_overlapping},~\Cref{algo:exist_chain_overlapping},~\Cref{algo:connect_two_chains} and~\Cref{algo:get_closest_chain_logic} describes the step. 

\begin{algorithm}[!htbp]
  \KwIn{  $l\_ch_{s}$,// chains list \\
  $l\_nodes_s$,// nodes list \\
  $cy$,  $cx$, // pith's coordinates x, in pixels\\
  $nr$// number of total rays\\
  }
\KwOut{A list $l\_ch^{f}_{c}$, $f=1,2, \cdots, N_f$, where each element is a \textit{chain}; \\
  $l\_nodes^{f}_{c}$, $c=1,2, \cdots, N_f$, where each element is a \textit{Node}\\}
\SetAlgoLined
\LinesNumbered
\BlankLine 
    $M$ $\leftarrow$ compute\_intersection\_matrix($l\_ch_{s}$, $l\_nodes_s$, $nr$).\\
    \tcc{Loop for  connecting chain main logic, losing the restrictions at each iteration}
    \For{$i \leftarrow 1$ \KwTo $9$}{
         $iteration\_params$ $\leftarrow$ get\_iteration\_parameters($i$)//Table (\ref{tab:connectivityParameters})\\
            $l\_ch_c$, $l\_nodes_c$,  $M$ $\leftarrow$  connect\_chain\_main\_logic($M$, $cy$, $cx$, $nr$, $iteration\_params$) \\
        update\_list\_for\_next\_iteration($l\_ch_c,l\_nodes_c$)
    }
\KwRet{$l\_ch_{c}$,$l\_nodes_{c}$}
\caption{Connect Chains}
\label{algo:connect}
\end{algorithm} 

\begin{algorithm}[!htbp]
  \KwIn{ $state$,// class object that has a pointer to all the system objects\\
  $Ch_i$, // current support chain \\
  $l\_s\_{outward}$,// outward chain list \\
  $l\_s\_{inward}$,// inward chain list \\
  }
\KwOut{index of chain for next iteration. Stored in class $state$\\}
\SetAlgoLined
\LinesNumbered
\BlankLine 
$l\_ch_s$ $\leftarrow$ $state$.get\_list\_chains()\\
\If{$state$.system\_status\_change()}{
    sort\_chain\_list\_by\_descending\_size($l\_ch_s$)\\
    $l\_current\_iteration$ $\leftarrow$ $Ch_i$ + $S_{outward}$ + $S_{inward}$\\
    sort\_chain\_list\_by\_descending\_size($l\_current\_iteration$)\\
    $longest\_chain$ $\leftarrow$ $l\_current\_iteration[0]$//$l\_current\_iteration$ is sorted by size\\
    \If{$Ch_i = longest\_chain$}
    {
        $next\_chain\_index$ $\leftarrow$ get\_next\_chain\_index\_in\_list($l\_ch_s$, $Ch_i$)\\
    }
    \Else{
        $next\_chain\_index$ $\leftarrow$ get\_chain\_index\_in\_list($l\_ch_s$,$longest\_chain$)\\
    }
    
}
\Else{
    $next\_chain\_index$ $\leftarrow$ get\_next\_chain\_index\_in\_list($l\_ch_s$, $Ch_i$)\\
}
$state$.$next\_chain\_index$ $\leftarrow$ $next\_chain\_index$\\
\KwRet{}
\caption{update\_system\_status}
\label{algo:update_system}
\end{algorithm} 

\begin{algorithm}[!htbp]
  \KwIn{
  $state$,// class object that has  pointers to all the system objects\\
  }
\KwOut{next support chain\\}
\SetAlgoLined
\LinesNumbered
\BlankLine 
$l\_ch_s$ $\leftarrow$ $state$.get\_list\_chains()\\
$Ch_i$ $\leftarrow$ $l\_ch_s$[$state$.next\_chain\_index]\\
$state$.size\_l\_chain\_init $\leftarrow$ length($l\_ch_s$)\\
$state$.fill\_chain\_if\_there\_is\_no\_overlapping($Ch_i$),// See \Cref{algo:fill_chain_if_there_is_no_overlapping} \\
\KwRet{$Ch_i$}
\caption{get\_next\_chain}
\label{algo:get_next_chain}
\end{algorithm} 

\begin{algorithm}[!htbp]
  \KwIn{
  $state$,// class object that has  pointers to all the system objects\\
  $chain$, // chain to be completed if conditions are met. Passed by reference.
  }
\KwOut{Void. If nodes are created, they are added to $chain$ and $state$ directly\\}
\SetAlgoLined
\LinesNumbered
\BlankLine 
$l\_ch_s$ $\leftarrow$ $state$.get\_list\_chains()\\
$threshold$ $\leftarrow$ 0.9 \\  
\If{$chain$.size $\not \in$ [$threshold, 1 ) \times chain$.Nr}{
\KwRet{}
}
$Ch_i$, $l\_nodes$, $endpoint\_type$ $\leftarrow$ $state$.compute\_all\_elements\_needed\_to\_check\_if\_exist\_chain\_overlapping($chain$)\\
$exist\_chain$ $\leftarrow$ exist\_chain\_overlapping($l\_ch_s$, $l\_nodes$, $chain$, $chain$, $endpoint\_type$, $Ch_i$)//See  \Cref{algo:exist_chain_overlapping}\\
\If{$exist\_chain$}{
\KwRet{}
}
$state$.add\_nodes\_list\_to\_system($chain$, $l\_nodes$)\\
\KwRet{}
\caption{fill\_chain\_if\_there\_is\_no\_overlapping}
\label{algo:fill_chain_if_there_is_no_overlapping}
\end{algorithm} 

\begin{algorithm}[!htbp]
  \KwIn{
  $l\_ch_s$,//list chains\\
  $l\_nodes$, // list of interpolated nodes plus the  endpoints\\
  $Ch_j$,// source chain.\\
  $Ch_k$,// destination chain.\\
  $endpoint\_type$,// source chain endpoint (A or B)\\
  $Ch_i$,// support chain\\
  }
\KwOut{Boolean. True if exist chain belonging to $l\_ch_s$ in band}
\SetAlgoLined
\LinesNumbered
\BlankLine 
$info\_band$ $\leftarrow$ InfoVirtualBand($l\_nodes$, $Ch_j$, $Ch_k$, $endpoint\_type$,$Ch_i$)\\
$l\_chains\_in\_band$ $\leftarrow$ exist\_chain\_in\_band\_logic($l\_ch\_s$, $info\_band$)\\
$exist\_chain$ $\leftarrow$ len($l\_chains\_in\_band$) $>$ 0\\
\KwRet{$exist\_chain$}
\caption{exist\_chain\_overlapping}
\label{algo:exist_chain_overlapping}
\end{algorithm} 

\begin{algorithm}[!htbp]
  \KwIn{$state$,\\
  $Ch_j$, // current chain to be connected  \\
  $Ch_k$, // closest chain to be connected with $Ch_j$ \\
  $l\_candidates\_Ch_i$, // set of chains where to pick chains to connect with $Ch_j$\\
  $endpoint$, //$Ch_j$ endpoint to be connected\\
  $Ch_i$, //support chain $Ch_i$\\
  }
\KwOut{Void. Nodes are added to $Ch_j$ and system list are updated in $state$}

\SetAlgoLined
\LinesNumbered
\BlankLine 
\If{$endpoint$ is None }{
\tcc{If $endpoint$ is None, there is no closest chain to connect with.}
\KwRet{}
}
\If{$Ch_i$ == $Ch_k$ }{
\tcc{If $Ch_i$ and $Ch_k$ are the same chains, connecting them is impossible.}
\KwRet{}
}

generate\_new\_nodes($state$, $Ch_j$, $Ch_k$, $endpoint$, $Ch_i$)\\
updating\_chain\_nodes($state$, $Ch_j$, $Ch_k$)\\
update\_chain\_after\_connect($state$,  $Ch_j$, $Ch_k$)\\
delete\_closest\_chain($state$,  $Ch_k$, $l\_candidates\_Ch_i$)\\
update\_intersection\_matrix($state$, $Ch_j$, $Ch_k$)\\
update\_chains\_ids($state$, $Ch_k$)\\
\KwRet{}
\caption{connect\_two\_chains}
\label{algo:connect_two_chains}
\end{algorithm} 

\begin{algorithm}[!htbp]
  \KwIn{$state$,\\
  $Ch_j$, // current chain  \\
  $l\_candidates\_Ch_i$, // set of visible chains from $Ch_i$\\
  $l\_no\_intersection\_j$,//list of chains belonging to  $l\_candidates\_Ch_i$ that do not intersect with $Ch_j$\\
  $Ch_i$, //support chain \\
  $location$, // location of set $l\_candidates\_Ch_i$ with respect to $Ch_i$ (inward or outward)\\
  $enpoint$,// $Ch_j$ endpoint A or B to be connected\\
  }
\KwOut{closest chain to $Ch_j$\\}
\SetAlgoLined
\LinesNumbered
\BlankLine 
$M$ $\leftarrow$ $state$.M\\
$Ch_k$ $\leftarrow$ get\_closest\_chain($state$, $Ch_j$, $l\_no\_intersection\_j$, $Ch_i$, $location$, $endpoint$, $M$) \\
\If{$Ch_k$ is None}{
\KwRet{$Ch_k$}
}
$l\_no\_intersection\_k$ $\leftarrow$ get\_non\_intersection\_chains($M$, $l\_candidates\_Ch_i$, $Ch_k$)\\
\If{$endpoint$ = B}{
$endpoint_k$ = A
}
\Else{
$endpoint_k$ = B
}
$symmetric\_chain$ $\leftarrow$ get\_closest\_chain($state$, $Ch_k$,  $l\_no\_intersection\_k$, $Ch_i$, $location$, $endpoint_k$, $M$)\\
\If{not ($symmetric\_chain$ = $Ch_j$) and not $\left( length(Ch_k) + length(Ch_j) \right) \leq Nr$ }
{
$Ch_k$ = None\\
}
\KwRet{$Ch_k$}
\caption{get\_closest\_chain\_logic}
\label{algo:get_closest_chain_logic}
\end{algorithm}

\begin{table}[!htbp]
\centering
\begin{tabular}{|l|l|l|l|l|l|l|l|l|l|}
\hline
                   & 1   & 2   & 3   & 4   & 5   & 6   & 7   & 8   & 9   \\ \hline
\textit{$Th_{Radial\_tolerance}$}        & 0.1 & 0.2 & 0.1 & 0.2 & 0.1 & 0.2 & 0.1 & 0.2 & 0.2 \\ \hline
\textit{$Th_{Distribution\_size}$} & 2   & 2   & 3   & 3   & 3   & 3   & 2   & 3   & 3   \\ \hline
\textit{$Th_{Regular\_derivative}$}      & 1.5 & 1.5 & 1.5 & 1.5 & 1.5 & 1.5 & 2   & 2   & 2   \\ \hline
\textit{NeighbourdhoodSize} & 10  & 10  & 22  & 22  & 45  & 45  & 22  & 45  & 45  \\ \hline
\textit{derivFromCenter }   & 0   & 0   & 0   & 0   & 0   & 0   & 1   & 1   & 1   \\ \hline
\end{tabular}
\caption{Connectivity Parameters. Each column is the parameter set used on that iteration.}
\label{tab:connectivityParameters}
\end{table}

\begin{algorithm}[!htbp]
  \KwIn{$state$,\\
  $Ch_j$, // current chain   \\
  $Ch_i$, // support chain\\
  $endpoint$, // $Ch_j$ endpoint \\
  $candidate\_chain\_radial\_distance$, //radial difference between $Ch_j$  and $candidate\_chain$ endpoints \\
  $candidate\_chain$, // angular closer chain to $Ch_j$\\
  $M$,// intersection matrix\\
  $l\_sorted\_chains\_in\_neighbourhood$, // chains in  $Ch_j$ endpoint neighbourhood sorted by angular distance\\
  }
\KwOut{closest chain to $Ch_j$ that satisfies connectivity goodness conditions.\\}
\SetAlgoLined
\LinesNumbered
\BlankLine 
$l\_intersections\_candidate$ $\leftarrow$ intersection\_chains($M$, $candidate\_chain$, $l\_sorted\_chains\_in\_neighbourhood$)\\
$l\_intersections\_candidate\_set$ $\leftarrow$ get\_all\_chain\_in\_subset\_that\_satisfy\_condition($state$, $Ch_j$, $Ch_i$, $endpoint$, $candidate\_chain\_radial\_distance$, $candidate\_chain$, $l\_intersections\_candidate$)\\
sort\_set\_list\_by\_distance($l\_intersections\_candidate\_set$)\\
$Ch_k$ $\leftarrow$ l\_intersections\_candidate\_set[0].cad\\
\KwRet{$Ch_k$}
\caption{get\_the\_closest\_chain\_by\_radial\_distance\_that\_does\_not\_intersect}
\label{algo:get_the_closest_chain_by_radial_distance_that_does_not_intersect}
\end{algorithm}

\paragraph{postprocessing}

\begin{algorithm}[!htbp]
  \KwIn{
  $l\_ch_{c}$, // chains list \\
  $l\_nodes_c$, // nodes list \\
}
\KwOut{A list of post-processed \textit{chains} $l\_ch^{k}_{p}$, $k=1,2, \cdots, N_f$ \\}
    $l\_ch_p$, $idx\_start$ $\leftarrow$ initialization($l\_ch_c$)\\
    \While{True}{
        ctx $\leftarrow$ DiskContext($l\_ch_{p}$, $idx\_start$)\\
        \While{$len(ctx.completed\_chains) > 0$}
        {
            $l\_within\_chains$, $inward\_ring$, $outward\_ring$ $\leftarrow$ ctx.update()\\
            $chain\_was\_completed$ $\leftarrow$ split\_and\_connect\_chains($l\_within\_chains$, $inward\_ring$, $outward\_ring$, $l\_ch_p$,  $l\_nodes_c$,$ctx.neighbourhood\_size$) // See  \Cref{algo:split_and_connect_chains}\\
            \If {chain\_was\_completed}{
                    idx\_start $\leftarrow$ ctx.idx\\
                    break\\
            }
            connect\_chains\_if\_there\_is\_enough\_data($ctx$, $l\_nodes_c$, $l\_ch_p$)\\
            \If {ctx.exit()}
            {
                break\\
            }
        }
        \If{not $chain\_was\_completed$}
        {break\\
        }
   
}
complete\_chains\_if\_required($l\_ch_p$)\\
\SetAlgoLined
\LinesNumbered
\BlankLine 
\KwRet{$l\_ch_{p}$}
\caption{Posprocessing Main Logic}
\label{algo:posprocessing}
\end{algorithm}

The method ${split\_and\_connect\_chains}$ is described in \Cref{algo:split_and_connect_chains}. It iterates over all the chains within a region to connect them. At every chain endpoint, the method cuts all the chains that intersect the ray passing through that endpoint and checks the connectivity goodness conditions between the divided chains to find connections. Notice that this module removes the chain overlapping constraint. The parameters used by this module are:
\begin{enumerate}
    \item $neighbourhood\_size$ = 45 
    \item $Th_{Radial\_tolerance}$ = 0.2
    \item $Th_{Distribution\_size}$ = 3
    \item $Th_{Regular\_derivative}$ = 2
\end{enumerate}

The parameter $neighbourhood\_size$  defines the maximum angular distance to consider candidate chains departing from an endpoint in both directions. Given a source chain (the current chain $Ch_j$), every chain in the region that overlaps $Ch_j$ in more than $neighbourhood\_size$ is not considered a candidate chain to connect because if there is a very long overlapping, that chain is probably part of another ring. In line 1, the method is initialized, and variables $connected$, $completed\_chain$, and $Ch_j$ are set to FALSE. Also, the \textit{chains} in $l\_within\_chains$ list are sorted by size in descending order. The chain nodes inside the region are stored in the $l\_inward\_nodes$ list (line 2). The main loop, lines 3 to 15, iterates over the chains in $l\_within\_chains$. The loop terminates when either current chain $Ch_j$ is closed or all chains in list $l\_within\_chains$ have been tested. In line 10, a new (non-treated) chain is extracted for the current iteration and stored in $Ch_j$. The splitting and connecting logic called $split\_and\_connect\_neighbouring\_chains$ is executed in line 13. The best candidate chain for endpoint A ($Ch_k^a$) is determined at this point, while the best candidate chain for endpoint B ($Ch_k^b$) is obtained in line 14. $Ch_i^a$ and $Ch_i^b$ are the support chains of chains $Ch_k^a$ and $Ch_k^b$ respectively. The radial distance of chain $Ch_k^a$($Ch_k^b$) to $Ch_j$ trough endpoint $A$($B$) is  $diff_a$($diff_b$). The radially closest candidate chain is connected in function $connect\_radially\_closest\_chain$ (line 15), and the maximum number of nodes ($Nr=360$) constraint is verified. The same node interpolation as in line 15 of \Cref{algo:posprocessing} is used.

\begin{algorithm}[!htbp]
\KwIn{
  $l\_within\_chains$, // uncompleted chains delimited by inward\_ring and outward\_ring \\
  $inward\_ring$, // inward ring of the region \\
  $outward\_ring$, // outward ring of the region \\
  $l\_ch_p$, // chain list \\
  $l\_nodes_c$, // full nodes list\\
  Parameter:\\
  $neighbourhood\_size$ //  size to search for chains that intersect the other endpoint\\
}
\KwOut{boolean value indicating if a chain has been completed on the region}
$connected$, $completed\_chain$, $Ch_j$ $\leftarrow$ initialization\_step($l\_within\_chains$)\\
$l\_inward\_nodes$ $\leftarrow$ get\_nodes\_from\_chain\_list($l\_within\_chains$)\\
\While{True}{
\If{not $connected$}{

\If{$Ch_j$ is not None and $Ch_j.is\_closed(threshold=0.75)$}{
complete\_chain\_using\_2\_support\_ring($inward\_ring$, $outward\_ring$, $Ch_j$)\\
completed\_chain $\leftarrow$ True\\
$Ch_j$ $\leftarrow$ None\\
}
\Else{
$Ch_j$ $\leftarrow$ get\_next\_chain( $l\_within\_chains$)\\
}
}
\If{$Ch_j$ == None}
{break}
$Ch_k^a$, $diff\_a$, $Ch_i^a$ $\leftarrow$ split\_and\_connect\_neighbouring\_chains($l\_inward\_nodes$,
$l\_within\_chains$, $Ch_j$, $A$, $outward\_ring$, $inward\_ring$, $neighbourhood\_size$)\\ 

$Ch_k^b$, $diff\_b$, $Ch_i^b$ $\leftarrow$ split\_and\_connect\_neighbouring\_chains($l\_inward\_nodes$,
$l\_within\_chains$, $Ch_j$, $B$, $outward\_ring$, $inward\_ring$, $neighbourhood\_size$, $aux\_chain=Ch_k^a$)//See  \Cref{algo:split_and_connect_neighbouring_chains}\\ 

$connected$, $Ch_i$, $endpoint$ $\leftarrow$ connect\_radially\_closest\_chain($Ch_j$, $Ch_k^a$, $diff\_a$, $Ch_i^a$, $Ch_k^b$, $diff\_b$, $Ch_i^b$, $l\_ch_p$,
$l\_within\_chains$, $l\_nodes_c$, $inward\_ring$, $outward\_ring$)\\

}
\SetAlgoLined
\LinesNumbered
\BlankLine 
\KwRet{$completed\_chain$}
\caption{split\_and\_connect\_chains}
\label{algo:split_and_connect_chains}
\end{algorithm}

Given a chain $Ch_j$, the logic for splitting neighborhood chains and searching for candidates is implemented by \Cref{algo:split_and_connect_neighbouring_chains}. Chains that intersect the ray supporting $Ch_j$ endpoint are split. In line 1, the angle domain of $Ch_j$ is stored in $Ch_j\_angle\_domain$. In line 2, variable $Ch_j\_node$ stores the $Ch_j$ node endpoint. The closest chain ring to the $Ch_j$ endpoint (in Euclidean distance) is selected as the support chain, $Ch_i$. From lines 4 to 7, we have the logic to get all chains in the region delimited by two rings intersecting the ray $Ray_e$ supporting the endpoint. First, we store in $l\_nodes\_ray$ all the nodes over $Ray_e$, pinpointing the chains supporting those nodes, and keep those chains in $l\_endpoint\_chains$. To be cut, the overlapping between a chain in the set $l\_endpoint\_chains$ and $Ch_j$ must be smaller than $neighbourhood\_size$. Otherwise, it is filtered because that chain belongs to another ring (line 7).  The method in line 8 effectively cut the chosen chains. Once a chain is cut, it produces a  $sub\_chain$ nonintersecting $Ch_j$, stored as a candidate chain in $l\_candidates$ (line 8) and in the list $l\_no\_intersections\_j$ (line 9). In line 10, all chains that intersect $Ch_j$ in the second endpoint and are in the neighborhood of the first endpoint are added to $l\_candidates$. Also, the chains in the $Ch_j$ chain neighborhood, which does not intersect its endpoint but intersects in the other endpoint, are split (line 11). In line 12,  all chains in $l\_no\_intersections\_j$ that are far away regarding angular distance from the given endpoint of $Ch_j$ are removed. The nonintersecting chains in the endpoint neighborhood are stored in $l\_filtered\_no\_intersection\_j$. In line 13, chains in $l\_filtered\_no\_intersection\_j$ are added to $l\_candidates$. In line 14, all chains from $l\_candidates$ which do not satisfy the connectivity goodness conditions are discarded. In line 15, the closest chain that meets the connectivity goodness conditions is returned, $Ch_k$, the radial difference between $Ch_k$ and $Ch_j$ endpoints $diff$, and the support chain, $Ch_i$.

\begin{algorithm}[!htbp]
  \KwIn{$l\_within\_nodes$, // nodes within region\\
  $l\_within\_chain$,  //  uncompleted chains delimited by inward and outward rings\\
  $Ch_j$, // current source chain. The one that is being  connected if conditions are met\\
  $endpoint$,// endpoint of source chain to find candidate chains to connect \\
  $outward\_ring$, // outward support chain ring \\
  $inward\_ring$, // inward support chain ring\\
  $neighbourhood\_size$, // total nodes size to search for chains that intersect the other endpoint
}
\KwOut{Source chain $Ch_k$,  $radial\_distance$ and closest support chain to endpoint, $Ch_i$}

$Ch_j\_angle\_domain$  $\leftarrow$ get\_angle\_domain($Ch_j$)\\
$Ch_j\_node$ $\leftarrow$ get\_node\_endpoint($Ch_j$,$endpoint$)\\
$Ch_i$ $\leftarrow$ select\_support\_chain($outward\_ring$, $inward\_ring$, $Ch_j\_node$)\\
$l\_nodes\_ray$ $\leftarrow$ select\_nodes\_within\_region\_over\_ray($Ch_j$, $Ch_j\_node$, $l\_within\_nodes$)\\
$l\_chain\_id\_ray$$\leftarrow$extract\_chains\_ids\_from\_nodes($l\_nodes\_ray$)\\
$l\_endpoint\_chains$ $\leftarrow$ get\_chains\_from\_ids($l\_within\_chains$, $l\_chain\_id\_ray$)\\

$l\_filtered\_chains$ $\leftarrow$ remove\_chains\_with\_higher\_overlapping\_threshold($Ch_j\_angle\_domain$, $l\_endpoint\_chain$,$neighbourhood\_size$)\\
$l\_candidates$ $\leftarrow$ split\_intersecting\_chains($Ch_j\_node.angle$, $l\_filtered\_chains$, $Ch\_j$)//See  \Cref{algo:split_intersecting_chains}\\
$l\_no\_intersections\_j$ $\leftarrow$ get\_chains\_that\_no\_intersect\_src\_chain($Ch_j$, $Ch_j\_angle\_domain$, $l\_within\_chains$,$l\_endpoint\_chains$)\\
add\_chains\_that\_intersect\_in\_other\_endpoint($l\_within\_chains$, $l\_no\_intersections\_j$, $l\_candidates$, $Ch_j$, $neighbourhood\_size$, $endpoint$)\\
$l\_candidates$ $\leftarrow$ split\_intersecting\_chain\_in\_other\_endpoint($endpoint$, $Ch_j$, $l\_within\_chains$, $l\_within\_nodes$, $l\_candidates$)\\
$l\_filtered\_no\_intersection\_j$ $\leftarrow$ filter\_no\_intersected\_chain\_far($l\_no\_intersections\_j$, $Ch_j$, $endpoint$, $neighbourhood\_size$)\\
$l\_candidates$ $\leftarrow$ $l\_candidates$ + $l\_filtered\_no\_intersection\_j$\\
$l\_ch_k\_ Euclidean\_distances$, $l\_ch_k\_radial\_distances$, $l\_ch_k$ $\leftarrow$ \
        get\_chains\_that\_satisfy\_similarity\_conditions($Ch_i$, $Ch_j$, $\_candidates$, $endpoint$)\\
$Ch_k$, $diff$ $\leftarrow$ select\_closest\_candidate\_chain($l\_ch_k$, $l\_ch_k\_ Euclidean\_distances$, $l\_ch_k\_radial\_distances$, $l\_within\_chains$, $aux\_chain$)\\
\SetAlgoLined
\LinesNumbered
\BlankLine 
\KwRet{$Ch_k$,$diff$,$Ch_i$}
\caption{split\_and\_connect\_neighbouring\_chains}
\label{algo:split_and_connect_neighbouring_chains}
\end{algorithm}

Method $split\_intersecting\_chains$ is  described in \Cref{algo:split_intersecting_chains}. Given an endpoint ray direction, we iterate over all the intersecting chains in that  $direction$. Given a chain to be split, $inter\_chain$, and the node, $split\_node$, we divide the chain nodes in two chains cutting the nodes list in the position of $split\_node$. Remember that the list of nodes within a chain is clockwise sorted. After splitting the chain in $sub\_ch1$ and $sub\_ch2$, we select the sub chain that does not intersect $inter\_chain$, line 5. Then, if $Ch_k$ intersects $Ch_j$ in the other endpoint, this means that $inter\_chain$ intersects  $Ch_j$ at both endpoints, we repeat the logic over the other endpoint but for $Ch_k$ instead of $inter\_chain$. The split chain list is returned in $l\_search\_chains$.

\begin{algorithm}[!htbp]
  \KwIn{$direction$, // endpoint direction for split chains\\
  $l\_filtered\_chains$, // list of chains to be split \\
  $Ch_j$, // source chain. The one that is being to connect if conditions are met\\
  }
\KwOut{split chain list}
$l\_search\_chains$ $\leftarrow$ []\\
\For{$inter\_chain$ in $l\_filtered\_chains$}{
$split\_node$ $\leftarrow$get\_node\_by\_angle($direction$)\\
$sub\_ch1$, $sub\_ch2$ $\leftarrow$ split\_chain($inter\_chain$, $split\_node$)\\
$Ch_k$ $\leftarrow$ select\_no\_intersection\_chain\_at\_endpoint($sub\_ch1$, $sub\_ch2$, $Ch_j$, $direction$)\\
\tcc{Longest chains intersect two times}
\If{intersection\_between\_chains($Ch_k$, $Ch_j$)}{
$split\_node\_2$ $\leftarrow$ get\_node\_by\_angle($node\_direction\_2$)\\
$sub\_ch1$, $sub\_ch2$ $\leftarrow$ split\_chain($Ch_k$, $split\_node\_2$)\\
$Ch_k$ $\leftarrow$ select\_no\_intersection\_chain\_at\_endpoint($sub\_ch1$, $sub\_ch2$, $Ch_j$, $node\_direction\_2$)\\
}
change\_id($Ch_k$)\\
$l\_search\_chains$ $\leftarrow$ $l\_search\_chains$ + $Ch_k$\\
}
\SetAlgoLined
\LinesNumbered
\BlankLine 
\KwRet{$l\_search\_chains$}
\caption{split\_intersecting\_chains}
\label{algo:split_intersecting_chains}
\end{algorithm}

{\small
\bibliographystyle{siam}
\bibliography{TreeRingDetection}
}